%% file: _main.tex
\input{_constants}
\arxiv

\pdfoutput=1
\documentclass[10pt,twocolumn,letterpaper]{article}
\input{cvpr_header}

\begin{document}
\title{\paperTitle}
\author{\authorBlock}
\maketitle

\input{00_abstract}
\input{01_intro}

\input{02_related}

\input{03_method}
\input{10_conclusion}

{\small
\bibliographystyle{ieeenat_fullname}
\bibliography{11_references}
}

\ifarxiv \clearpage

\maketitlesupplementary

\onecolumn
\tableofcontents
\clearpage
\addcontentsline{toc}{section}{List of Figures}
\listoffigures
\clearpage

\twocolumn

\appendix \input{12_appendix} \fi
 \end{document}

%% file: _constants.tex
\def\paperID{38250}
\def\confName{CVPR}
\def\confYear{2026}

\def\paperTitle{RoadTones: Tone Controllable Text Generation from Road Event Videos}

\def\authorBlock{
    Chirag Parikh\thanks{Equal contribution} \qquad
    Siddhi Pravin Lipare\footnotemark[1] \qquad
    Ravi Kiran Sarvadevabhatla \\
    CVIT \& iHub-Data, IIIT Hyderabad, India \\
}

\newif\ifreview 
\newif\ifarxiv \newcommand{\arxiv}{\arxivtrue}
\newif\ifcamera 
\newif\ifrebuttal 

%% file: cvpr_header.tex
\ifreview \usepackage[review]{cvpr} \fi
\ifarxiv \usepackage[pagenumbers]{cvpr} \fi
\ifrebuttal \usepackage[rebuttal]{cvpr} \fi
\ifcamera \usepackage{cvpr} \fi

\input{_macros}  

\usepackage{xr-hyper}

\makeatletter
\newcommand*{\addFileDependency}[1]{
  \typeout{(#1)}
  \@addtofilelist{#1}
  \IfFileExists{#1}{}{\typeout{No file #1.}}
}

\makeatother

\definecolor{cvprblue}{rgb}{0.21,0.49,0.74}
\usepackage[pagebackref,breaklinks,colorlinks,allcolors=cvprblue]{hyperref}

\usepackage[capitalize]{cleveref}
\crefname{section}{Sec.}{Secs.}
\crefname{table}{Table}{Tables}
\crefname{figure}{Fig.}{Figs.}

\ifarxiv \crefname{appendix}{App.}{Apps.}
\else \crefname{appendix}{Suppl.}{Suppls.} \fi

%% file: _macros.tex
\usepackage{graphicx}	
\usepackage{amsmath}	
\usepackage{amssymb}	
\usepackage{booktabs}
\usepackage{times}
\usepackage{microtype}
\usepackage{epsfig}
\usepackage{caption}
\usepackage{float}
\usepackage{placeins}
\usepackage{color, colortbl}
\usepackage{stfloats}
\usepackage{enumitem}
\usepackage{tabularx}
\usepackage{xstring}
\usepackage{multirow}
\usepackage{xspace}
\usepackage{url}
\usepackage{subcaption}
\usepackage{xcolor}
\usepackage{soul}
\usepackage[normalem]{ulem}
\usepackage[hang,flushmargin]{footmisc}
\usepackage[most]{tcolorbox}
\usepackage{circledsteps}
\usepackage[table]{xcolor}
\usepackage{array}
\usepackage{arydshln}

\definecolor{lightgray}{RGB}{226,226,226}  
\definecolor{darkpink}{RGB}{102,0,102}  
\definecolor{lightpink}{RGB}{255,210,239} 
\definecolor{darkbrown}{RGB}{102,51,0} 
\definecolor{darkblue}{RGB}{0,0,153}  
\definecolor{palegreen}{RGB}{0,108,108}  
\definecolor{paleyellow}{RGB}{255, 253, 158}
\definecolor{lightergray}{RGB}{241, 241, 241}
\definecolor{lightorange}{RGB}{252, 229, 205}
\definecolor{darkorange}{RGB}{180,95,6}
\definecolor{lightpurple}{RGB}{220,167,255}
\definecolor{lavender}{RGB}{153, 51, 255}
\definecolor{mangoyellow}{RGB}{255, 229, 153}
\definecolor{lightbrown}{RGB}{45,37,35}  

\pgfkeys{/csteps/inner color=white}
\pgfkeys{/csteps/outer color=darkbrown}
\pgfkeys{/csteps/fill color=darkbrown}

\DeclareRobustCommand{\CircledColored}[2]{%
\begingroup
\pgfkeys{/csteps/.cd,
inner color=white,
outer color=#1,
fill color=#1}%
\Circled{#2}%
\endgroup
}

\DeclareRobustCommand{\CircledBlue}[1]{\CircledColored{darkblue}{#1}}

\definecolor{Gray}{gray}{0.85}
\definecolor{SkyBlue}{rgb}{0.88,1,1}
\definecolor{PasteGreen}{RGB}{204, 226, 215}
\definecolor{PastePink}{RGB}{253, 223, 236}

\newcolumntype{a}{>{\columncolor{Gray}}c}

\definecolor{LightYellow}{RGB}{252, 255, 125}
\definecolor{PasteYellow}{RGB}{254, 255, 214}
\definecolor{PasteLavender}{RGB}{229, 228, 244}
\definecolor{NewGreen}{RGB}{17,207,156}
\definecolor{pinkcircle}{RGB}{255,138,189}
\definecolor{lightlavender}{RGB}{220,167,255}

\newcommand{\hlpastegreen}[1]{{\sethlcolor{PasteGreen}\hl{#1}}}

\newcolumntype{g}{>{\columncolor{gray!15}}c}



%% file: 00_abstract.tex
\begin{abstract}
Existing video-language models can generate factual descriptions of road events but lack control over how these events are expressed: their tone, urgency, or style. This limits deployment in communication-critical settings where the effectiveness of a message depends on both content and presentation, not just factual accuracy. To mitigate this, we introduce a comprehensive dataset-model-evaluation suite for tone-controllable road video captioning. Our human-validated data generation pipeline expands road-video corpora with diverse tonal annotations and multi-tone captions, yielding the \textbf{RoadTones-51K} dataset. We propose \textbf{RoadTones-VL-CoT}, a controllable video-to-text model that also generates tone-conditioned Chain-of-Thought intermediate drafts for interpretability. We also introduce \textbf{RoadTones-Eval}, a new evaluation suite that jointly measures factual consistency and tone adherence. In addition, we conducted a user study whose results validate caption quality, tone control, and factual consistency. Together, these contributions lay the foundation for context-sensitive tone-controllable video captioning. 
\end{abstract}


%% file: 01_intro.tex

\section{Introduction}
\label{sec:intro}

Understanding road events from videos is central to modern AI-enabled mobility systems and applications - from intelligent traffic monitoring and post-incident analysis to citizen-friendly public safety reporting via social media. In ADAS/AV pipelines, video understanding has positive implications for driver monitoring, behavior attribution and fleet-level quality assessment. Recent video-language models partially address this need by generating neutral, factual text descriptions of road events~\cite{Dolphin, roadsocial, bdd-x, drama, SUTD-TrafficQA, LingoQA, embodied_driving, DriveLM}. But often, it is not enough to convey only factual information. For effective human-AI communication, systems must also know \textit{how} to stylize the content, with the right \textit{tone} for the use case, audience, or intent.


Imagine a video where a motorcycle abruptly cuts in front of a car during dense traffic.
An ADAS/AV engineering team reviewing the clip may require a neutral, analytical description for logs and post-drive analysis: ``Sudden motorcycle cut-in from the right; ego vehicle performs controlled deceleration; no lane instability observed\#MotorcycleCutIn". This supports scenario tagging, system debugging, and incident triage. A municipal transport authority preparing a safety advisory would prefer an instructional and civic tone: ``Sudden cut-ins are common in peak traffic. Keep safe following distance and slow early to avoid collisions." For the general public on social platforms, an engaging tone may be more effective: ``Rush-hour chaos strikes again{\includegraphics[width=0.02\textwidth]{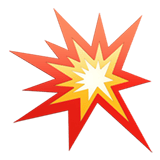}}. Watch those surprise cut-ins and stay alert out there. \#DriveSafe". These perspectives show that different stakeholders - engineers, agencies, and everyday road users - often need different narrative framings for the exact same road event, and tone is the mechanism that adapts communication to each audience.

However, current road-video datasets and models treat event captions as fixed, offering no control over tone~\cite{Dolphin, drama, bdd-x, roadsocial, embodied_driving, LingoQA}. Even when captions are expressive~\cite{roadsocial}, each video clip typically has a single or small, fixed set of associated tones. This mismatch is especially limiting for workflows that must generate both human-readable safety summaries and public-facing explanations. There are no mechanisms to (i) generate alternate tones for the same video, (ii) control tone intensity, hashtags, emoji usage, or (iii) evaluate tone adherence alongside content fidelity. 

\begin{figure}[!ht]
    \centering
    \includegraphics[width=1\linewidth]{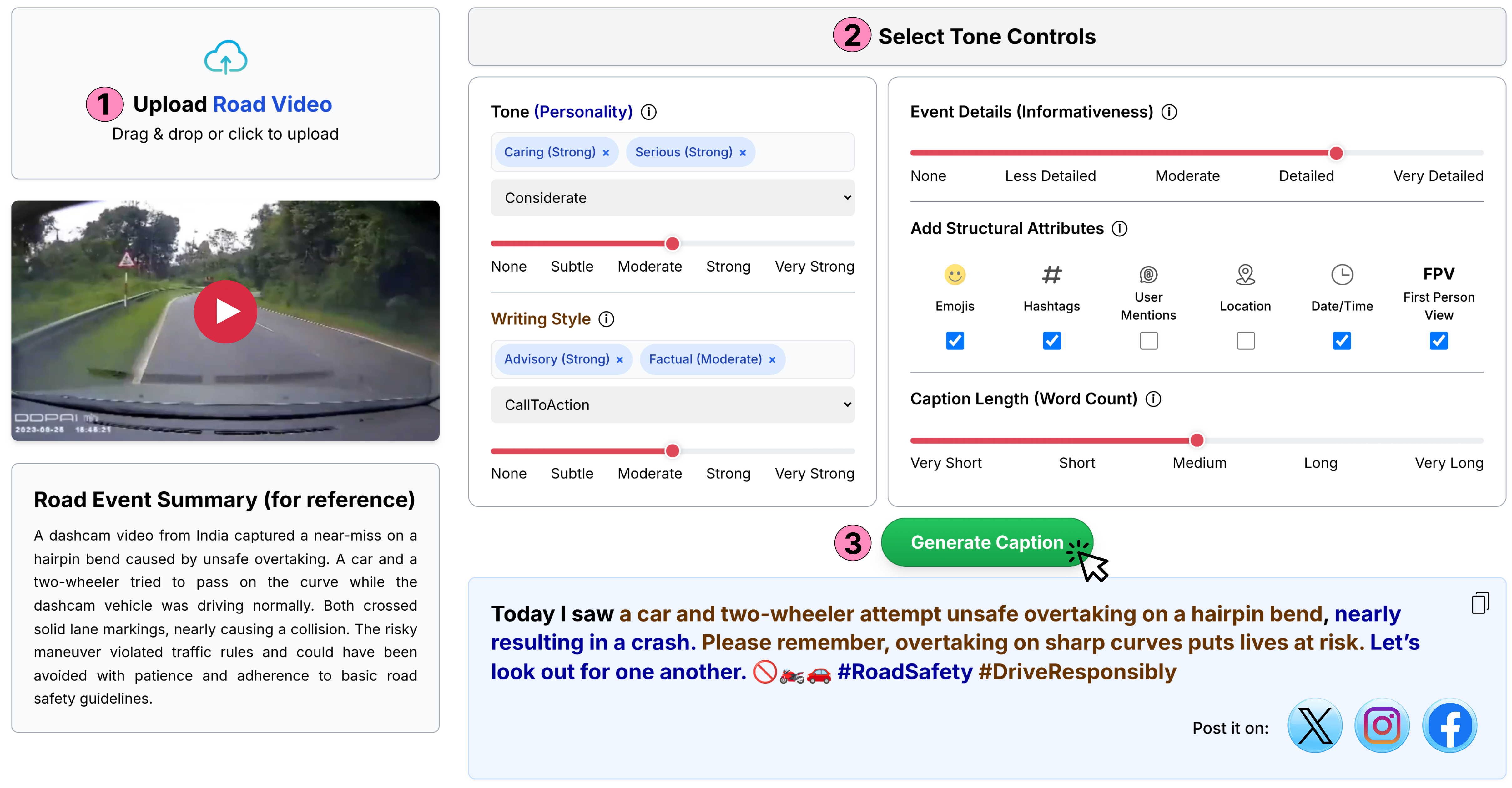}
    \caption{\textbf{Tone-Controlled Road Video Captioning. } 
Our user interface demonstrates captioning capability with fine-grained control across five tone  dimensions: \textcolor{darkblue}{Personality} \textcolor{darkbrown}{Writing Style}, Event Details (Informativeness), Structural Attributes, and Caption Length. 
\CircledText[inner color=black, outer color=black, fill color=pinkcircle]{\textbf{1}} The user uploads a Road Video, \CircledText[inner color=black, outer color=black, fill color=pinkcircle]{\textbf{2}} selects the desired controls with specific intensities, and adds structural attributes such as `Emojis' and `Hashtags'.
\CircledText[inner color=black, outer color=black, fill color=pinkcircle]{\textbf{3}}. Our approach then generates a caption that reflects all specified controls, as shown in the output. In the generated caption above, ``Let's look out for one another'' reflects a \textbf{``Caring''} \textcolor{darkblue}{Personality} while ``Please remember\dots \#DriveResponsibly'' reflect an \textbf{``Advisory''} \textcolor{darkbrown}{Writing Style}. The Road Event Summary is provided for reference. As additional utility, the generated content can be posted on social media platforms (e.g. X, Instagram).}
    \label{fig:main-ui}
\end{figure}

To fill this gap, we introduce a comprehensive dataset-model-evaluation stack for tone-controlled road video captioning. Our approach is designed to be modular yet cohesive. A human-validated data generation pipeline generates tone-rich captions at scale, enabling fine-grained tone control (see \cref{fig:caption-after-one-change}). A video-language model maps video and tone signals to factual-yet-stylized outputs (see \cref{fig:main-ui}). A new evaluation suite provides tools to benchmark both tone fidelity and factual consistency. This capability supports not only public communication but also for creating multiple narrative forms of the same event (e.g. for debugging, driver-behavior annotation, human-in-the-loop review). We validate our system through a user study, showing that our captions are perceived as high-quality, tone-aligned and factually consistent.


Our key contributions are:
\begin{itemize}
\item \textbf{A scalable human-validated data generation pipeline} that augments a road-video dataset with rich tonal annotations and multiple tone-controlled captions per clip, resulting in the \textbf{RoadTones-51K dataset}.
\item \textbf{\textsc{RoadTones-VL-CoT}:} A multi-tone controllable video-to-text model that also generates Chain-of-Thought style intermediate caption drafts for partial interpretability.
\item \textbf{RoadTones-Eval Suite:} New evaluation metrics,  benchmarks to assess tone adherence and factual consistency.
\end{itemize}


Broadly, our work provides a unified dataset-model-evaluation framework for multi-axis tone control in road videos, enabling audience-adaptive communication across mobility, AV/ADAS development, safety monitoring, and public engagement.

%% file: 02_related.tex
\section{Related Work}
\label{sec:related}


\textbf{Road Event Video Captioning:} Several recent datasets pair road videos with textual annotations, enabling Video LLMs to describe on-road scenes and events \cite{Dolphin, roadsocial, LingoQA, SUTD-TrafficQA, embodied_driving, IDD-X}. Many of the representative works focus on dashcam perspectives \cite{Dolphin, LingoQA, bdd-x, DriveLM, IDD-X}, while others broaden coverage to diverse geographies and camera viewpoints \cite{roadsocial}. These efforts have advanced road event video understanding, but provide a one-size-fits-all factual description per clip and do not expose style or tone as a controllable dimension. Even when social-media-style captions are included~\cite{roadsocial}, tone per video is typically limited and lacks reusable control signals that would let a system restyle the same event for different audiences or risk levels. This limits downstream applications that require socially engaging, context-specific communication.


\textbf{Controllable Video Captioning:} Traditional video captioning aims to generate a single, generic description of a video. In contrast, controllable video captioning allows users to guide the generated caption using specific constraints that influence both linguistic form and semantic content. Prior work has explored control mechanisms such as exemplar sentences, Part-of-Speech (POS) sequences, or the desired caption length \cite{CVC_exemplar, CVC_POS, length_CVC}. Wang et al. \cite{emvidcap} introduce a framework that injects the emotional states of people in the video into its factual description. Other efforts propose object-oriented control that centers the generated caption around user-specified entities \cite{intentvcnet, liu2021o2na}. Despite these advances, no existing work offers control over the  communicative \emph{tone}-a crucial dimension involving multiple stylistic attributes (e.g., assertive, critical, sarcastic).

%


\textbf{Controllable Image Captioning:} Image captioning has progressed from neutral, factual descriptions to subjective control over attributes such as length, sentiment, and persona \cite{captionsmiths, senticap, factual_emotional, personality-cap, mscap}. CaptionSmiths offers flexible control of length and descriptiveness for factual captions \cite{captionsmiths}. Some methods generate captions with target styles (humorous, romantic) or positive/negative sentiment while preserving core content \cite{senticap, factual_emotional, mscap}. Personality-Captions conditions generation on a single personality trait from a large category inventory to produce engaging text \cite{personality-cap}. However, these approaches control one attribute at a time, whereas real-world captions often blend multiple stylistic cues (e.g., cautious yet empathetic). Our work enables \textit{simultaneous} control over multiple attributes corresponding to style, persona, and length, while preserving factual content. Crucially, it extends these capabilities from static images to the challenging and dynamic setting of road event videos.




%% file: 03_method.tex
\begin{figure*}
    \centering
    \includegraphics[width=1\linewidth]{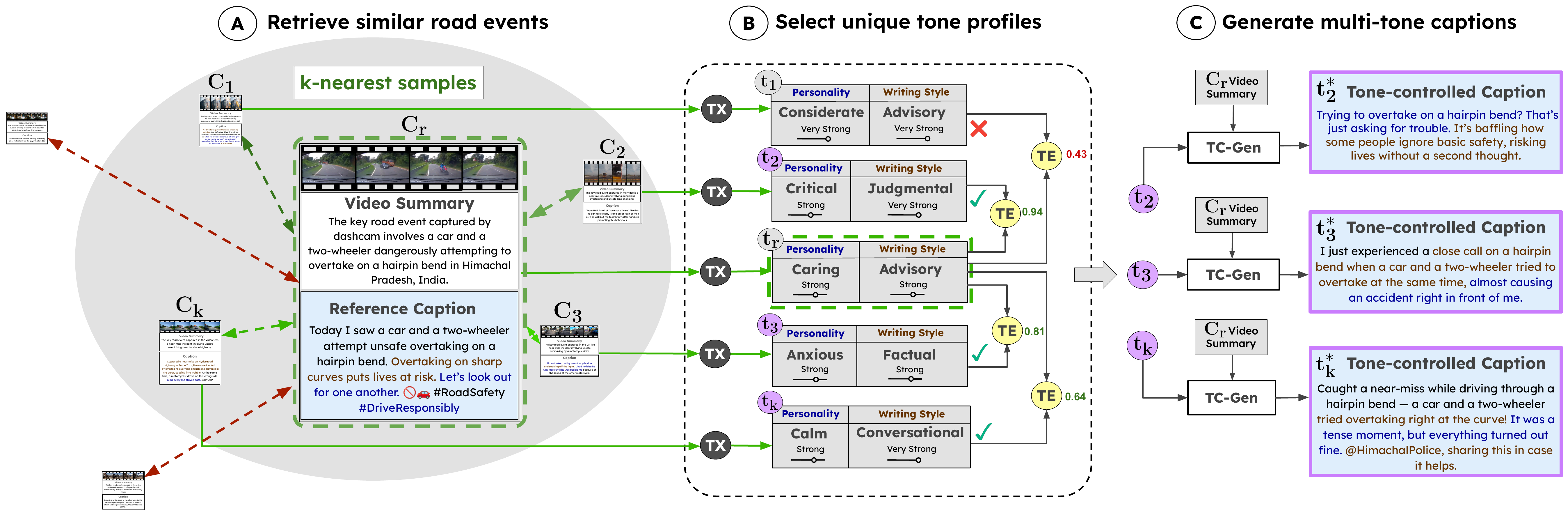}
    \caption{ \textbf{Generating Distinct Tone Captions Per-Video.} \CircledText[inner color=black, outer color=black, fill color=white]{A} Given a reference video $C_r$, we first retrieve similar road events using a $\boldsymbol{k}$-nearest neighbor approach. \CircledText[inner color=black, outer color=black, fill color=white]{B} We obtain tone profiles from captions using the Tone Extractor (\CircledText[inner color=white, outer color=darkgray, fill color=darkgray]{TX}, \cref{ssec: tone_extract}). The Tone Evaluator (\CircledText[inner color=black, outer color=black, fill color=paleyellow]{TE},\cref{sec:tone_eval}) then selects the tone profiles most dissimilar to the reference \CircledText[inner color=black, outer color=black, fill color=lightgray]{\textbf{$t_{r}$}}. \CircledText[inner color=black, outer color=black, fill color=white]{C} The selected tone profiles (\CircledText[inner color=black, outer color=black, fill color=lightpurple]{\textbf{$t_2$}}, \CircledText[inner color=black, outer color=black, fill color=lightpurple]{\textbf{$t_3$}}, \CircledText[inner color=black, outer color=black, fill color=lightpurple]{\textbf{$t_k$}}) are fed to   \CircledText[inner color=black, outer color=black, fill color=white]{TC-Gen} (\cref{ssec:caption_generation}) to produce distinct tone-controlled captions for the video. For e.g. the \textcolor{darkbrown}{Advisory} {Writing Style} in the reference caption changed to \textcolor{darkbrown}{Judgemental} in the tone caption $t_2^*$. The colors in the generated captions map to blue for \textcolor{darkblue}{Personality} and brown for \textcolor{darkbrown}{Writing Style}.}
    \label{fig:distinct-tone-captions-per-video}
\end{figure*}

\section{Tone Components}
\label{sec:tonetax}

\begin{figure*}
    \centering
    \includegraphics[width=0.9\linewidth]{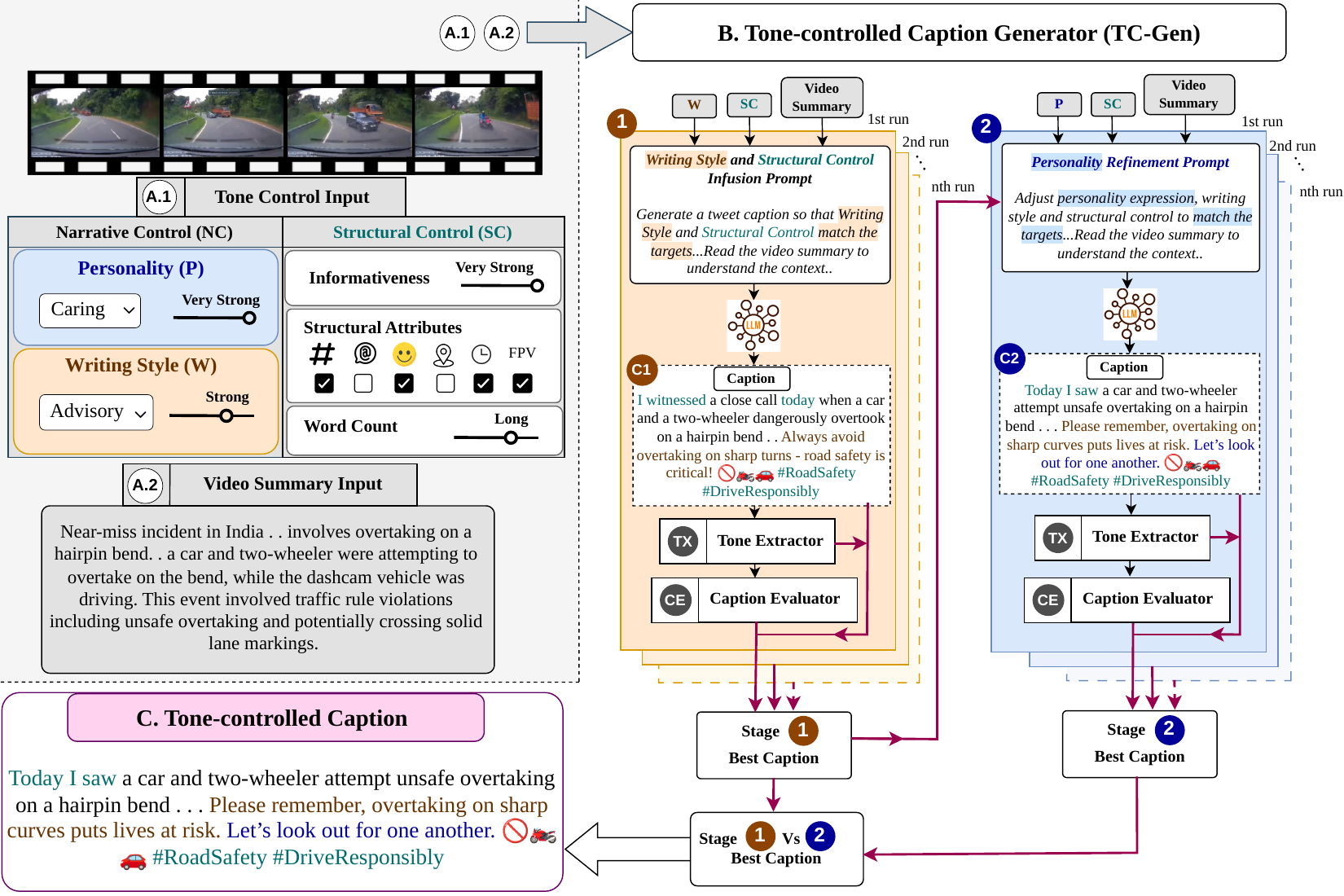}
    \caption{\textbf{Tone-controlled Caption Generation pipeline.} Inputs \CircledText[inner color=black, outer color=black, fill color=white]{A.1} and \CircledText[inner color=black, outer color=black, fill color=white]{A.2} include the target Tone Controls (Narrative and Structural) and a detailed video summary respectively. The inputs are fed to \CircledText[inner color=black, outer color=black, fill color=white]{B. Tone‑controlled Caption Generator}, a two‑stage pipeline (\Circled{1}, \CircledBlue{2}). At each stage, the generator conditions on the pipeline inputs to produce candidate captions. Stage \Circled{1} infuses \textcolor{darkbrown}{Writing style} and enforces \textcolor{palegreen}{Structural Controls}. Stage  \CircledBlue{2} restyles Stage-\Circled{1} caption with \textcolor{darkblue}{Personality} controls and jointly recalibrates, preserving all previous controls. The quality of the candidates produced at each stage is evaluated by extracting the realized narrative and structural attributes' profiles (\CircledText[inner color=white, outer color=darkgray, fill color=darkgray]{TX} Tone Extractor \cref{ssec: tone_extract}) and computing alignment scores along with factual consistency using \CircledText[inner color=white, outer color=darkgray, fill color=darkgray]{CE} Caption Evaluator \cref{sec:tone_eval}. The top scoring candidate is selected from each stage (Stage-wise Best Caption) and compared against each stage’s best caption (Cross Stage Best Caption) to return the final \CircledText[inner color=black, outer color=darkpink, fill color=lightpink]{C. Tone-controlled Caption}.  
    }
    \label{fig:caption-generation-pipeline}
\end{figure*}

We define tone as the attitude or stance conveyed by the captioner through wording, phrasing, hashtags, emojis, and other expressive markers. In this section, we introduce the controllable components that determine how tone is expressed in a caption. 
To enable precise, context-sensitive tone control in video captioning, we adopt a modular framework that decomposes tone into two interpretable and adjustable components: Narrative Control (NC) and Structural Control (SC) - see \CircledText[inner color=black, outer color=black, fill color=white]{A.1} Tone Control Input in \cref{fig:caption-generation-pipeline}.




\textbf{Narrative Control (NC)} governs the \textit{narrative voice} by specifying two key aspects of the captioner - their \textit{personality} (e.g., sarcastic, caring, serious) and their  \textit{writing style} (e.g., conversational, metaphorical, instructional). For personality attributes, we adopt the full set of 215 traits from Shuster et al.~\cite{personality-cap}, ensuring coverage and comparability.  For writing style, we curate a list of 16 unique attributes (see \cref{ssec:tone_extract_appendix} for the detailed schema). Regardless of aspect (viz. \textit{personality, writing style}), each tone attribute is also associated with an intensity (`Absent' to `Very Strong'), mapped to a [0,1] scale, where higher values indicate stronger presence in the caption. Together, these elements shape the message flavor, allowing the caption to resonate with the intended audience. 

 \textbf{Structural Control (SC)} regulates the \textit{presentation structure}-how the message is visually and contextually formatted. It includes settings for verbosity (e.g., compact vs. elaborate), inclusion of expressive markers like emojis and hashtags, and contextual cues such as time, location, or first-person framing and an informativeness score [0,1] that reflects how much factual content a caption conveys relative to a neutral, detailed description of the road video. 

The clean separation between NC and SC supports compositional tone specification, interpretable generation processes, and scalable dataset construction, all of which are essential for building tone-controllable captioning systems.

\section{Dataset Creation}
\label{sec:data_creation}

We build our dataset on top of RoadSocial \cite{roadsocial}, a large‑scale VideoQA corpora for generic road event understanding. We select RoadSocial because the community‑authored social‑media captions accompanying its videos naturally exhibit wide variation in tone and expressivity, providing a rich basis for tonal supervision. We extract video summary via QA aggregation and use it along with the social captions to extract tone profiles and drive our data generation pipeline.

A core objective is to augment existing road video captions with meaningful yet distinct tone-controlled variants. To obtain these meaningful caption variants, we design a nearest neighbor based selection procedure (see \cref{fig:distinct-tone-captions-per-video}). For a reference video $C_r$ in RoadSocial, we first retrieve its $k$ nearest neighbors ($C_1\ldots C_k$) that are semantically similar in terms of road event type (e.g. near-miss, road rage, defensive driving). 
We then extract the tonal content ($t_r$) of the reference and that of the neighbors' captions ($t_1\ldots t_k$) using the Tone Extractor (\CircledText[inner color=white, outer color=darkgray, fill color=darkgray]{TX}, \cref{ssec: tone_extract}). Next, we compare the extracted tones with the Tone Evaluator (\CircledText[inner color=black, outer color=darkgray, fill color=paleyellow]{TE}) and compute the pairwise dissimilarity  (i.e. $d(t_i,t_j);i,j\in \{1,2,\ldots k;r\}$). The most dissimilar tones (e.g., $t_2, t_3, t_k$ in \cref{fig:distinct-tone-captions-per-video}) are selected and fed to the tone-controlled caption generator (\CircledText[inner color=black, outer color=black, fill color=white]{TC-Gen}, \cref{ssec:caption_generation}), producing multiple, distinct tone variants for the same factual event. Thus, our data pipeline ensures that generated captions for each video are diverse in tone yet grounded in similar event contexts.





We apply this pipeline for every video in RoadSocial and obtain the RoadTones-51K dataset. We describe some key components of the pipeline in the content that follows.

\subsection{\CircledText[inner color=black, outer color=black, fill color=white]{TC-Gen} Tone‑Controlled Caption Generation }
\label{ssec:caption_generation}

In this section, we describe how the tone settings (\CircledText[inner color=black, outer color=black, fill color=white]{A.1} in \cref{fig:caption-generation-pipeline}) obtained via nearest neighbor selection procedure are used within our multi‑stage pipeline to produce tone‑controlled captions for road videos. 

\textbf{Stage \Circled{1} \textcolor{darkbrown}{Writing Style} and \textcolor{palegreen}{Structural Control} Infusion:} Conditioned on  video summary (\CircledText[inner color=black, outer color=black, fill color=white]{A.2} in \cref{fig:caption-generation-pipeline}), the target Writing style (W) and Structural Control (SC) inputs, an LLM~\cite{openai2025gpt4.1} is prompted to generate a candidate tone-conditioned caption \Circled{C1}. We extract the tone content from generated caption using the Tone Extractor (\CircledText[inner color=white, outer color=darkgray, fill color=darkgray]{TX}) and feed it to the Caption Evaluator (\CircledText[inner color=white, outer color=darkgray, fill color=darkgray]{CE}, \cref{sec:tone_eval}) to yield a score. We set a high temperature value for LLM decoding, thereby  generating multiple candidate captions across multiple runs ($1,2,\ldots n$). The highest scoring candidate (Stage-\Circled{1}'s best caption) along with its tonal and structural contents are carried forward as input to the next stage.


\textbf{Stage \CircledBlue{2} \textcolor{darkblue}{Personality} Refinement:} Based upon the Personality controls provided as input, we prompt the LLM to restyle Stage-\Circled{1}'s best caption while retaining the previously met constraints.
As in Stage-\Circled{1}, we generate multiple candidates, re-extract tones, and select the caption that maximizes the quality score. Again, out of both the stages, the candidate with the highest caption quality score is returned as the pipeline's final \CircledText[inner color=black, outer color=darkpink, fill color=lightpink]{Tone-controlled Caption} as shown in \cref{fig:caption-generation-pipeline}.



\begin{figure*}
    \centering
    \includegraphics[width=0.9\linewidth]{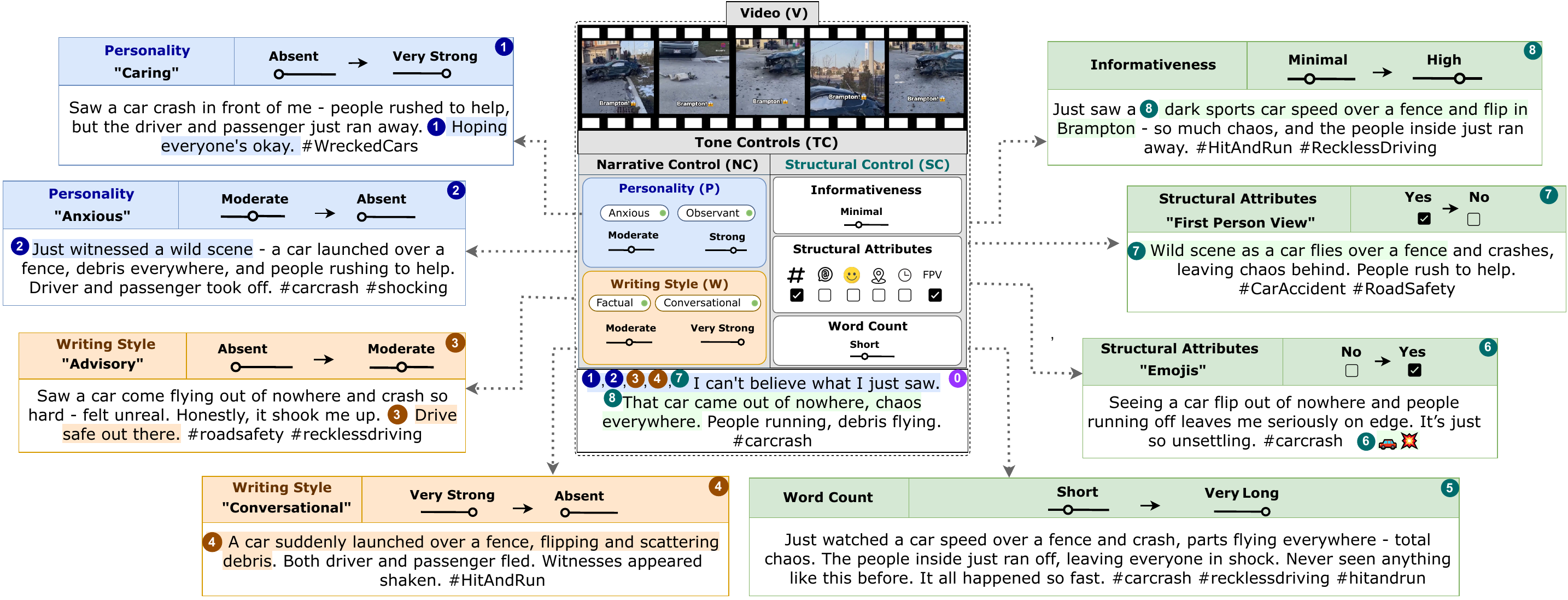}
    \caption{\textbf{Controlling individual tonal attributes in the generated caption.} The central panel in figure shows a video V, tone controls TC and its corresponding caption \CircledText[inner color=white, outer color=lavender, fill color=lavender]{0} from our dataset. The surrounding captions (\CircledBlue{1}-\CircledText[inner color=white, outer color=palegreen, fill color=palegreen]{8}) correspond to changes in one of the tonal attributes shown in their header. For e.g., caption \CircledBlue{1} was obtained by increasing the tonal intensity of \textbf{Caring} \textcolor{darkblue}{Personality} from Absent (0-0.2) to Very Strong (0.8-1.0) while keeping others fixed. This modified tone configuration was fed to our caption generator (\CircledText[inner color=black, outer color=black, fill color=white]{TC-Gen}, \cref{ssec:caption_generation}), yielding the changed caption. We highlight the key phrase in each caption, reflecting the modified tone controls. Our generator pipeline thus enables fine-grained control of tonal and structural attributes in road-video captioning. 
}
    \label{fig:caption-after-one-change}
\end{figure*}

Ablations on the order of application of two stages and their individual contribution towards tone-controlled caption generation along with the prompt details are provided in \cref{ssec:caption_generation_appendix}. 
In \cref{fig:caption-after-one-change}, we additionally show how altering different tone control values affects the final generated caption. The caption generation quality is validated through a user study described in \cref{sec:user_study}. The detailed prompts for each stage are provided in \cref{ssec:caption_generation_appendix}.

\subsection{\CircledText[inner color=white, outer color=darkgray, fill color=darkgray]{TX} Tone Extractor}
\label{ssec: tone_extract}

We adopt a four‑step, LLM‑based pipeline that processes a video summary together with its caption to extract per‑attribute tone control values. 
\textit{(1) Writing Style extraction:} the LLM~\cite{openai2025gpt4.1} scores the intensities of all writing style attributes given the caption’s expressive form. \textit{(2) Personality extraction:} the LLM shortlists relevant traits from 215 category inventory and assigns intensities based on the author's persona projected in the caption. 
\textit{(3) Informativeness extraction:} the LLM assesses how much factual content the caption conveys relative to a neutral, detailed reference description of the video. \textit{(4) Structural attribute extraction:} the LLM detects the presence of location and date/time information and identifies captioner's point-of-view (first‑person vs. external/third‑person). Hashtags, user mentions, emojis, and word count are extracted directly via regular expressions.
All prompts used in the four‑step pipeline are iteratively refined based on the alignment between the expected and the extracted tonal intensities. 
We additionally validate the change in intensity levels for tone attributes through a user study in \cref{sec:user_study}.  
Details on the finalized prompts for each extraction step are provided in \cref{ssec:tone_extract_appendix}.




\subsection{Data Statistics}

\begin{figure}[!t]
    \centering
    \includegraphics[width=1\linewidth]{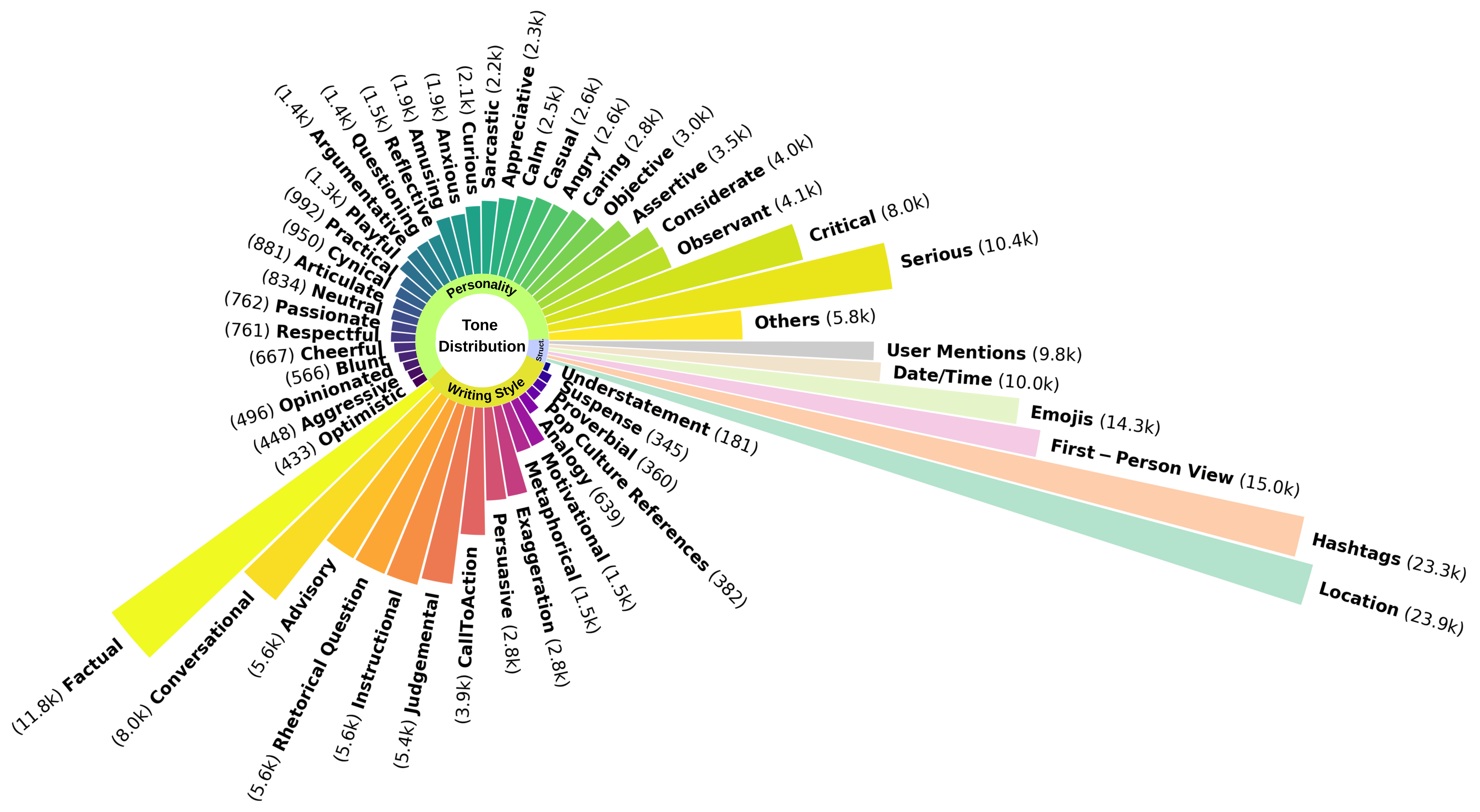}
    \caption{Distribution of Personality, Writing Style and Structural attributes (shown as ``Struct'') annotations in RoadTones-51K dataset. Only top 30 attributes of Personality tone are shown and remaining 86 are shown as ``Others''. }
    \label{fig:tone-distribution}
\end{figure}

Our resulting dataset contains 7681 road‑event videos paired with 51K tone‑aware captions (approx. 3 distinct tone control variants per clip and intermediate stages' outcomes for each variant). Each caption in our dataset is annotated with continuous intensity values for the narrative controls-Personality (215 traits) and Writing style (16 types)-and with 8 structural controls: informativeness, target word count, viewpoint (first‑person vs. external), hashtags, emojis, user mentions, location, and date/time. 
The RoadTones-51K captions comprise a total of over 129K dominant tone attribute annotations whose distribution is shown in \cref{fig:tone-distribution}. More stats in \cref{ssec:tone_stats_appendix}.

\section{VLM for Tone-Controllable Captioning}
\label{sec:vlm}

Given a road video and a target tone profile, our goal is to generate a caption that (i) conforms to the specified tonal attributes and (ii) remains factually accurate to the visual content. To this end, we train a Vision-Language Model (VLM) using instruction-tuning triplets \{road video, structured query with tone specification, tone-controlled caption with optional rationale\}, structured under our tone schema (\cref{sec:tonetax}). To further strengthen factual grounding, we include an auxiliary fine-tuning objective for video summarization using the triplets \{road video, summarization query, neutral detailed summary\}. The VLM is conditioned through a structured query encoding both the target tone profile and its interpretation. Each query template has four parts: (i) a task instruction (e.g., ``Describe the key road event shown in the video as a tone-controlled caption'') (ii) a Chain-of-Thought (CoT) flag requesting either a rationale+caption or a direct caption (iii) a tone schema defining control axes-Personality, Writing Style, and Structural controls-with attribute intensity levels and (iv) the target tone specification formatted per our taxonomy (\cref{sec:tonetax}). We retain continuous attribute intensities in [0,1].
Structural controls (e.g. viewpoint, hashtags, emojis, mentions, location, date/time) and target word count are represented as key-value pairs. Full templates, instruction variants, and examples are included in the \cref{sec:vlm_appendix}.

\textbf{\textsc{RoadTones-VL} w/ CoT:} We fine-tune Qwen3-VL~\cite{qwen3vl} to generate Chain-of-Thought (CoT) style reasoning alongside the final tone-controlled caption. CoT supervision is derived from intermediate outputs of our tone-controlled caption generation pipeline (\CircledText[inner color=black, outer color=black, fill color=white]{TC-Gen}), specifically the best captions after Stage~\Circled{1} (\textcolor{darkbrown}{Writing Style} and \textcolor{palegreen}{Structure} Infusion) and Stage~\CircledBlue{2} (\textcolor{darkblue}{Personality} Refinement; see \cref{fig:caption-generation-pipeline}). The model learns to (i) outline tonal calibration steps and (ii) produce the final tone-controlled caption while preserving factual content. As we shall see, this improves interpretability without compromising performance (Table~\ref{tab:roadtones_eval}). For flexibility, we mix CoT and non-CoT instruction-tuning triplets during training, enabling the model to generate either a direct caption or a rationale-augmented caption on demand.

\textbf{\textsc{RoadTones-VL} w/o CoT:} This variant is trained only with non-CoT instruction pairs to directly produce the final tone-controlled caption. It serves as an ablation counterpart to assess the effect of CoT supervision (\cref{sec:exp,sec:results}).

\begin{table*}[!t]
\centering
\resizebox{0.96\linewidth}{!}{
\begin{tabular}{c|l|c|ccc|cccc|g|g|g}
\toprule
\rowcolor{white} \multirow{2}{*}{\#} & \multirow{2}{*}{Model} & \multirow{2}{*}{Params} & \multicolumn{3}{c|}{Narrative Alignment} & \multicolumn{4}{c|}{Structural Alignment} & \multirow{2}{*}{TAS} & \multirow{2}{*}{FC} & \multirow{2}{*}{Overall}  \\
\cmidrule(rl){4-6} \cmidrule(rl){7-10}
\rowcolor{white} & & & P & WS & NAS & A & I & wc & SAS & & & \\ [1.2pt]
\cmidrule(r){1-1} \cmidrule(r){2-2} \cmidrule(r){3-3} \cmidrule(rl){4-6} \cmidrule(rl){7-10} \cmidrule(rl){11-11} \cmidrule(rl){12-12} \cmidrule(l){13-13} 

 \multirow{2}{*}{A} & \cellcolor{lightorange} GPT-5 \cite{gpt5} & - & 63.1 &	65.0 &	64.1 &	95.7 &	74.3 &	87.3 &	85.8 &	75.0 & 55.7 &  65.4 \\

& \cellcolor{lightorange} Gemini-2.5-pro \cite{gemini2.5pro} & - &  68.5 &	72.5 &	70.5 &	95.8 &	74.9 &	91.7 &	87.5 & \underline{79.0} &	\underline{59.4} & \underline{69.2} \\

\midrule
& \cellcolor{PasteGreen} Dolphins \cite{Dolphin} & 9B & 19.5 & 32.6 & 26.1 & 63.0 & 56.5 & 37.3  & 52.3 & 39.2 & 40.2 & 39.7 \\
B & \cellcolor{PasteGreen} RoboTron-Drive \cite{robotrondrive} & 8B & 27.8 & 32.9 & 30.3  & 61.1 & 58.1 & 32.9 & 47.8 & 39.1 & 41.5 & 40.3 \\
& \cellcolor{PasteGreen} RoadSocial \cite{roadsocial} & 7B &  40.8 &	33.7 &	37.2 &	63.4 &	73.0 &	52.2 &	62.9 &	50.1 & 44.5 & 47.3 \\

\midrule

&\cellcolor{PasteYellow} VideoLLaMA3 \cite{videollama3} & 7B & 40.8 & 43.6 & 42.2 & 68.9 & 64.0 & 12.0 & 48.3 & 45.3 & 41.7 & 43.5 \\
&\cellcolor{PasteYellow} InternVL-3.5 \cite{internvl3} & 8B & 50.3 &	52.4 &	51.3 &	77.8 &	64.8 &	22.9 &	55.2 &	53.3 &	44.1 & 48.7 \\
C &\cellcolor{PasteYellow} Qwen2.5-VL \cite{qwen2.5} & 7B &  37.5 &	37.1 &	37.3 &	65.0 & 66.0 &	18.9 &	50.0 & 43.7 & 48.3 & 46.0 \\
&\cellcolor{PasteYellow} MiniCPM-V 4.5 \cite{minicpmv} & 8B &  59.9 &	55.0 &	57.4 &	79.3 &	69.6 &	18.3 &	55.7 & 56.6 & 47.6 & 52.1 \\
&\cellcolor{PasteYellow} Qwen3-VL \cite{qwen3vl} & 8B &  60.7	& 57.9 &	59.3 &	82.6 &	69.0 &	17.5 &	56.4 & 57.9 & 52.8  & 55.4 \\

\midrule

\rowcolor{SkyBlue} \textbf{Ours (D)} & \textbf{\textsc{RoadTones-VL} w-CoT} & 8B & 72.7 &	81.5 &	77.1 &	98.2 &	74.5 &	94.5 &	89.1 &	\textbf{83.1} &	\textbf{57.7}&	\textbf{70.4} \\[1.4pt]

\rowcolor{SkyBlue} (Qwen3-VL ft.) & \textsc{RoadTones-VL} w/o-CoT & 8B & 72.7 &	81.6 &	77.2 &	98.1 &	74.2 &	94.1 &	88.8 &	83.0 &	57.2 & 70.1 \\[0.5pt]

\hdashline
\noalign{\vskip 3.8pt}


 \rowcolor{white} \multirow{4}{*}{\shortstack{Ablations (E)}}   &  Single-Cap & 8B & 70.9 & 79.5  & 75.2 & 98.4 & 69.5 & 92.1 & 86.6 & 80.9 & 53.0 & 67.0 \\

 & Single-Run & 8B & 72.1 & 80.8 & 76.5 & 97.9 & 73.2 & 94.2 & 88.4 & \cellcolor{white} 82.4 & \cellcolor{white} 56.3 & \cellcolor{white} 69.4 \\


\cmidrule(lr){2-13}

\rowcolor{white} &  RoadSocial-Cap & 8B & 60.0 &	64.9 &	62.4 &	93.9 &	68.1 &	77.7 &	79.9 &	71.2 &	46.2 &	58.7 \\

\rowcolor{white} &  \textbf{RoadTones}-Cap & 8B & 65.4 &	73.1 &	69.2 &	98.1 &	73.3 &	80.6 &	84.0 &	76.6 & 54.2	& 65.4 \\










\bottomrule
\end{tabular}
}
\caption{\textbf{Performance of our proposed method and zero-shot baselines on RoadTones-Eval.} Tone and factual alignment score abbreviations - Personality traits (P), Writing Style (WS), Structural Attributes (A), Informativeness (I), word count (wc), Narrative Alignment Score (NAS), Structural Alignment Score (SAS), Factual Consistency (FC), and Tone-controlled caption quality (Overall). Rows A, B, C correspond to zero-shot baselines using 2 closed-source general-purpose, 3 open-source driving-specific, and 5 open-source general-purpose VideoLLMs respectively. Rows D shows performance comparison between fine-tuned VideoLLMs and ours \textsc{RoadTones-VL}. Rows E, show ablation results of the fine-tuned model on different dataset variants. Qwen3-VL ft. model abbreviations: w-CoT (with Chain-of-Thought reasoning). All scores are in percentages (\%).}
\label{tab:roadtones_eval}
\end{table*}

\section{\CircledText[inner color=black, outer color=black, fill color=paleyellow]{TE} Tone Evaluation Metrics}
\label{sec:tone_eval}

We evaluate the tonal and factual consistency of a caption aspects along two complementary metrics, each ranging from $0$ to $1$ (higher is better): \textbf{Tone Alignment Score (TAS)} and \textbf{Factual Consistency (FC)}.




\noindent \textbf{Tone Alignment Score (TAS):} TAS quantifies how well a generated caption adheres to the specified tone controls. We first extract the caption’s tone profile using the Tone Extractor 
(\CircledText[inner color=white, outer color=darkgray, fill color=darkgray]{TX}, \cref{ssec: tone_extract}). Alignment is then computed across two complementary axes: \textit{Narrative Controls (NC)} and \textit{Structural Controls (SC)} (\cref{sec:tonetax}), corresponding to tone expression and surface form respectively.

\textit{Narrative Alignment Score (NAS)} (\CircledText[inner color=black, outer color=black, fill color=paleyellow]{TE} in \cref{fig:distinct-tone-captions-per-video}) quantifies how faithfully the generated caption reflects the target \textit{Personality} and \textit{Writing Style} settings. We employ the LLM-as-a-judge framework~\cite{llmjudge} to assess semantic alignment for each component, yielding $S_p$ and $S_w \in [0,1]$. An LLM~\cite{openai2025gpt4.1} is prompted to reward agreement when related or synonymous attributes appear at comparable intensities and penalize disagreements. The overall score is computed as $NAS = (S_w + S_p)/2$. Detailed scoring prompts are provided in the \cref{sec:tone_eval_appendix}.

\textit{Structural Alignment Score (SAS)} evaluates how well the caption matches the specified structural tone targets (\cref{sec:tonetax}). For scalar controls, we compare the predicted informativeness level $\hat{I}$ (inferred from the caption) against the specified target $I^\star \in [0,1]$ and compute $e_I = |\hat{I} - I^\star|$. Similarly, we compute word count error as $e_\ell = |\widehat{\ell} - \ell^\star| / \ell^\star$, where $\widehat{\ell}$ and $\ell^\star$ are the actual and target word counts, respectively. For each binary attribute $a \in \mathcal{A}$ (hashtags, emojis, mentions, location, time, viewpoint), we set $e_a = 1$ if the caption disagrees with the target and $e_a = 0$ otherwise. The final structural alignment score is: $SAS = 1 - \frac{e_I + e_\ell + \sum_{a \in \mathcal{A}} e_a}{2 + |\mathcal{A}|}$,
where higher values indicate stronger structural fidelity.

The final Tone Alignment Score is the average of the narrative and structural alignments: $TAS = (NAS + SAS) / 2$.

\textbf{Factual Consistency score (FC)} measures the factual quality of a tone-controlled caption relative to a neutral, detailed video summary. We adopt the LLM-as-a-judge paradigm~\cite{llmjudge}, prompting a language model to compare the set of factual statements expressed in the caption against those in the summary. The LLM is instructed to ignore differences in tone, style, or phrasing (e.g., emojis or hashtags) and focus solely on semantic overlap, factual accuracy, and absence of hallucinations or contradictions. The final score $FC \in [0,1]$ is determined by the model's rubric-based evaluation. Scoring prompts and the detailed rubric are included in the \cref{sec:tone_eval_appendix}.

Finally, we report an overall quality score for each caption by averaging tone and factual dimensions: $Overall = (TAS + FC)/2$. This unified metric serves as the core output of our Caption Evaluator (\CircledText[inner color=white, outer color=darkgray, fill color=darkgray]{CE} in \cref{fig:caption-generation-pipeline}) and is used to assess generation quality (\CircledText[inner color=black, outer color=black, fill color=paleyellow]{TE} in \cref{fig:distinct-tone-captions-per-video}).

\section{Experiments}
\label{sec:exp}

We ensure strict separation of videos for training and evaluation. The dataset is divided into 5398/1523 clips for train/val, with no overlap. RoadTones-Eval uses 760 clips (2.2K tone-controlled captions) excluded from training and validation. 


\textit{Model training and evaluation details:} For \textsc{RoadTones-VL}(w/CoT), CoT and non-CoT data are mixed in a 1:3 ratio. Both \textsc{RoadTones-VL}(w/CoT) and \textsc{RoadTones-VL}(w/o-CoT) involve fine-tuning all layers of the base Qwen3-VL-8B-Instruct model~\cite{qwen3vl}-vision encoder, MLP projector, and LLM-with learning rates of $1\times10^{-5}$, $1\times10^{-5}$, and $2\times10^{-6}$, respectively. Training uses 4 $\times$H100 GPUs (effective batch size 64) for one epoch with 2 fps video sampling. 

Captions generated by \textsc{RoadTones-VL} (excluding CoT responses) are assessed using the same metrics - Tone Alignment Score (TAS) and Factual Consistency (FC) with per-component scores (Narrative and Structural Alignment) reported for detailed analysis. Neutral video summaries are hidden during inference and used only for FC computation.



\textit{Zero-shot:} We benchmark 10 Video-LLMs in zero-shot mode : two commercial (rows A in \cref{tab:roadtones_eval}), three driving-specific open-source (B), and five general-purpose open-source (C). All models receive the same structured tone query adapted to their native prompt format and are evaluated with identical metrics and settings.


 \textbf{Ablations:} To isolate the effects of tone supervision, reasoning, and data design, we conduct the following studies: \textit{Single-Cap:} Fine-tuning with only one tone-controlled caption per video (vs.\ multiple tone profiles) tests whether tone diversity improves controllability and content preservation. 
 \textit{Each-Stage Single-Run:} Training targets are reconstructed using a single sample per stage-removing stage-wise best-of-$n$ resampling and TAS/FC-based selection (\cref{ssec:caption_generation})-to quantify the contribution of multi-sample validation. 
 \textit{RoadSocial-Cap:} Fine-tuning on original RoadSocial captions (conditioned on extracted tone profiles) provides a comparison between raw social captions and our pipeline-generated tone-controlled captions. Both are evaluated on RoadSocial’s split for fairness. All ablations use the same architecture and hyperparameters as \textsc{RoadTones-VL}(w/o CoT).

\begin{figure}
    \centering
    \includegraphics[width=1\linewidth]{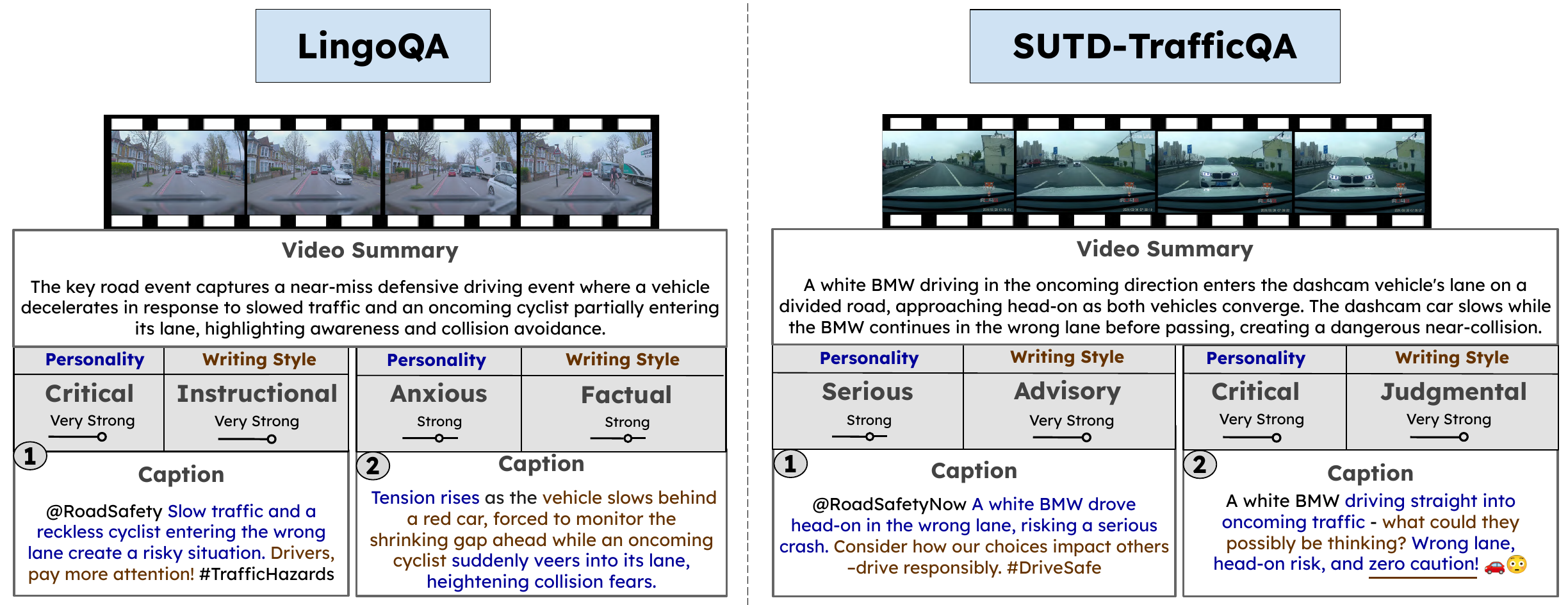}
    \caption{\textbf{Applicability of our tone-controlled caption generation pipeline on popular road video datasets: SUTD-TrafficQA\cite{SUTD-TrafficQA} and LingoQA\cite{LingoQA}}. For each video sample, we show two distinct tone captions (\CircledText[inner color=black, outer color=black, fill color=gray!35]{1},\CircledText[inner color=black, outer color=black, fill color=gray!35]{2}) with corresponding dominant \textcolor{darkblue}{Personality} and \textcolor{darkbrown}{Writing Style} attributes highlighted in text. Video summary is shown for reference. More samples are provided in \cref{ssec:caption_generation_appendix}.
    }
    \label{fig:other-dataset-captions}
\end{figure}

\section{Results}
\label{sec:results}

We summarize key findings on RoadTones‑Eval (\cref{tab:roadtones_eval}), followed by observations on zero‑shot models, ablations and the utility of our data pipeline.

\textbf{\textsc{RoadTones‑VL} w/ CoT} achieves the best performance on the leaderboard (Overall: 70.4\%, Tone Adherence Score [TAS]: 83.1\%, Factual Consistency [FC]: 57.7\% - see last 3 columns of \cref{tab:roadtones_eval}). It surpasses the strongest zero-shot commercial model (Gemini‑2.5‑pro \cite{gemini2.5pro}) by 4.1\% in TAS and the strongest zero-shot open-source model (Qwen3‑VL-8B-Instruct \cite{qwen3vl}) by 25.2\% in TAS. Relative to zero-shot (Qwen3-VL \cite{qwen3vl}), our model improves Narrative alignment by 17.8\% and Structural alignment by 32.7\% after fine-tuning. \textsc{RoadTones‑VL} w/ CoT matches or slightly improves overall tonal quality of caption compared to the non‑CoT variant while also providing rationales, making tone infusion and refinement transparent. Qualitative results are provided in \cref{sec:results_appendix}.

Driving‑specific open‑source models (block B in \cref{tab:roadtones_eval}) lag behind in TAS and Overall despite domain familiarity, indicating limited narrative tone controllability without targeted supervision. Open‑source models struggle with Narrative tone alignment (Personality and Writing Style) more than Structural tone alignment (structural Attributes and Informativeness)  in zero-shot mode, suggesting the non-trivial nature of obtaining precise control over narrative tone without task‑specific tuning afforded by our dataset (i.e. \textit{RoadTones-51K}).

\noindent \textbf{Utility of our data generation pipeline:} We use our pipeline (\cref{fig:distinct-tone-captions-per-video,fig:caption-generation-pipeline}) to generate two distinct tone captions for each sample video (with associated summary extracted via QA aggregation) from SUTD-TrafficQA~\cite{SUTD-TrafficQA} and LingoQA~\cite{LingoQA} datasets. The outcomes shown in \cref{fig:other-dataset-captions} demonstrate our pipeline's utility in augmenting existing road‑video datasets with diverse tones and controllable captions. Additional examples can be found in \cref{ssec:caption_generation_appendix}. 



\noindent \textbf{Ablations:}
\textit{Single-Cap:} Training with single tone caption per video reduces Overall score by 2.7\% compared to \textsc{RoadTones-VL w/o CoT}, that is trained with multiple distinct tone captions. Thus, per‑video tone diversity during training enhances both controllability and factual stability at inference.
\textit{Single‑Run:} Removing best‑of‑n selection and TAS/FC gating during data construction lowers Overall score by 0.7\%, validating the utility of stage‑wise multiple runs and best caption selection for tone adherence and factual quality.
\textit{RoadSocial-Cap vs RoadTones-Cap:} Training on raw social captions of RoadSocial~\cite{roadsocial} (with extracted tone profiles) underperforms training on our pipeline‑generated tone‑controlled captions by 6.7\% Overall. Refined tone‑controlled targets are more effective than noisy social captions for learning controllability and reducing hallucinations.
These ablations highlight that diverse tonal supervision per video, quality training targets, and domain‑specific knowledge are critical to achieving strong tone controllability and factual consistency in caption generation-even when starting from strong VLM backbones.

\section{User Study}
\label{sec:user_study}

To assess the real-world effectiveness of our tone-controlled caption generator \CircledText[inner color=black, outer color=black, fill color=white]{TC-Gen} and automated evaluation metrics, we conducted a user study approved by an Institutional Review Board (IRB). Thirty-four participants (19 male, 15 female; age 21-45, mean 25.1) evaluated 84 tone-controlled captions across 28 videos from our dataset using a custom-built web interface. The study consisted of tasks designed to measure caption quality and tone controllability.

\noindent \textbf{\CircledText[inner color=black, outer color=black, fill color=white]{TC-Gen} Caption Quality Assessment:} Participants rated each caption on a 5-point Likert scale (1 = lowest, 5 = highest) across the following aspects: \textit{Tone Alignment:} Agreement between the caption’s tone and the specified personality and writing style, \textit{Tonal Relevance:} Appropriateness of the tone for the video’s context and events, \textit{Factual Consistency:} Accuracy of the caption relative to the depicted road event, \textit{Usefulness:} Practical value for applications such as safety awareness or driver behavior feedback, \textit{Human-likeness:} Perceived naturalness of the caption’s phrasing and expression.

\noindent \textbf{Agreement on RoadTones-Eval Metrics:} Participants rated captions produced by \textsc{RoadTones‑VL-CoT} on Tone Alignment and Factual Consistency. We then correlated human ratings with our automated metrics-Tone Alignment Score (TAS) and Factual Consistency (FC) (\cref{sec:tone_eval}).



\noindent\textbf{Tone Controllability Evaluation:} Participants evaluated whether \CircledText[inner color=black, outer color=black, fill color=white]{TC-Gen} could adjust tonal attribute intensity while preserving factual accuracy. For each video, two captions were shown-one with default tone settings and another with a modified intensity for a single tonal attribute (e.g., \textit{Sarcasm: Low} $\rightarrow$ \textit{High}). Users indicated whether change in tone was perceptible (\textit{Yes/No}) and whether  factual content remained consistent (\textit{Yes/No}). Agreement rates for  tone controllability and factual consistency were computed across participants.

\begin{table}[t!]
\centering
\resizebox{0.88\columnwidth}{!}{
\begin{tabular}{lcc}
\toprule
\textbf{Aspect} & \textbf{Average} & \textbf{p-value}\\
\midrule
Tone Alignment & 4.06 $\pm$ 0.78 & $1.2\times10^{-6}$\\
Factual Consistency  & 4.44 $\pm$ 0.63 & $8.6\times10^{-7}$\\
Human-Likeness  & 4.01 $\pm$ 0.87 & $8.6\times10^{-7}$\\
Tone-Relevance  & 3.71 $\pm$ 0.85 & $8.3\times10^{-7}$\\
Usefulness  & 3.52 $\pm$ 0.90 & $8.6\times10^{-7}$\\

\bottomrule
\end{tabular}
}
\caption{\textbf{User study results for \CircledText[inner color=black, outer color=black, fill color=white]{TC-Gen} Caption Quality Assessment } with 5-point Likert scale (higher = better).}
\label{tab:task1_results}
\end{table}

\noindent\textbf{Analysis of Results:} The average scores from the \textit{Caption Quality Assessment} (Table~\ref{tab:task1_results}) indicate that captions generated by our pipeline were rated highly across all dimensions. A one-sample Wilcoxon signed-rank test confirmed that all mean ratings were significantly above the neutral midpoint of~3 ($p<.001$). There was substantial agreement on automated evaluation metrics (Spearman $\rho = 0.723$ for TAS, $\rho = 0.773$ for FC). In the \textit{Tone Controllability Evaluation}, binomial tests showed strong agreement for both tone controllability ($89.71\%$, $p=2.5\times10^{-8}$) and factual consistency ($82.10\%$, $p=1.2\times10^{-14}$ ), validating the reliability of our tone-control mechanism ($p <.001$). Overall, participants found \CircledText[inner color=black, outer color=black, fill color=white]{TC-Gen} effective, demonstrating that our pipeline enhances factual precision and tone reliability.

%% file: 10_conclusion.tex
\section{Conclusion}
\label{sec:conclusion}

This work introduces a unified dataset-model-evaluation framework for tone-controllable road-video captioning. The proposed RoadTones stack comprises the \textbf{RoadTones-51K} dataset built through a scalable, human-validated data generation pipeline, the \textbf{\textsc{RoadTones-VL-CoT}} model for tone-aware video-to-text generation, and the \textbf{RoadTones-Eval} suite for standardized benchmarking. Our experiments and user studies show that tone can be infused effectively without compromising factual accuracy. Together, these contributions establish the groundwork for tone-controllable video-to-text generation. Our dataset is derived from social-media captions in English and inherits cultural, demographic, and platform biases. Cross-lingual and cross-cultural tone modeling is an interesting area for further work. 


\textbf{Acknowledgement.} This project was supported by the iHubData and Mobility at IIIT Hyderabad.




%% file: 12_appendix.tex
\section{Dataset Creation}
\label{sec:data_creation_appendix}

\subsection{\CircledText[inner color=black, outer color=black, fill color=white]{\textbf{TC-Gen}} Tone‑Controlled Caption Generation }
\label{ssec:caption_generation_appendix}

\textbf{Prompt schema and stage templates.} We provide the exact Stage-\Circled{1} and Stage-\CircledBlue{2} structured queries as JSON-like prompts (see \cref{fig:tc-gen_prompt1,fig:tc-gen_prompt2}). 
Each stage prompt contains five parts: system prompt, task instruction, contextual inputs (e.g., target tone specification, video summary), tone schema (how to interpret control axes and intensities), and rules or constraints for caption generation (e.g., keep content consistent with the neutral video summary). Target tone controls are passed as key-value pairs with continuous attribute intensities in [0,1] for Personality and Writing Style, and scalar/binary values for structural controls (informativeness, word count, viewpoint, hashtags, emojis, user mentions, location, date/time).

\CircledText[inner color=black, outer color=black, fill color=white]{\textbf{TC-Gen}} \textbf{configuration.} For both stages we use the same LLM \cite{openai2025gpt4.1} and decoding settings: temperature = 0.7, top\_p = 1.0, max\_tokens = 2048. Each stage generates n = 2 candidates by re‑prompting the LLM with the same controls. For every candidate, the tone profile is extracted using the Tone Extractor (\CircledText[inner color=white, outer color=darkgray, fill color=darkgray]{TX}) and the caption quality is computed with the Caption Evaluator (\CircledText[inner color=white, outer color=darkgray, fill color=darkgray]{CE}). Each stage returns: (i) the best candidate caption, (ii) its tone specification from \CircledText[inner color=white, outer color=darkgray, fill color=darkgray]{TX}, and (iii) its TAS/FC/Overall scores. 
Note that our \textbf{RoadTones-51K} dataset stores the tone specifications of the generated captions (extracted via \CircledText[inner color=white, outer color=darkgray, fill color=darkgray]{TX}) and not the original target controls, which are used only to drive generation and for internal comparison.

\begin{table*}[!t]
\centering
\renewcommand{\arraystretch}{1.3}
\resizebox{\linewidth}{!}{
\begin{tabular}{c|l|ccc|cccc|g|g|g}
\toprule

\rowcolor{white}
\multicolumn{1}{c}{\multirow{2}{*}{\#}} &
\multicolumn{1}{l}{\multirow{2}{*}{Generation Method}} &
\multicolumn{3}{c}{Narrative Alignment} &
\multicolumn{4}{c}{Structural Alignment} &
\multicolumn{1}{g}{\multirow{2}{*}{TAS}} &
\multicolumn{1}{g}{\multirow{2}{*}{FC}} &
\multicolumn{1}{g}{\multirow{2}{*}{Overall}} \\
\cmidrule(rl){3-5} \cmidrule(rl){6-9}

\rowcolor{white}
\multicolumn{1}{c}{} &
\multicolumn{1}{l}{} &
\multicolumn{1}{c}{P} & \multicolumn{1}{c}{WS} & \multicolumn{1}{c}{NAS} &
\multicolumn{1}{c}{A} & \multicolumn{1}{c}{I} & \multicolumn{1}{c}{wc} & \multicolumn{1}{c}{SAS} &
\multicolumn{1}{g}{} &
\multicolumn{1}{g}{} &
\multicolumn{1}{g}{} \\[1.2pt]

\cmidrule(r){1-1} \cmidrule(r){2-2} \cmidrule(rl){3-5} \cmidrule(rl){6-9} \cmidrule(rl){10-10} \cmidrule(rl){11-11} \cmidrule(l){12-12}


\rowcolor{PasteGreen} \textbf{Ours} & \CircledText[inner color=black, outer color=black, fill color=white]{\textbf{TC-Gen}}
& 77.3 & 85.0 & 86.2
& 96.8 & 84.9 & 91.3 & 91.0
& \textbf{86.2} & \textbf{99.0} & \textbf{92.5} \\

A & Order reversal
& 72.7 & 78.6 & 80.0
& 97.2 & 85.4 & 91.4 & 91.2
& \underline{82.7} & 99.0 & \underline{90.9} \\

B & Single-stage
& 70.3 & 72.9 & 71.6
& 99.2 & 85.8 & 93.7 & 92.9
& 82.3 & 98.9 & 90.6 \\

C & Writing Style-only
& 58.9 & 78.3 & 68.6
& 97.6 & 80.0 & 94.0 & 90.5
& 79.3 & 99.5 & 89.4 \\

D & Personality-only
& 67.6 & 67.3 & 67.4
& 98.8 & 81.5 & 84.9 & 88.4
& 78.0 & \underline{99.8} & 88.9 \\

\bottomrule
\end{tabular}
}
\caption{\textbf{Ablation of infusion and refinement strategies for \CircledText[inner color=black, outer color=black, fill color=white]{\textbf{TC-Gen}}.} We analyze the impact of multi-stage refinement and its ordering on tone-controlled caption generation. We report the alignment of target tone controls with that of the generated captions. Tone alignment abbreviations - Personality traits (P), Writing Style (WS), Structural Attributes (A), Informativeness (I), word count (wc), Narrative Alignment Score (NAS), Structural Alignment Score (SAS), Factual Consistency (FC), and Tone-controlled caption quality (Overall).
All scores are percentages (\%).
 }
\label{tab:tcgen_ablations}
\end{table*}

\textbf{Ablations (\cref{tab:tcgen_ablations}).} We next dissect the design choices of \CircledText[inner color=black, outer color=black, fill color=white]{\textbf{TC-Gen}} through targeted ablations, isolating the impact of stage ordering, multi‑stage refinement, and selective tone infusion on tone alignment and factual consistency.
\begin{itemize}
    \item Order reversal (row A, \cref{tab:tcgen_ablations}): Swap the stages- first infuse Personality + Structure, then refine Writing Style controls, to test sensitivity to control ordering.
    \item Single‑stage (B): Infuse all tone controls in one pass (no refinement) to quantify the gain from multi‑stage calibration.
    \item Writing Style‑only (C): Apply only Writing Style + Structural controls (no Personality) to isolate the contribution of writing style.
    \item Personality‑only (D): Apply only Personality + Structural controls (no Writing Style) to isolate the contribution of persona.
\end{itemize}
All ablations and the reference \CircledText[inner color=black, outer color=black, fill color=white]{\textbf{TC-Gen}} pipeline use the same configuration and the same set of target tone controls derived via the nearest‑neighbor selection procedure (\cref{fig:distinct-tone-captions-per-video}, Section \cref{sec:data_creation} in the main paper). The captions generated by \CircledText[inner color=black, outer color=black, fill color=white]{\textbf{TC-Gen}} and ablations are evaluated using Tone Alignment (TAS) and Factual Consistency (FC) scores. \hlpastegreen{The results reported in }(\cref{tab:tcgen_ablations})\hlpastegreen{ demonstrate that }\CircledText[inner color=black, outer color=black, fill color=white]{\textbf{TC-Gen}}\hlpastegreen{ consistently produces captions with the highest alignment to the target tone controls (TAS), outperforming all ablations.}


\begin{figure}
    \centering
    \includegraphics[width=1\linewidth]{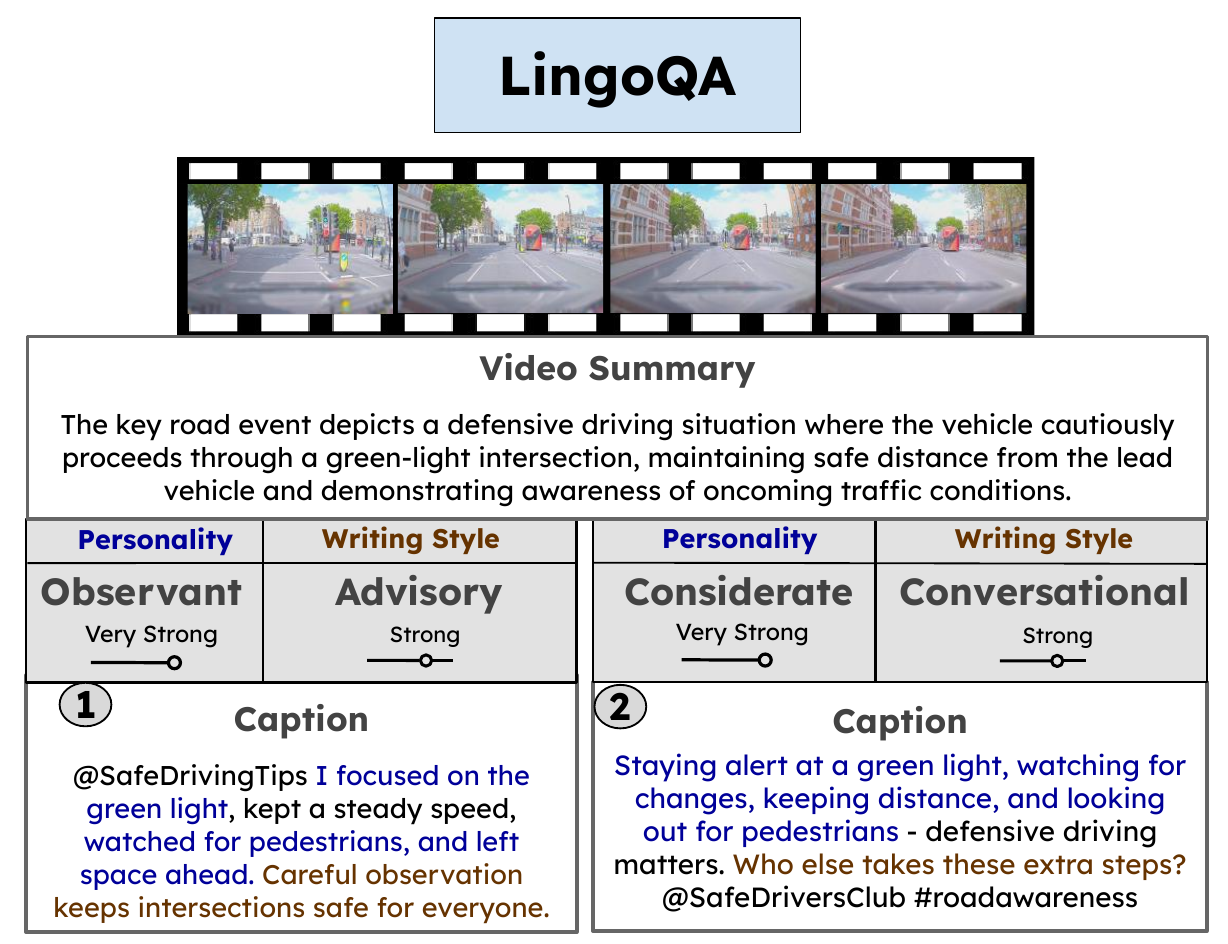}
    \caption{\textbf{Applicability of our tone-controlled caption generation pipeline on popular road video dataset: LingoQA\cite{LingoQA}}. For each video sample, we show two distinct tone captions (\CircledText[inner color=black, outer color=black, fill color=gray!35]{1},\CircledText[inner color=black, outer color=black, fill color=gray!35]{2}) with corresponding dominant \textcolor{darkblue}{Personality} and \textcolor{darkbrown}{Writing Style} attributes highlighted in text. Video summary is shown for reference.
    }
    \label{fig:LingoQA_cap1}
\end{figure}

\begin{figure}
    \centering
    \includegraphics[width=1\linewidth]{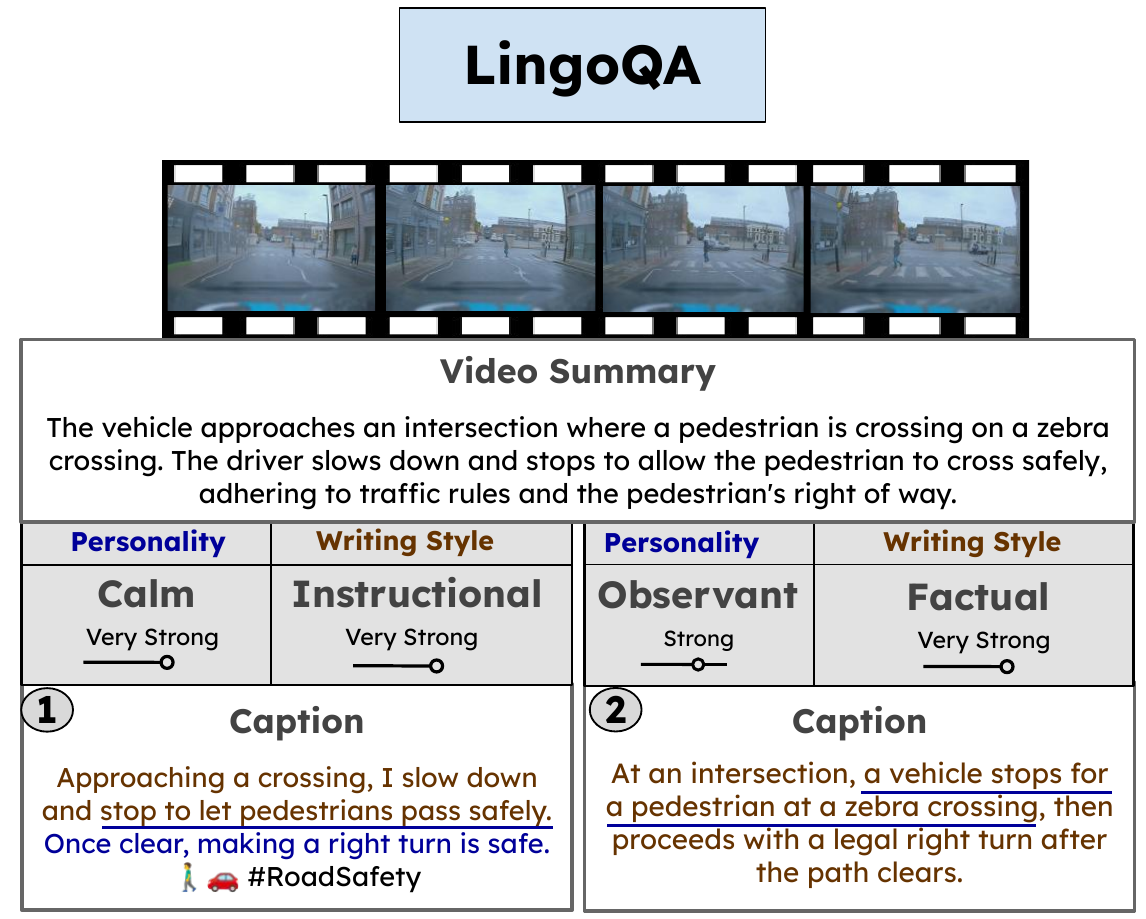}
    \caption{\textbf{Applicability of our tone-controlled caption generation pipeline on popular road video dataset: LingoQA\cite{LingoQA}}. For each video sample, we show two distinct tone captions (\CircledText[inner color=black, outer color=black, fill color=gray!35]{1},\CircledText[inner color=black, outer color=black, fill color=gray!35]{2}) with corresponding dominant \textcolor{darkblue}{Personality} and \textcolor{darkbrown}{Writing Style} attributes highlighted in text. Video summary is shown for reference. 
    }
    \label{fig:LingoQA_cap2}
\end{figure}

\begin{figure}
    \centering
    \includegraphics[width=1\linewidth]{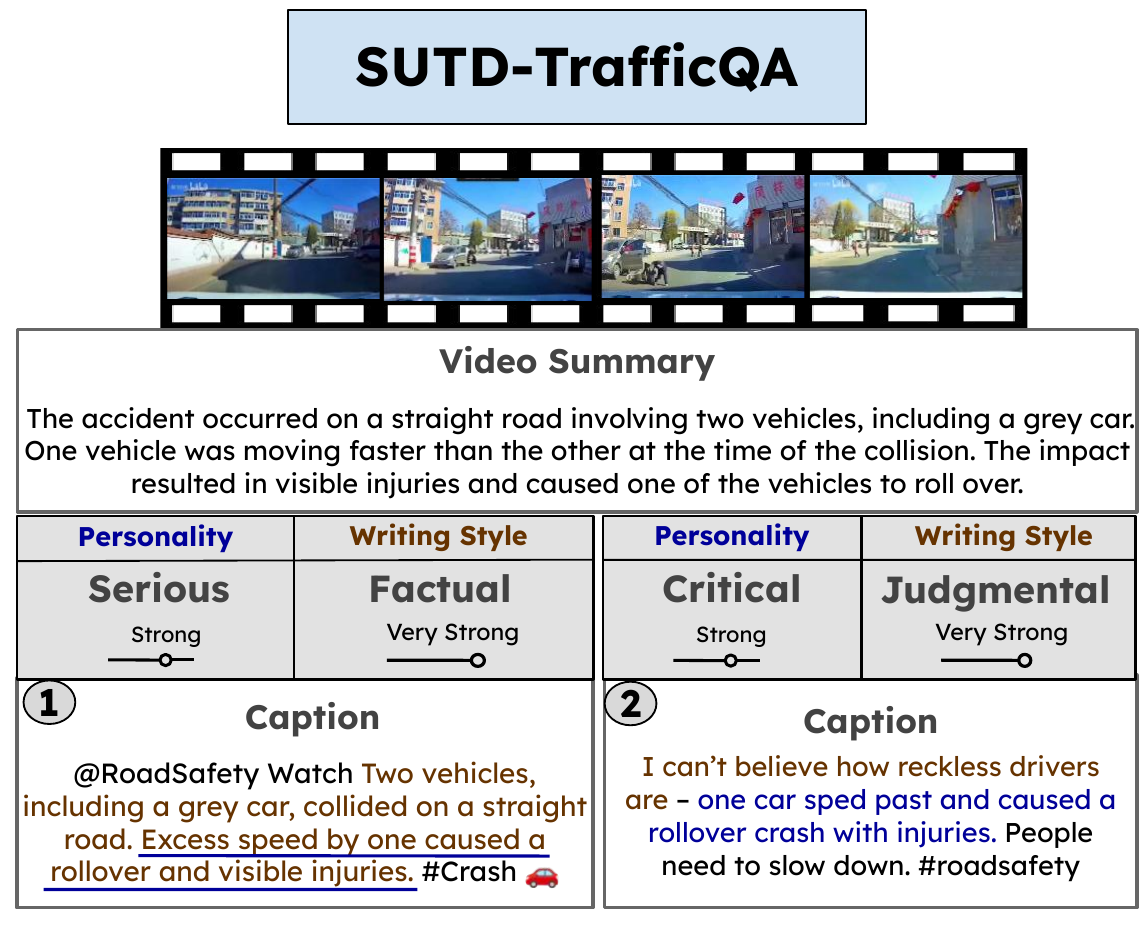}
    \caption{\textbf{Applicability of our tone-controlled caption generation pipeline on popular road video dataset: SUTD-TrafficQA\cite{SUTD-TrafficQA}}. For each video sample, we show two distinct tone captions (\CircledText[inner color=black, outer color=black, fill color=gray!35]{1},\CircledText[inner color=black, outer color=black, fill color=gray!35]{2}) with corresponding dominant \textcolor{darkblue}{Personality} and \textcolor{darkbrown}{Writing Style} attributes highlighted in text. Video summary is shown for reference. 
    }
    \label{fig:SUTD_cap1}
\end{figure}

\begin{figure}
    \centering
    \includegraphics[width=1\linewidth]{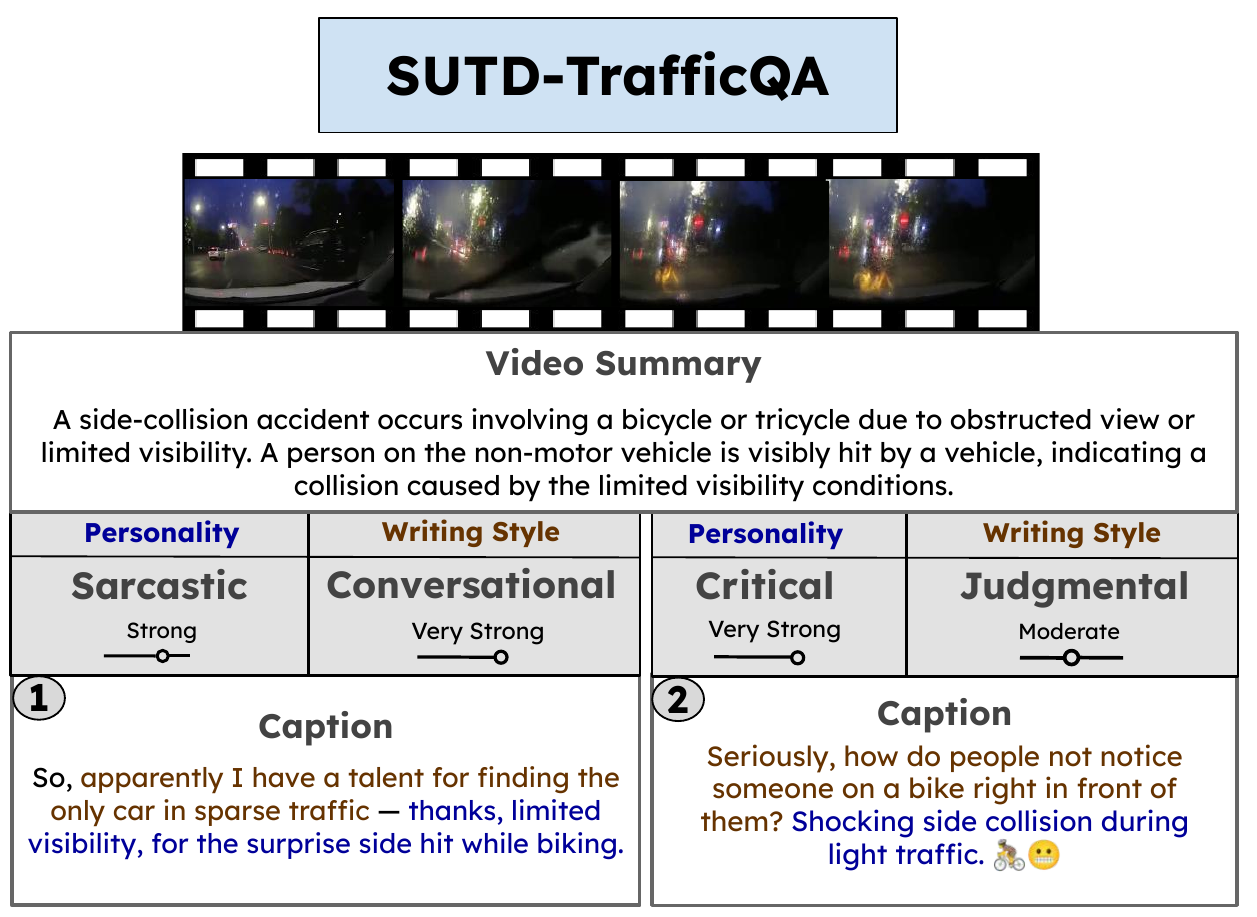}
    \caption{\textbf{Applicability of our tone-controlled caption generation pipeline on popular road video dataset: SUTD-TrafficQA\cite{SUTD-TrafficQA}}. For each video sample, we show two distinct tone captions (\CircledText[inner color=black, outer color=black, fill color=gray!35]{1},\CircledText[inner color=black, outer color=black, fill color=gray!35]{2}) with corresponding dominant \textcolor{darkblue}{Personality} and \textcolor{darkbrown}{Writing Style} attributes highlighted in text. Video summary is shown for reference.
    }
    \label{fig:SUTD_cap2}
\end{figure}

\CircledText[inner color=black, outer color=black, fill color=white]{\textbf{TC-Gen}} \textbf{features.} To demonstrate fine‑grained tone control via \CircledText[inner color=black, outer color=black, fill color=white]{\textbf{TC-Gen}}, we include single-attribute perturbation examples as shown in (Figs. \ref{fig:caption-after-one-change1},\ref{fig:caption-after-one-change2}, and \ref{fig:caption-after-one-change3}). Starting from a base tone profile, we vary exactly one control at a time (e.g., increase ``Caring''  Writing Style intensity from Absent to Moderate) while holding all other controls fixed, and regenerate the caption. For each case, we present the modified target tone profile and the updated caption, with the phrases or markers reflecting the altered attribute highlighted. Additionally, \hlpastegreen{to demonstrate the utility and portability of our pipeline (nearest neighbor distinct tones selection + }\CircledText[inner color=black, outer color=black, fill color=white]{\textbf{TC‑Gen}})\hlpastegreen{, we apply it to popular road‑video datasets- LingoQA} \cite{LingoQA}\hlpastegreen{, SUTD‑TrafficQA }\cite{SUTD-TrafficQA}\hlpastegreen{, and RoadSocial }\cite{roadsocial}\hlpastegreen{, and present distinct tone captions for sample clips from each }(Figs. \ref{fig:LingoQA_cap1}, \ref{fig:LingoQA_cap2}, \ref{fig:SUTD_cap1} and \ref{fig:SUTD_cap2}).

\subsection{\CircledText[inner color=white, outer color=darkgray, fill color=darkgray]{TX} Tone Extractor}
\label{ssec:tone_extract_appendix}

\textbf{Writing Style Inventory refinement:} We initialize the inventory with commonly used writing styles: factual, conversational, and instructional. Subsequently, for each tone-aware caption, we extract the attribute intensities using the Tone Extractor pipeline. If all existing writing style attributes get low intensity scores (i.e., no prominent style), we prompt the LLM~\cite{openai2025gpt4.1} to propose a new candidate that strongly captures the caption’s expressive form. Proposed styles are added to the inventory only if manual review deems them relevant and non‑redundant. We repeat this procedure for all captions in RoadSocial~\cite{roadsocial} and iteratively update the inventory with new attributes resulting in a total of 16 unique writing styles. A schema describing each style with example captions reflecting the style are provided in \cref{fig:writing_style_schema}.

\textbf{Tone Extraction Prompts.}
We provide the prompts to extract the tone content of a caption along the dimensions: Writing Style (\cref{fig:writing_style_tx_prompt}), Personality (\cref{fig:personality_tx_prompt}), Informativeness (\cref{fig:informativeness_tx_prompt}), and Structural attributes (\cref{fig:structural_attr_prompt}). For Writing Style, we follow the schema in \cref{fig:writing_style_schema} to return continuous intensities in [0,1] for the 16 style attributes. All tone‑extraction calls use GPT‑4.1‑mini \cite{openai2025gpt4.1} with temperature 0.4.

\begin{figure}
    \centering
    \includegraphics[width=1\linewidth]{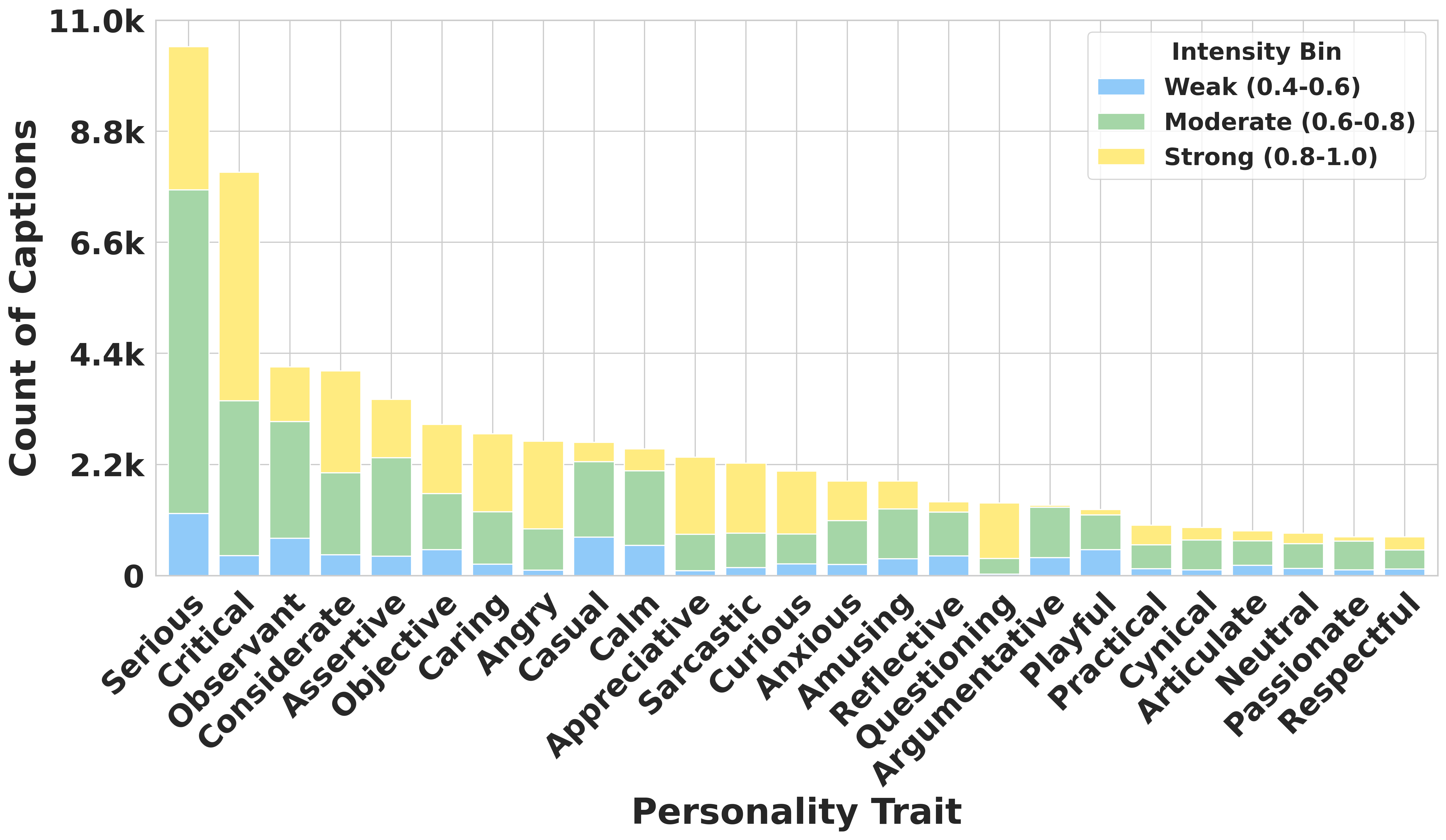}
    \caption{\textbf{Top 25 Personality Traits Intensity Distribution in RoadTones-51K.} The chart visualizes the total instances of 25 most frequent personality traits segmented into three intensity bins: \textit{Weak} (0.4-0.6), \textit{Moderate} (0.6-0.8), and \textit{Strong} (0.8-1.0). Traits with intensity level less than 0.4 are not considered for tone-controllable captioning.}
    \label{fig:personality-intensity-dist}
\end{figure}

\begin{figure}
    \centering
    \includegraphics[width=1\linewidth]{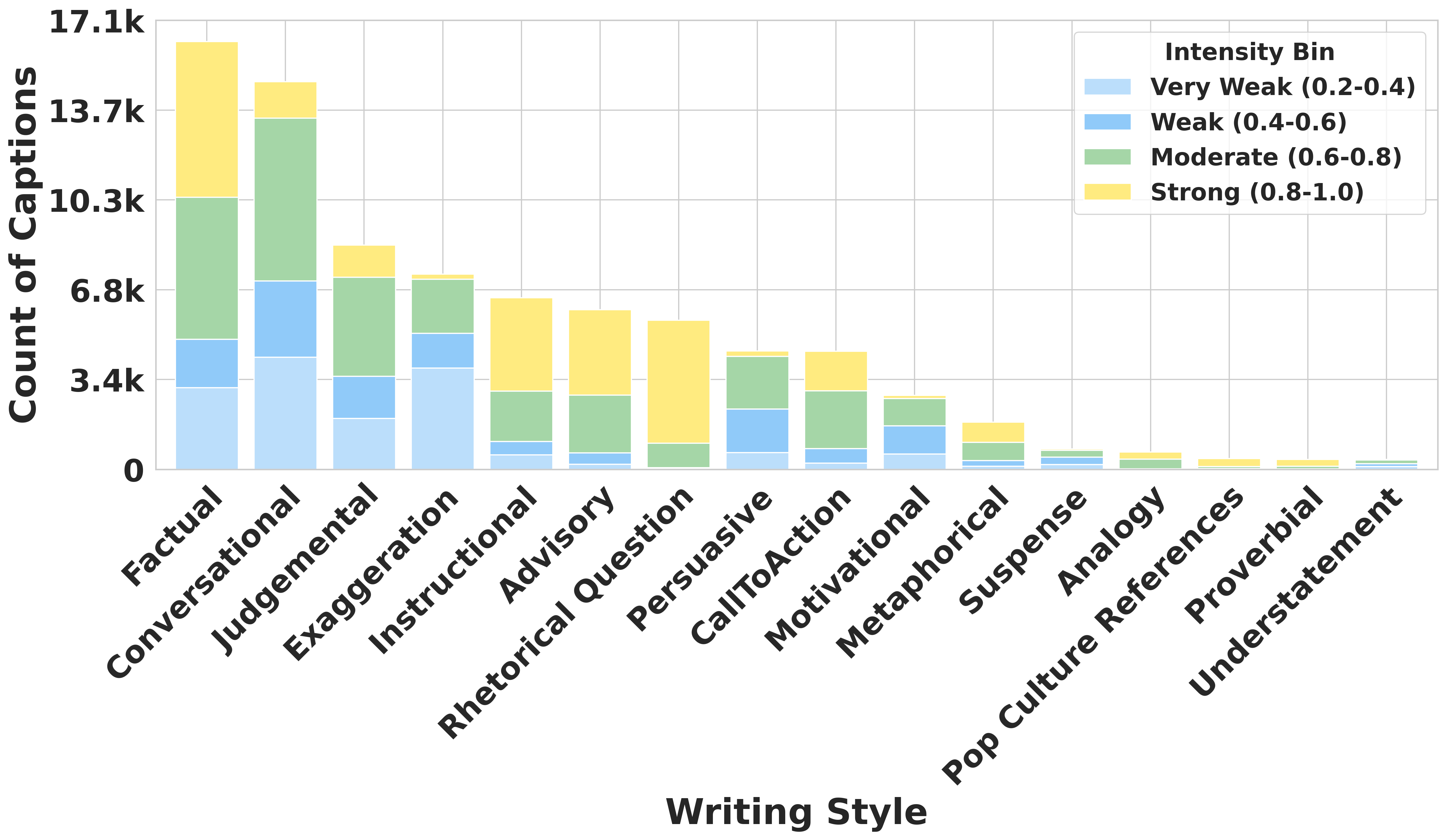}
    \caption{\textbf{Intensity Distribution of Writing Styles in RoadTones-51K.} The chart visualizes the total instances of 16 writing styles segmented into four intensity bins: \textit{Very Weak} (0.2-0.4), \textit{Weak} (0.4-0.6), \textit{Moderate} (0.6-0.8), and \textit{Strong} (0.8-1.0). Attributes with intensity level less than 0.2 are not considered for tone-controllable captioning.}
    \label{fig:writing-style-intensity-dist}
\end{figure}

\begin{figure}
    \centering
    \includegraphics[width=1\linewidth]{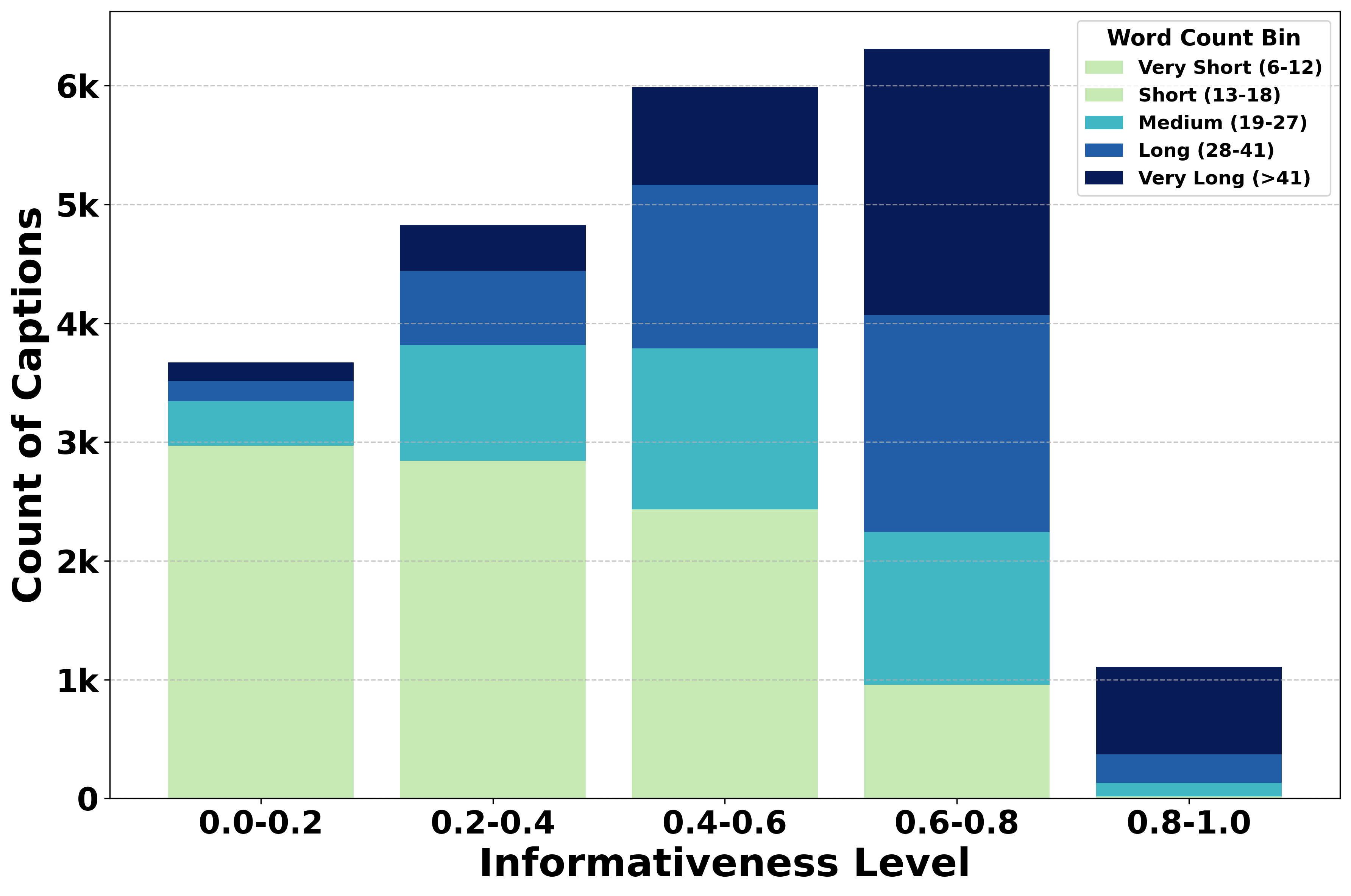
    }
    \caption{\textbf{Word Count Correlation with Informativeness Level in RoadTones-51K.} The distribution shows a clear relationship: captions with a low informativeness (0.0-0.4) are predominantly \textit{Very Short} or \textit{Short} (low word count; $<18$), while those with high informativeness (0.6-1.0) are strongly associated with \textit{Long} and \textit{Very Long} caption lengths (high word count; $>30$).}
    \label{fig:informativeness-vs-wordcount}
\end{figure}

\begin{figure}
    \centering
    \includegraphics[width=1\linewidth]{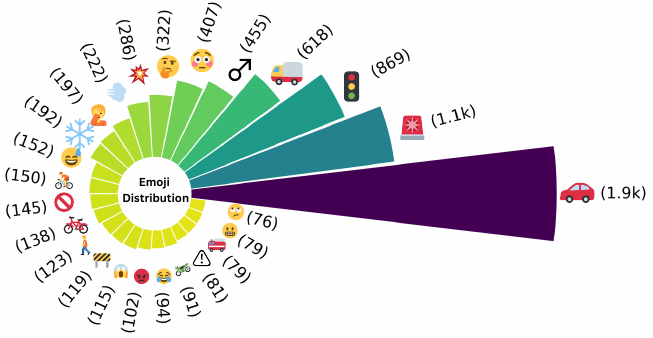}
    \caption{\textbf{Distribution of Top 25 Emojis in RoadTones-51K.} This chart visualizes the usage frequency of the most common emojis found in RoadTones-51K captions.}
    \label{fig:emoji-dist}
\end{figure}

\begin{figure*}[h!]
    \centering
    \includegraphics[width=1\textwidth]{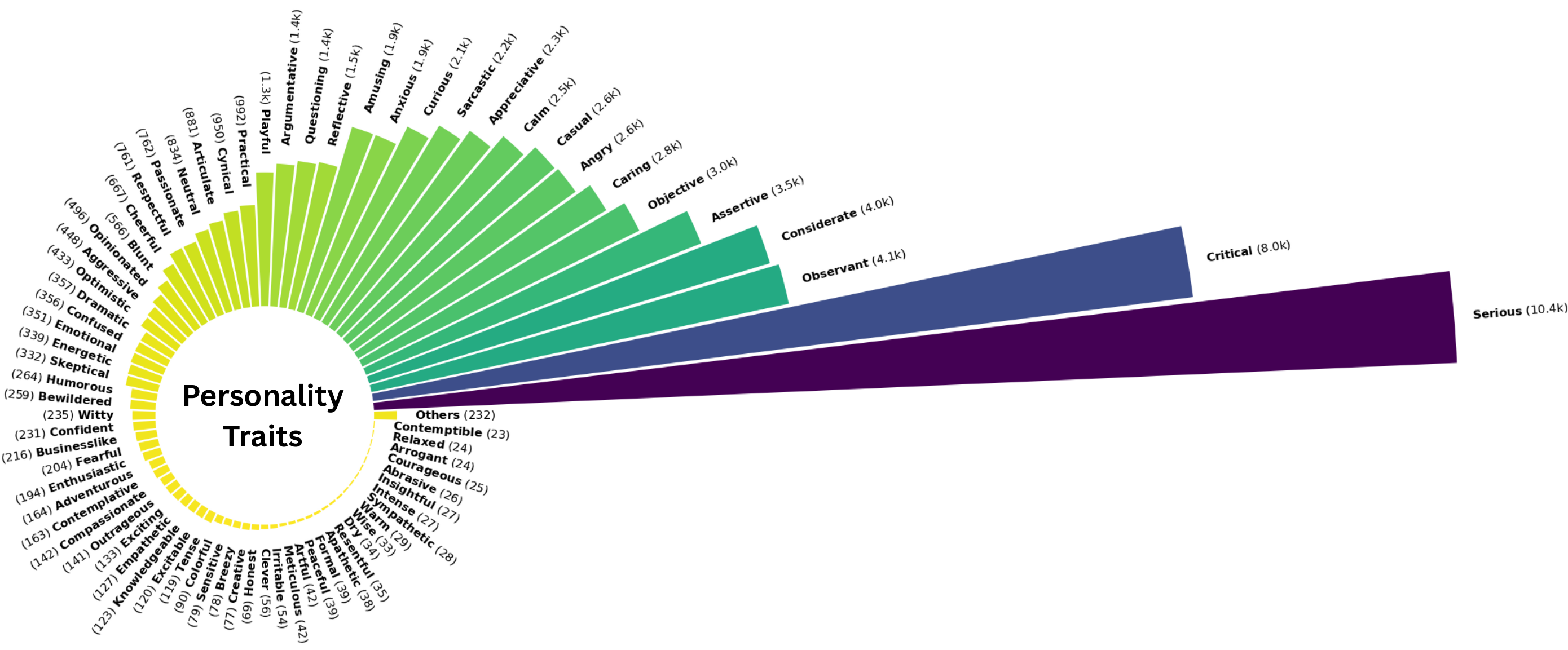}
    \caption{\textbf{Personality Trait Distribution.} This figure displays the distribution of 75 most frequent personality traits in \textbf{RoadTones-51K}. Less frequent traits are aggregated into the ``Others'' category.}
    \label{fig:personality_top75}
\end{figure*}

\subsection{Data Statistics} 
\label{ssec:tone_stats_appendix}

We report additional statistics for \textbf{RoadTones-51K}  (see \cref{fig:personality_top75,fig:personality-intensity-dist,fig:writing-style-intensity-dist,fig:informativeness-vs-wordcount,fig:emoji-dist}). \cref{fig:personality_top75} shows a long‑tailed distribution of dominant Personality trait annotations in the 51K captions; \cref{fig:personality-intensity-dist} further breaks down the top‑25 traits by intensity coverage (weak/moderate/strong). \cref{fig:writing-style-intensity-dist} summarizes the 16 Writing Styles and their intensity distributions. 
\cref{fig:informativeness-vs-wordcount} plots Informativeness levels against word‑count bins, revealing that low informativeness aligns with short captions, while high informativeness correlates with longer ones. Finally, \cref{fig:emoji-dist} presents the emoji usage distribution, dominated by domain‑relevant vehicles, traffic, and alert symbols.

\begin{figure*}
    \centering
    \includegraphics[width=1\linewidth]{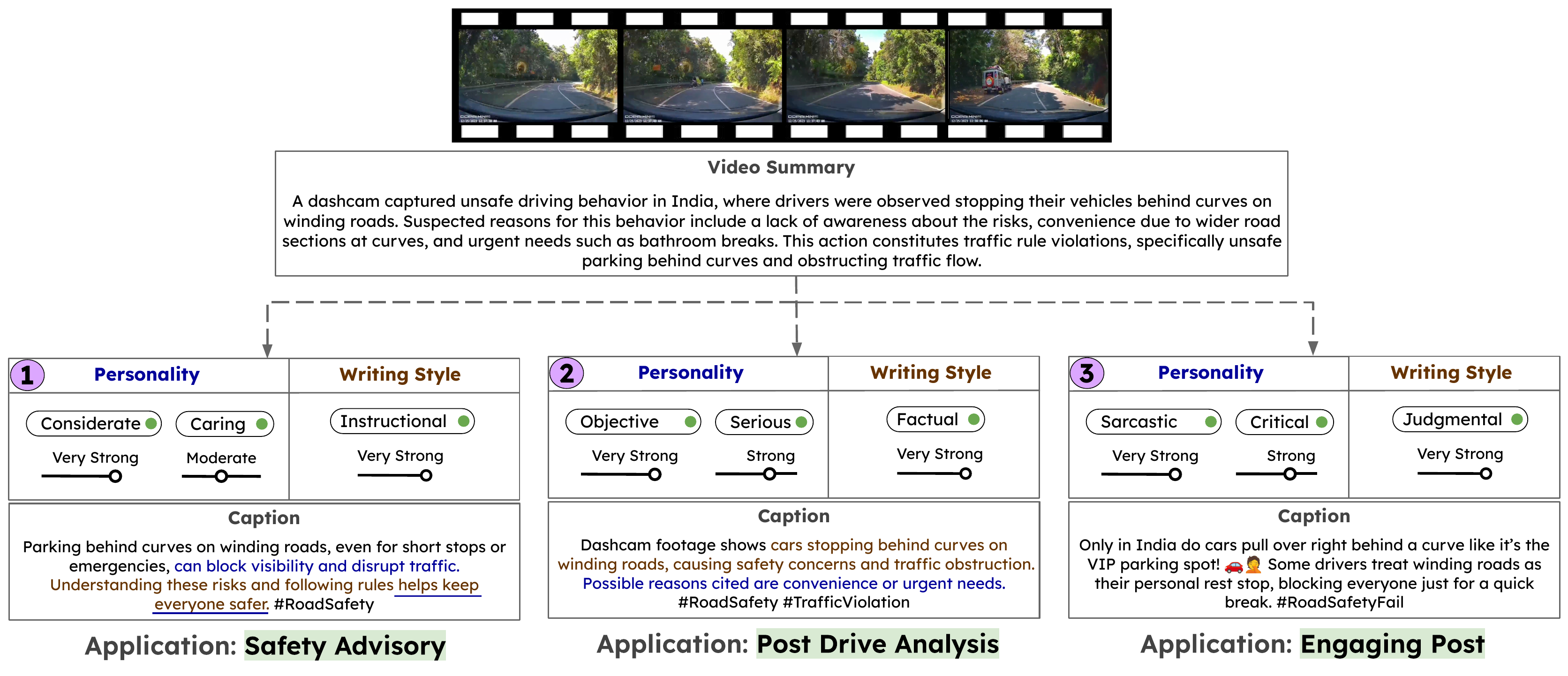}
    \caption{\textbf{Representative samples from RoadTones-51K with potential usecase/applications.} \CircledText[inner color=black, outer color=black, fill color=white]{TC-Gen}'s tone-controlled captions can be used in diverse domains, such as issuing \CircledText[inner color=black, outer color=black, fill color=lightlavender]{\textbf{1}} \textit{Safety Advisories}, conducting \CircledText[inner color=black, outer color=black, fill color=lightlavender]{\textbf{2}} \textit{Post-Drive Analysis}, or creating \CircledText[inner color=black, outer color=black, fill color=lightlavender]{\textbf{3}} \textit{Engaging Posts} for social media.} 
    \label{fig:intro_applications}
\end{figure*}

\textbf{Samples from RoadTones-51K.} 
We present representative samples with potential use cases for tone‑controlled captions (\cref{fig:intro_applications}), highlighting the intended audience/application to illustrate how the same factual event can be reframed for different stakeholders. 

\textbf{Tone's utility for downstream applications:} To demonstrate utility, we evaluate tone-conditioned retrieval and tone-based clustering of rare events on held-out datasets (nexar-collision, roadwork). As shown in \cref{tab:tone_retrieve_cluster} below, higher mAP scores confirm that tone embeddings help retrieve rare or safety-critical scenarios effectively, highlighting tone’s usefulness beyond stylistic variation.
\begin{table}[!h]
    \centering
    \resizebox{\linewidth}{!} {
    \begin{tabular}{ccccc}
        \hline
        \textbf{Dataset} & mAP@5 & mAP@all & ACC & NMI  \\
        \hline
        nexar-collision\cite{nexar}  & 0.986 & 0.852 & 0.75 & 0.341 \\
        \hline
        ROADWork \cite{roadwork}  & 0.983 & 0.884 & 0.847 & 0.504 \\
        \hline
    \end{tabular}
    }
    \caption{Tone-conditioned video retrieval and tone-based clustering.}
    \label{tab:tone_retrieve_cluster}
\end{table}

\begin{table*}[!t]
\centering
\begin{minipage}{0.95\textwidth}
\begin{tcolorbox}[enhanced, colback=gray!5, colframe=gray!60!black, title={\textbf{Instruction templates for Tone-Controlled Road Video Captioning}}]
\begin{itemize}
\item ``Give a tone-controlled caption of the primary traffic event unfolding in the scene."
    \item ``Please describe the key road event observed in this driving video as a tone-aware description."
    \item ``What is the key traffic event observed in this video? Respond with a tone-controlled description."
    \item ``Generate a tone-conditioned description of the main road event shown in the video."
    \item ``Briefly explain the central traffic event in this driving scenario with a tone-conditioned caption."
    \item ``What specific road event is taking place in this video? Provide a tone-conditioned description."
    \item ``Provide a tone-controlled natural-language description of the key road or traffic event."
    \item ``Describe the key road maneuver or traffic event occurring in this footage with a tone-controlled description."
    \item ``Write a tone-controlled caption that summarizes the key road event."
    \item ``What is the most notable road event or change captured in the video? Respond with a tone-controlled caption or description."
    \item ``Describe the main activity or incident occurring on the road with a tone-conditioned description."
    \item ``Based on the video, what is the main traffic event being presented? Provide a tone-aware caption."
    \item ``Summarize the primary road event depicted in the driving clip with a tone-controlled caption."
    \item ``Explain the key event occurring in this driving scenario with a tone-conditioned explanation."
    \item ``Give a concise, tone-controlled narrative of the primary road incident shown in this video segment."
    \item ``Provide a coherent, tone-controlled description of the key road event illustrated in the video."
\end{itemize}
\end{tcolorbox}
\begin{tcolorbox}[enhanced, colback=cyan!7, colframe=cyan!50!black, title={\textbf{Appended to each instruction template (binding rules)}}]
\small
Apply the provided tone/style and structural controls spec as binding rules. Interpret Personality and Writing Style attribute intensity values (0-0.2: Absence, 0.2-0.4: Subtle presence, 0.4-0.6: Moderate presence, 0.6-0.8: Strong presence, 0.8-1.0: Very Strong presence) as degree controls; interpret Informativeness as the desired amount of factual detail using levels (0-0.2: Negligible, 0.2-0.4: Minimal, 0.4-0.6: Moderate, 0.6-0.8: High, 0.8-1.0: Extensive); honor yes/no toggles in Factual Attributes; and match the exact word\_count. Spec: \texttt{\{0\}}
\end{tcolorbox}
\begin{tcolorbox}[enhanced, colback=violet!7, colframe=violet!50!black, title={\textbf{(Optional) CoT instruction appended to each template}}]
\small
Additionally, provide a step-by-step reasoning of how you arrived at the final tone-controlled caption.
\end{tcolorbox}
\end{minipage}
\caption{Structured Instruction templates used to fine-tune and benchmark MLLMs for Tone-controlled Road Video Captioning. The rule box is appended to every instruction template for interpreting the tonal and structural attribute control values. Floating-point intensity values ranging from 0 to 1 are used for Tonal attributes, Informativeness, and provided as input to the rule box in place of \texttt{\{0\}}. An additional CoT instruction is also provided for reference.}
\label{tab:tonedesc_instructions}
\end{table*}

\section{VLM for Tone-Controllable Captioning}
\label{sec:vlm_appendix}

\textbf{Structured query template.}
We use a single structured query format to  fine-tune \textsc{RoadTones-VL} (with and without CoT) for tone-controlled road video captioning (\cref{tab:tonedesc_instructions}). Each query is constructed as follows:
\begin{itemize}
    \item Randomly sample one instruction template from the top block of \cref{tab:tonedesc_instructions} to reduce prompt overfitting.
    \item Append the binding-rules block that explains how to interpret tone control axes (Personality, Writing Style, Structural controls) and their levels.
    \item Serialize the target tone specification into the Spec placeholder \{0\} as a key-value block with floating point intensities in [0,1] and binary toggles including word count.
    \item Append the Chain-of-Thought instruction shown in third box of \cref{tab:tonedesc_instructions}, if CoT output is requested, otherwise omit it.
\end{itemize}
A sample CoT input–output instruction-tuning triplets is provided in  (\cref{tab:sample_input_output_instruction_triplet}). The model’s response separates the final caption and rationale using tags: \textit{[FINAL]...[/FINAL][REASONING]...[/REASONING]}. Non-CoT targets contain only the final caption.

For the auxiliary video‑summarization objective, we provide separate instruction templates (\cref{tab:desc_instructions}). These contain no tone schema or controls; the instruction asks only for a neutral, detailed summary of the key road event in the video.

\section{\CircledText[inner color=black, outer color=black, fill color=paleyellow]{TE} Tone Evaluation Metrics}
\label{sec:tone_eval_appendix}

We provide the exact LLM-as-a-judge~\cite{llmjudge} prompts for computing the $NAS$ components: Personality similarity ($S_p$, \cref{fig:personality_llmeval_prompt}), Writing Style similarity ($S_w$, \cref{fig:writing_style_llmeval_prompt}), and Factual Consistency ($FC$, \cref{fig:factual_consistency_prompt}). All prompts output a single float score in [0,1]. We use GPT-4.1-mini \cite{openai2025gpt4.1} with deterministic decoding (temperature 0.0, top\_p 1.0, max\_tokens 256) and strict JSON parsing.

\section{Qualitative Results}
\label{sec:results_appendix}

We show some qualitative results \textsc{RoadTones-VL-CoT} against ground-truth (\cref{fig:roadtones_vl_captions-qualitative}) and against popular open and closed-source models (Figs. \ref{fig:compare_models_captions-eg1}, \ref{fig:compare_models_captions-eg2}, \ref{fig:compare_models_captions-eg3}, \ref{fig:compare_models_captions-eg4}, \ref{fig:compare_models_captions-eg5}, \ref{fig:compare_models_captions-eg6}, \ref{fig:compare_models_captions-eg7}, and \ref{fig:compare_models_captions-eg8}).

\begin{table*}[!t]
\centering
\begin{tabular}{c|l|ccc|cccc|c|c|c}
\toprule
\rowcolor{white} \multirow{2}{*}{\#} & \multirow{2}{*}{Model}  & \multicolumn{3}{c|}{Narrative Alignment} & \multicolumn{4}{c|}{Structural Alignment} & \multirow{2}{*}{TAS} & \multirow{2}{*}{FC} & \multirow{2}{*}{Overall}  \\
\cmidrule(rl){3-5} \cmidrule(rl){6-9}
\rowcolor{white} & & P & WS & NAS & A & I & wc & SAS & & & \\ [1.2pt]
\cmidrule(r){1-1} \cmidrule(r){2-2}  \cmidrule(rl){3-5} \cmidrule(rl){6-9} \cmidrule(rl){10-10} \cmidrule(rl){11-11} \cmidrule(l){12-12} 


\rowcolor{SkyBlue} \textbf{Ours}  & w similar-case & 72.7 &	81.6 &	77.2 &	98.1 &	74.2 &	94.1 &	88.8 &	83.0 &	57.2 & 70.1 \\

 & w/o similar-case  &  72.1  &	79.2 &	75.7 &	98.1 &	72.8 &	94.4 &	88.4	& 82.1	&56.4	& 69.3 \\

\midrule

Single-Cap &  n=1 & 70.9 & 79.5  & 75.2 & 98.4 & 69.5 & 92.1 & 86.6 & 80.9 & 53.0 & 67.0 \\

  &  n=2  & 72.5	& 80.7	&76.6	& 98.0 &	72.0	 & 91.6	& 87.7	&82.2&	56.1&	66.4 \\

\rowcolor{SkyBlue} \textbf{Ours}  & n=3  & 72.7 &	81.6 &	77.2 &	98.1 &	74.2 &	94.1 &	88.8 &	83.0 &	57.2 & 70.1 \\

&  n=4  & 72.1 & 80.4  & 76.3 & 98.1 & 72.7 & 94.6 & 88.5 & 82.4 & 57.5 & 70.0 \\
  
&  n=5 &  73.5 & 80.4 & 77.0 & 98.2 & 73.7 & 95.3 & 89.1 & 83.1  & 57.6 &  70.3\\

\bottomrule
\end{tabular}
\caption{Additional ablations with \textsc{RoadTones-VL} (\textbf{Ours}). 
}
\label{tab:roadtones_ablations}
\end{table*}

\section{Additional Ablations}

(1) \cref{tab:roadtones_ablations} compares our model v/s a variant fine-tuned without similar case retrieval (i.e., k-nearest neighbor data creation method described in Sec. 4 - Fig. 2, main paper). Our model outperforms variant by a noticeable margin (1.5 more points in NAS). (2)  \cref{tab:roadtones_ablations} additionally compares performances of models trained on TC-Gen generated data with varying numbers of retrieved cases (n=1 to 5). Our chosen n=3 obtains highest NAS (77.2) representing superior persona and style adherence compared to other choices. Factual Consistency (FC) saturates after n=3 with marginal gains. These  validate our retriever module effectiveness for diverse tone profiles, grounded in similar event contexts.

\section{User Study}
\label{sec:user_study_appendix}

\noindent\textbf{RoadTones familiarization.}
Before the main study, participants completed a brief familiarization to align their understanding of tone and evaluation criteria. 
Participants answered multiple‑choice practice questions covering three skills: (i) identifying the dominant narrative tone of a caption (Personality and Writing Style), (ii) detecting changes in tonal intensity (e.g., weak vs strong expression of an attribute), and (iii) assessing factual consistency of a caption with respect to the road video and its neutral summary.
Each QA provided immediate feedback: the correct answer and a short rationale explaining the decision (e.g., why a trait was dominant, which lexical or structural cues indicated intensity shifts, or which facts aligned/contradicted the summary). Distractor options were designed to be clearly incorrect. \cref{fig:streamlit-ui-quiz} shows the familiarization interface and a supplementary video (\textit{RoadTones\_UserStudy\_familiarization.mp4}) shows the questionnaire overview for all tasks. This calibration step standardized participants’ mental models of tone, reduced ambiguity in subsequent judgments, and strengthened the reliability of the main user study. 

\begin{figure*}[h!]
    \centering
    \includegraphics[width=1\linewidth]{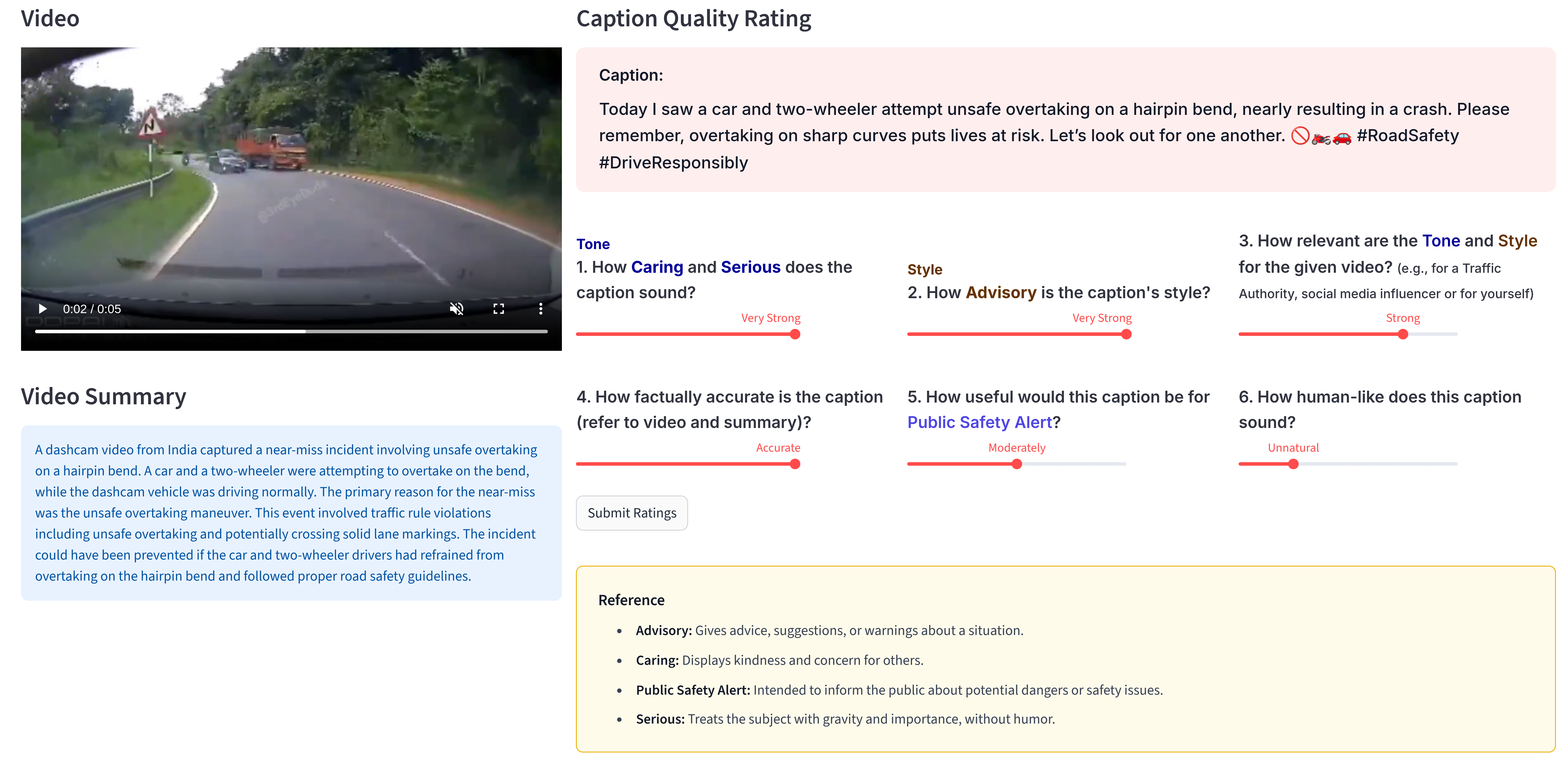}
    \caption{\textbf{User Study Interface for \CircledText[inner color=black, outer color=black, fill color=white]{\textbf{TC-Gen}} Caption Quality Assessment}. Participants viewed a video, its video summary and evaluated the quality of the corresponding caption generated by \CircledText[inner color=black, outer color=black, fill color=white]{\textbf{TC-Gen}} based on Tone Alignment, Tone Relevance, Factual Consistency, Usefulness and Human-Likeness on a 5-pt Likert Scale.}
    \label{fig:streamlit-ui-task1}
\end{figure*}

\noindent \CircledText[inner color=black, outer color=black, fill color=white]{\textbf{TC-Gen}} \textbf{Caption Quality Assessment.}  The users rated the quality of tone-controlled captions generated by \CircledText[inner color=black, outer color=black, fill color=white]{\textbf{TC-Gen}} pipeline (given their tone content extracted via \CircledText[inner color=white, outer color=darkgray, fill color=darkgray]{TX}) along five dimensions: Tone Alignment (Personality  and Writing Style), Tone Relevance, Factual Consistency, Usefullness, and Human-likeness as shown in \cref{fig:streamlit-ui-task1}.

\begin{figure*}[h!]
    \centering
    \includegraphics[width=1\linewidth]{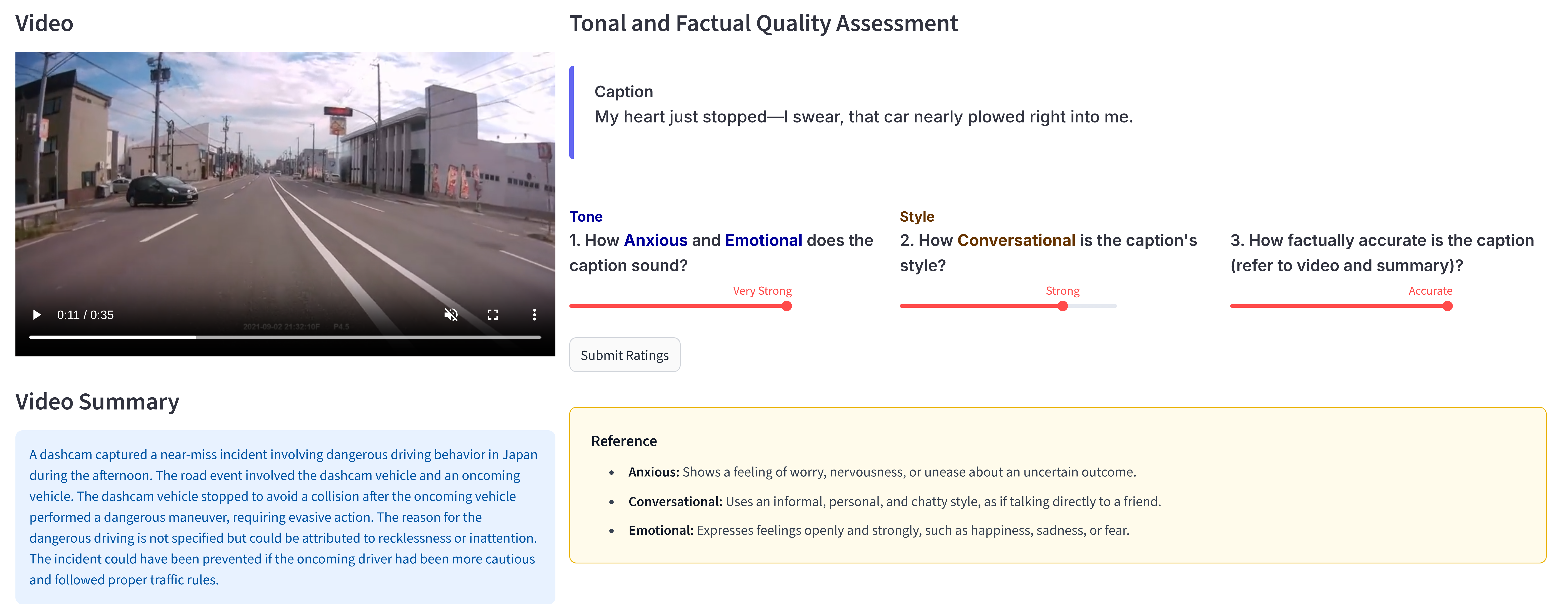}
    \caption{\textbf{User Study Interface for \textit{Agreement on RoadTones-Eval Metrics}}. Participants viewed a video, its video summary and rated the corresponding caption generated by \textsc{RoadTones-VL-CoT} based on Tone Alignment and Factual Consistency on a 5-pt Likert Scale. The user ratings were then correlated with scores computed by RoadTones-Eval metrics.}
    \label{fig:streamlit-ui-task2}
\end{figure*}

\noindent\textbf{Aggreement on RoadTones-Eval Metrics.} The users rated the quality of tone-controlled captions generated by \textsc{RoadTones-VL-CoT} model (given the ground-truth or target tone controls provided by the \textbf{RoadTones-51K} dataset) along two dimensions: Tone Alignment (Personality  and Writing Style) and Factual Consistency as shown in \cref{fig:streamlit-ui-task2}. This validates the alignment of user ratings with our proposed evaluation metrics.

\begin{figure*}[h!]
    \centering
    \includegraphics[width=1\linewidth]{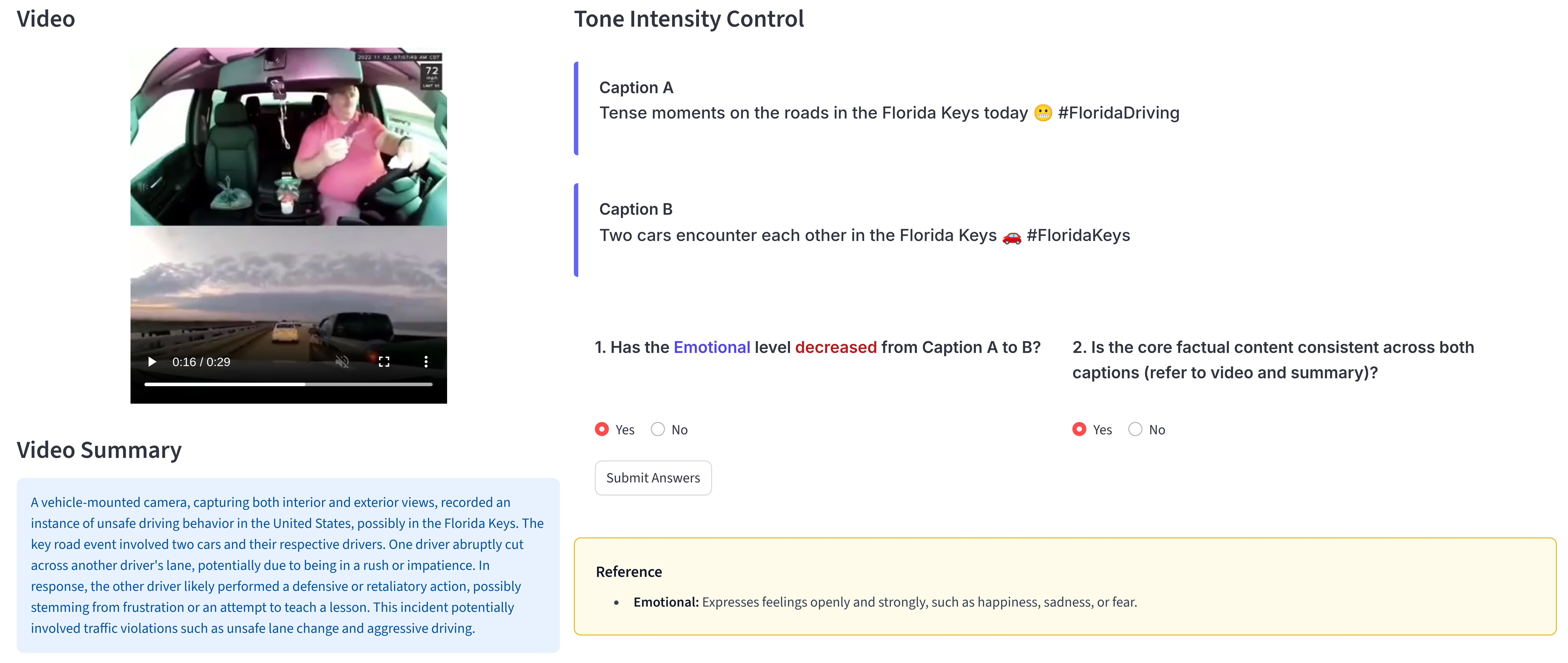}
    \caption{\textbf{User Study Interface for \textit{Tone Controllability Evaluation}}. Participants viewed a video, its video summary and evaluated Tone Controllability and Factual Consistency of the corresponding captions generated by \CircledText[inner color=black, outer color=black, fill color=white]{\textbf{TC-Gen}}.}
    \label{fig:streamlit-ui-task3}
\end{figure*}

\begin{figure*}
    \centering
    \includegraphics[width=0.9  \linewidth]{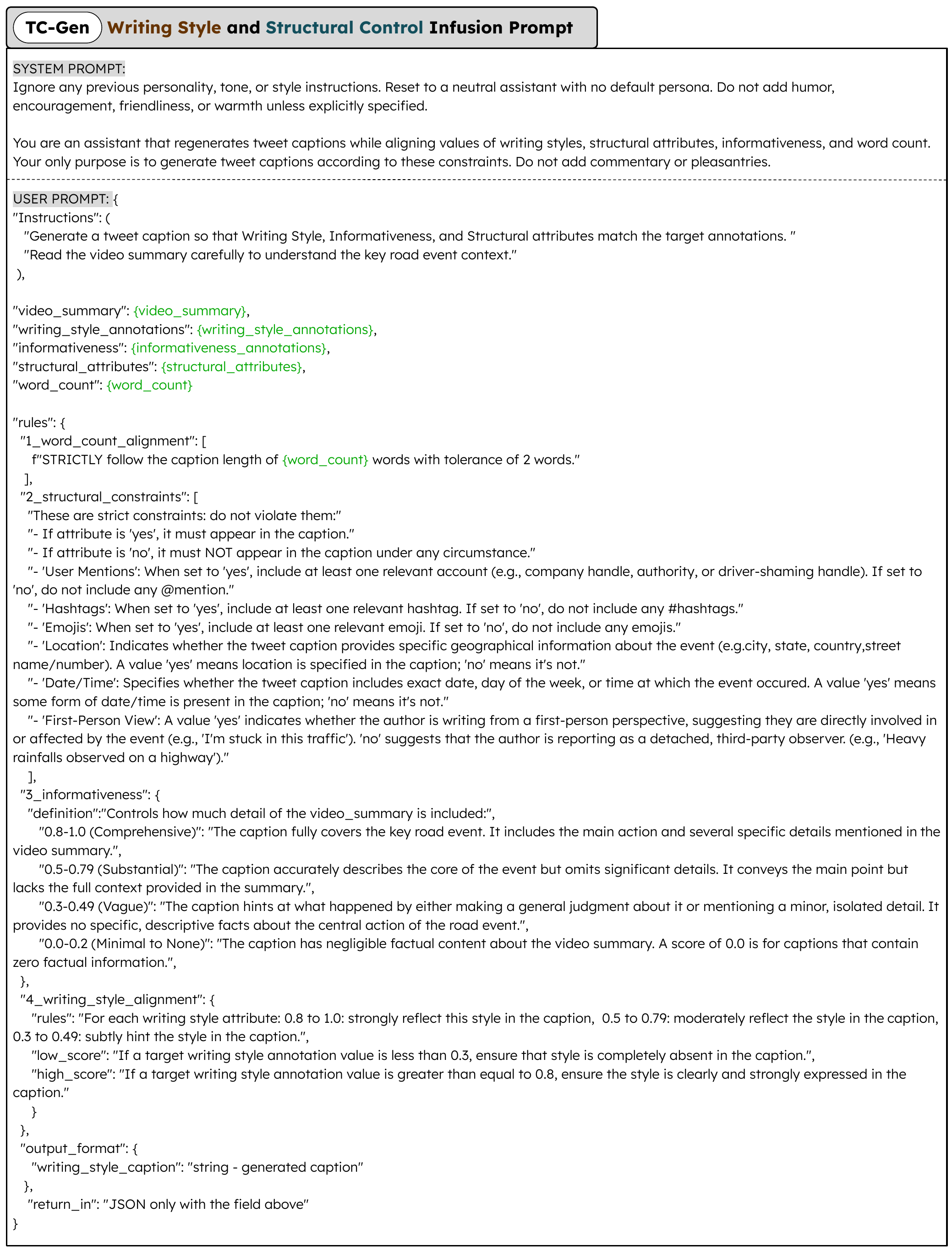}
    \caption{\textbf{\CircledText[inner color=black, outer color=black, fill color=white]{\textbf{TC-Gen}} Stage-\Circled{1} prompt.} }
    \label{fig:tc-gen_prompt1}
\end{figure*}

\begin{figure*}
    \centering
    \includegraphics[width=0.9  \linewidth]{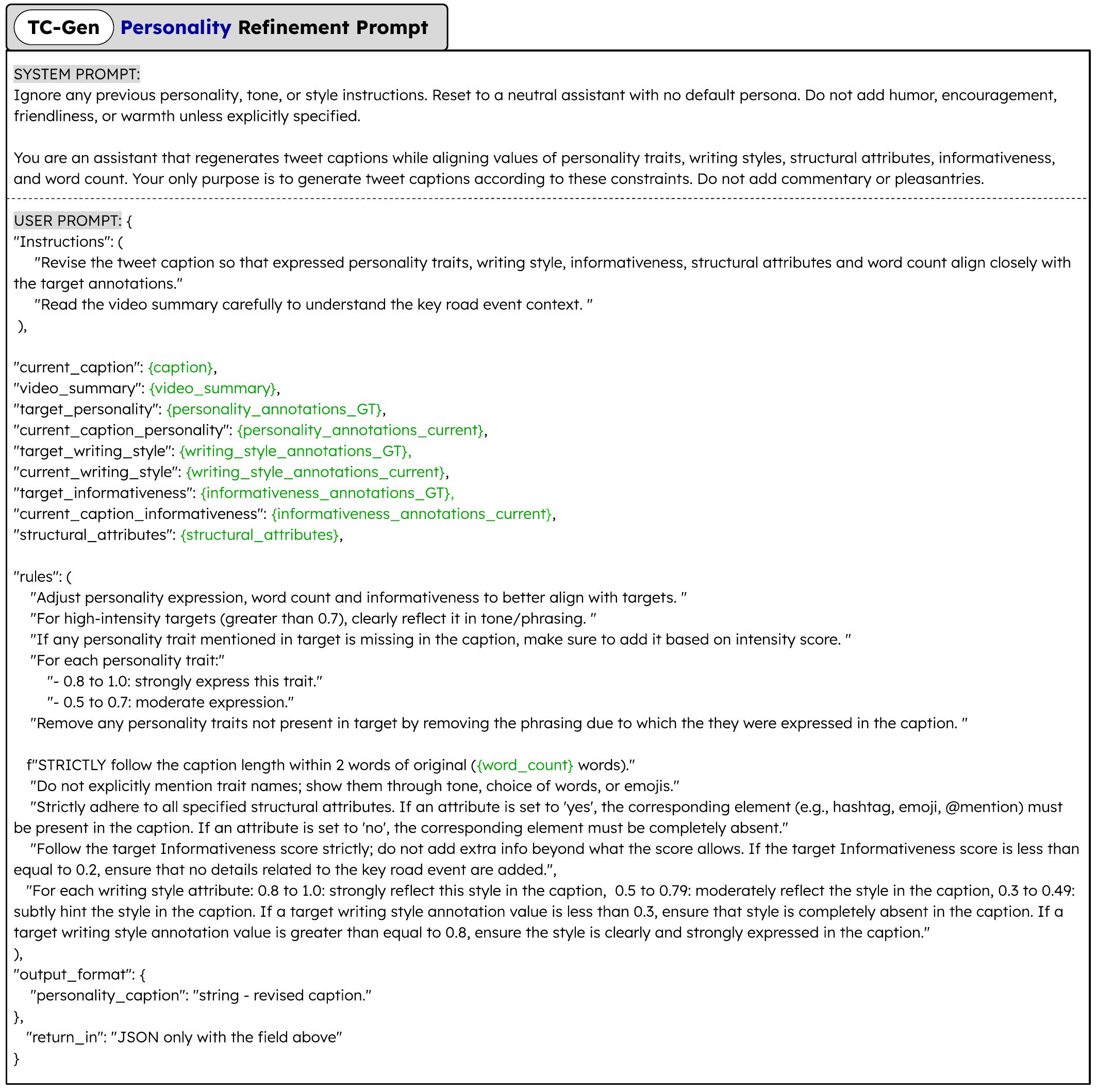}
    \caption{\textbf{\CircledText[inner color=black, outer color=black, fill color=white]{\textbf{TC-Gen}} Stage-\CircledBlue{2} Prompt.} }
    \label{fig:tc-gen_prompt2}
\end{figure*}

\begin{figure*}
    \centering
    \includegraphics[width=0.85\linewidth]{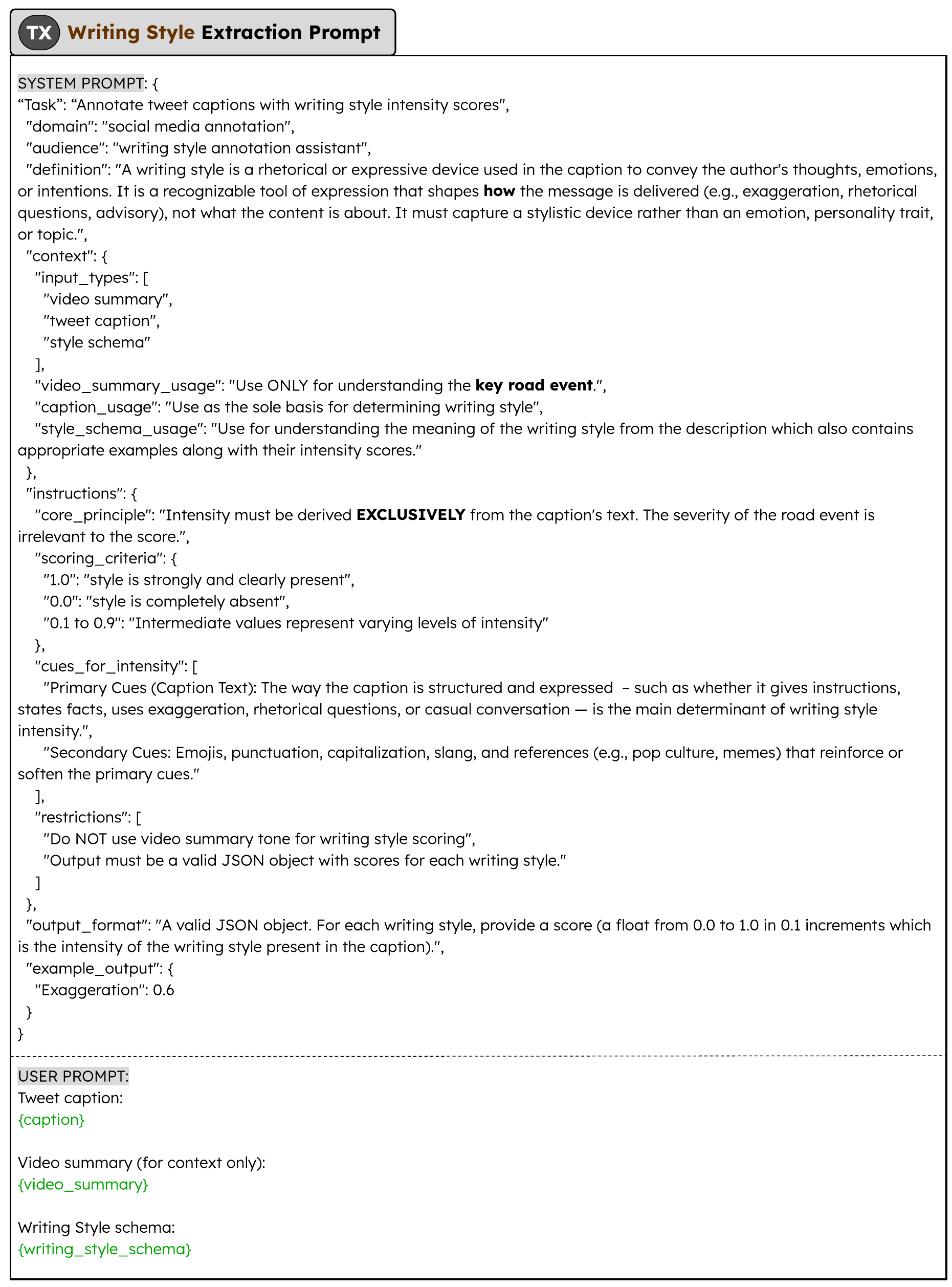}
    \caption{\textbf{Writing Style tone extraction prompt.} The prompt defines the task, context, scoring criteria, and restrictions provided to the LLM for writing style attributes intensity prediction based on caption text. Video summary about the key road event is also provided to disentangle the factual from the tonal content of the caption. }
    \label{fig:writing_style_tx_prompt}
\end{figure*}

\begin{figure*}
    \centering
    \includegraphics[width=0.9\linewidth]{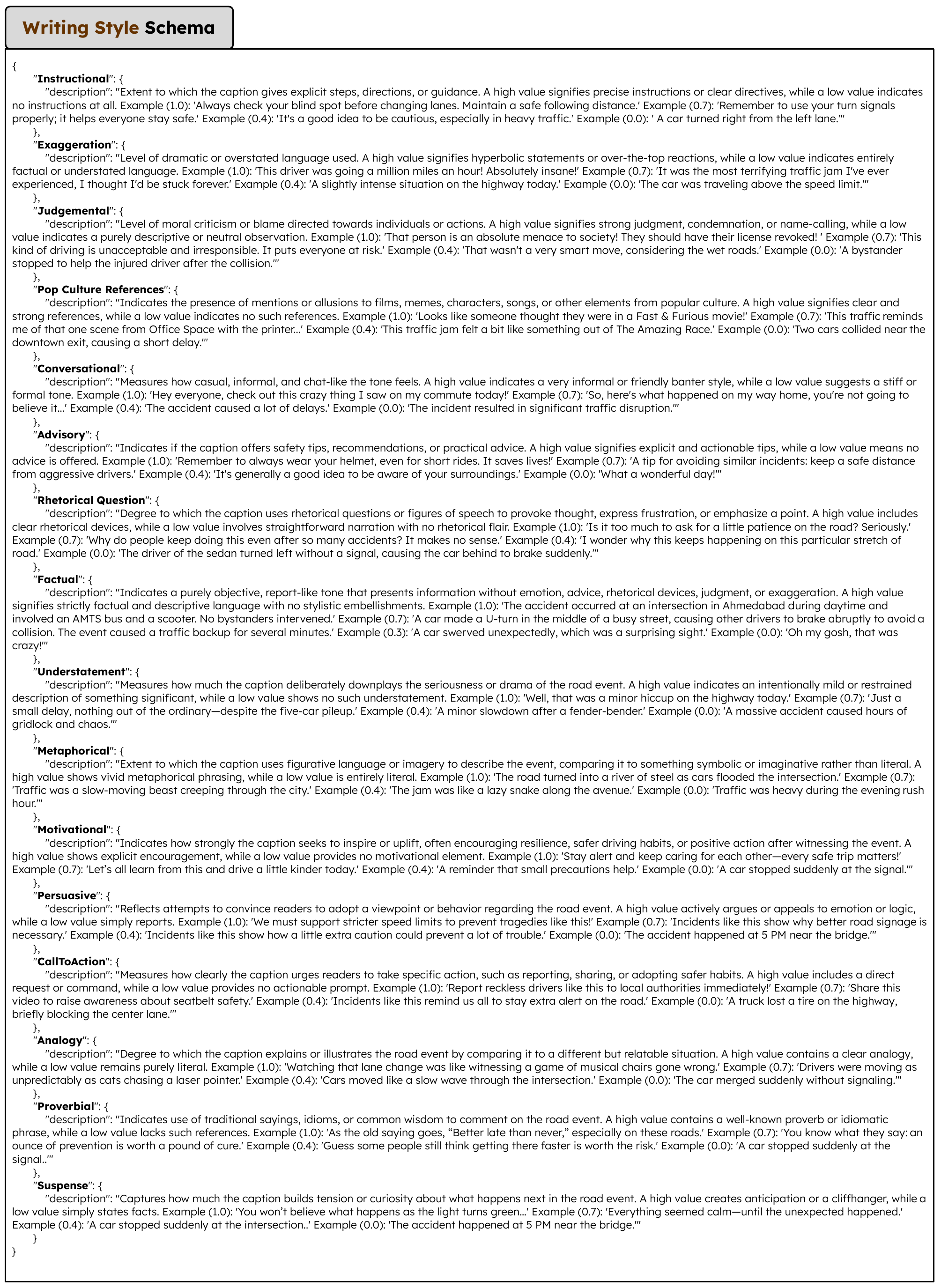}
    \caption{Writing Style tone schema defining the 16 attributes along with examples based on intensity levels.}
    \label{fig:writing_style_schema}
\end{figure*}

\begin{figure*}
    \centering
    \includegraphics[width=1\linewidth]{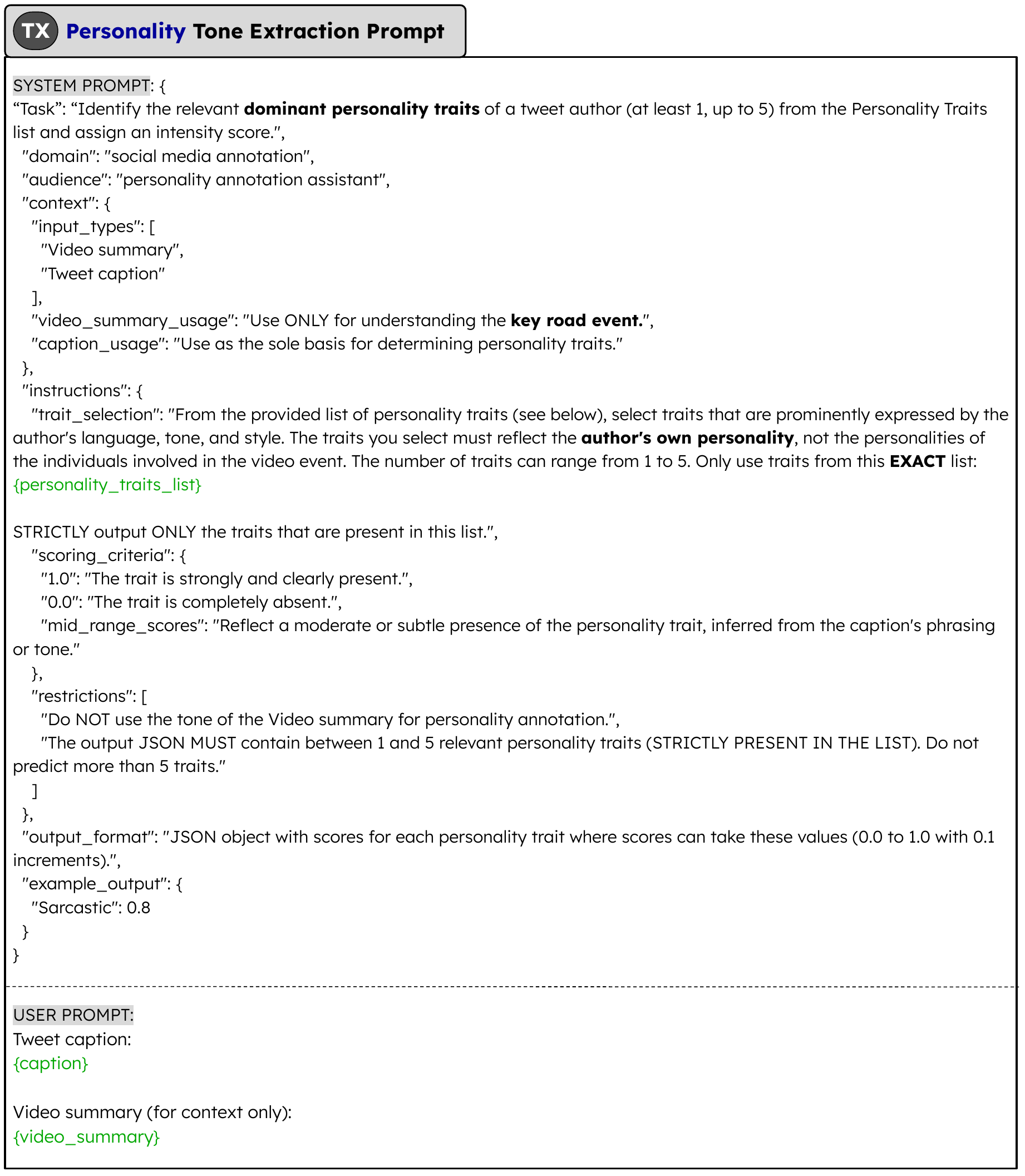}
    \caption{\textbf{Personality trait extraction prompt.} The prompt defines the task, context, scoring criteria, and restrictions provided to the LLM for personality traits intensity prediction based on caption text. Video summary about the key road event is also provided to disentangle the factual from the tonal content of the caption. }
    \label{fig:personality_tx_prompt}
\end{figure*}

\begin{figure*}
    \centering
    \includegraphics[width=1\linewidth]{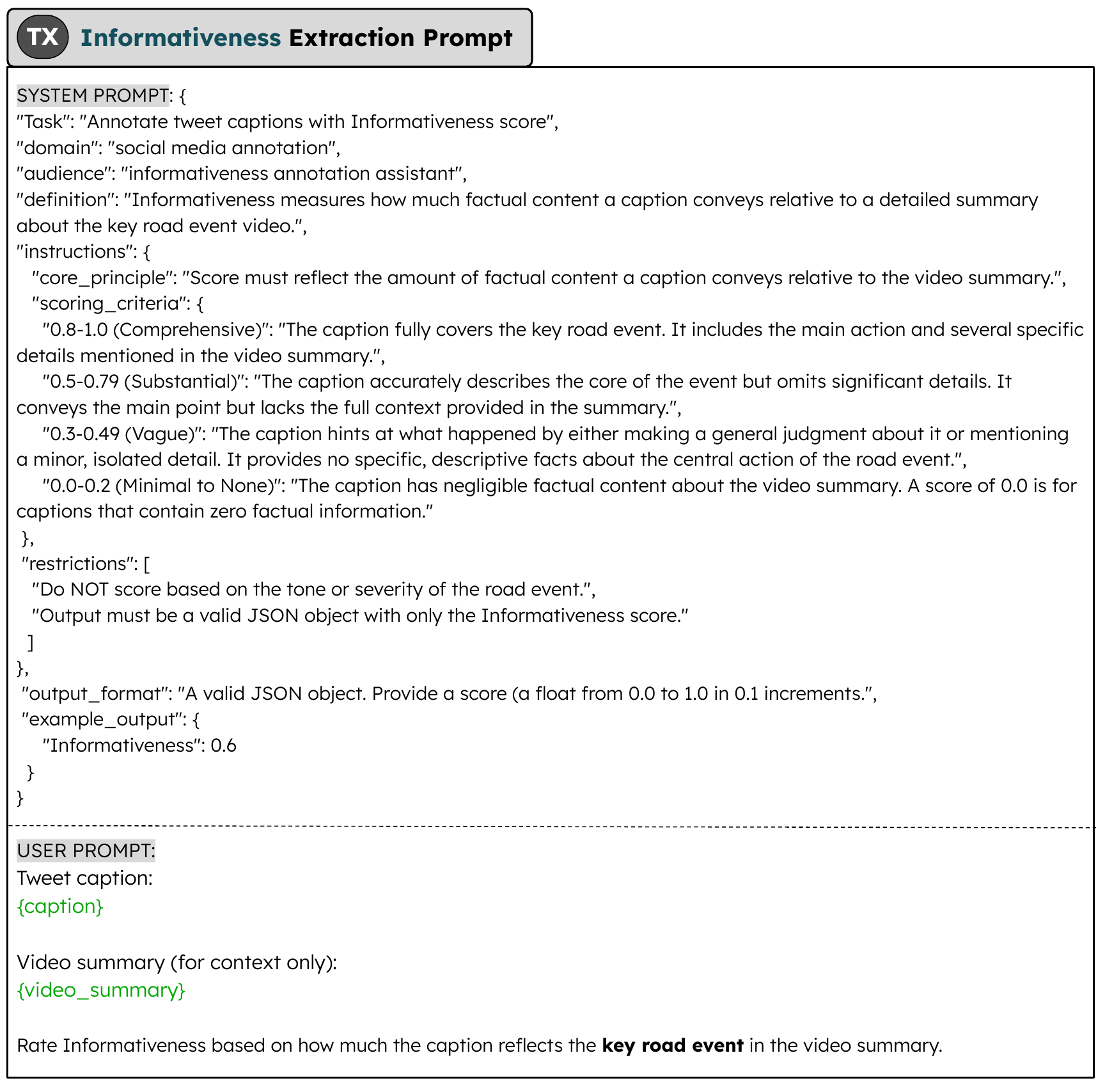}
    \caption{\textbf{Informativeness level extraction prompt.} The prompt defines the task, context, scoring criteria, and restrictions provided to the LLM for informativeness level prediction based on the amount of factual information conveyed through the caption relative to the detailed road video summary.}
    \label{fig:informativeness_tx_prompt}
\end{figure*}

\begin{figure*}
    \centering
    \includegraphics[width=1\linewidth]{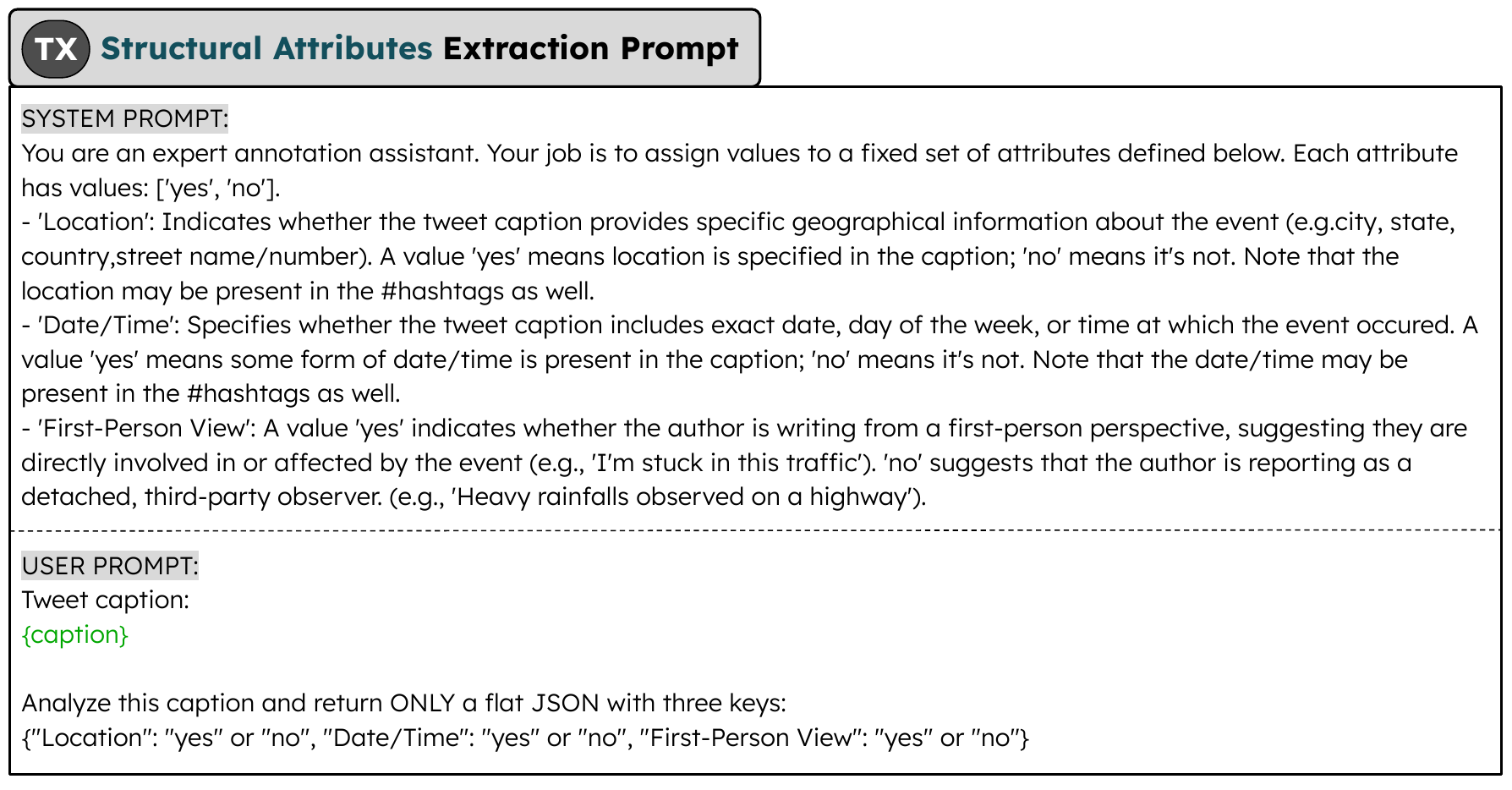}
    \caption{\textbf{Structural attributes extraction prompt.} The prompt guides the LLM to classify the presence ('yes' or 'no') of \textit{Location}, \textit{Date/Time}, and \textit{First-Person View} based on the provided definitions.}
    \label{fig:structural_attr_prompt}
\end{figure*}

\begin{figure*}
    \centering
    \includegraphics[width=1\linewidth]{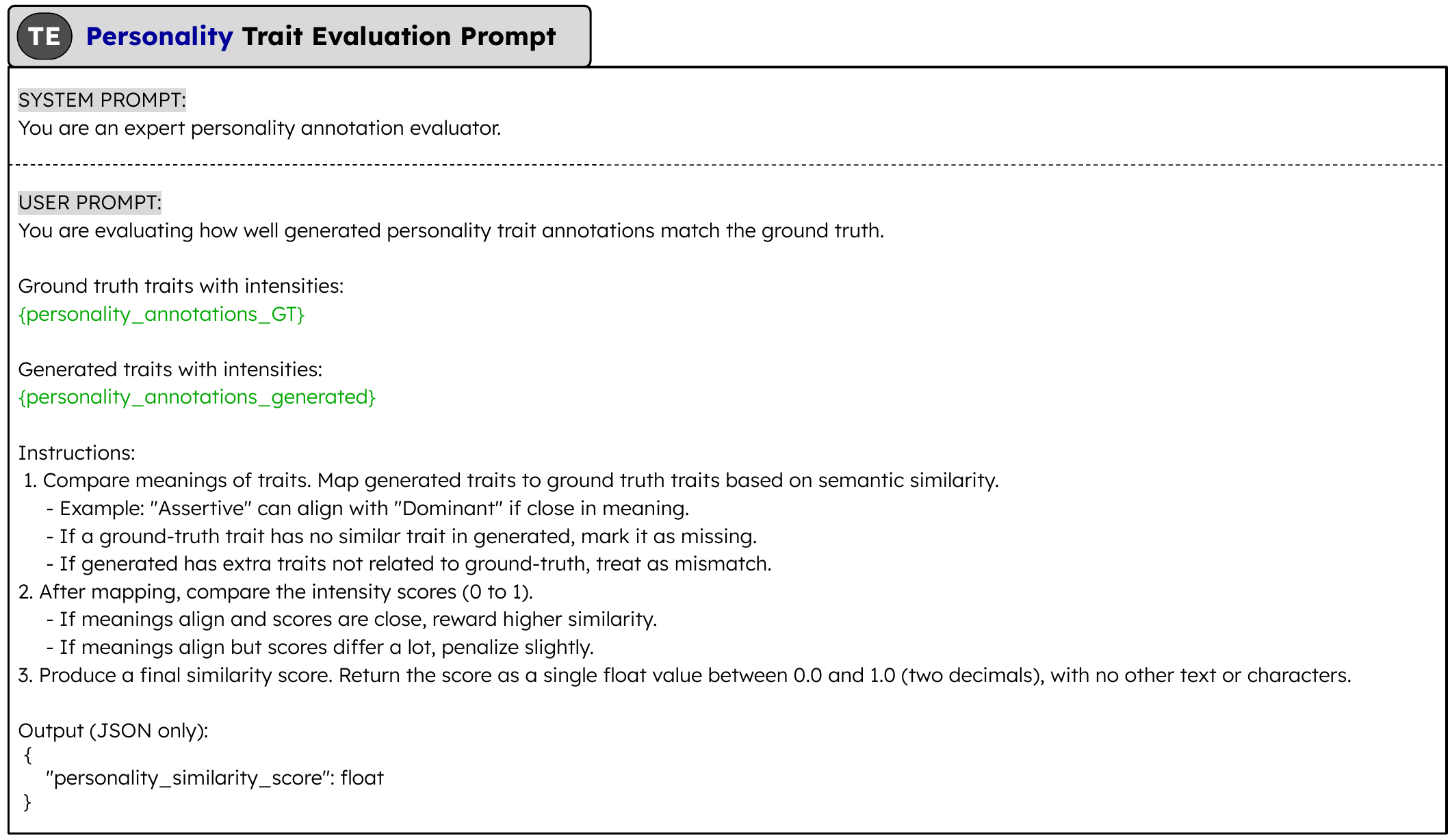}
    \caption{\textbf{Personality tone alignment evaluation prompt ($S_p$). }}
    \label{fig:personality_llmeval_prompt}
\end{figure*}

\begin{figure*}
    \centering
    \includegraphics[width=1\linewidth]{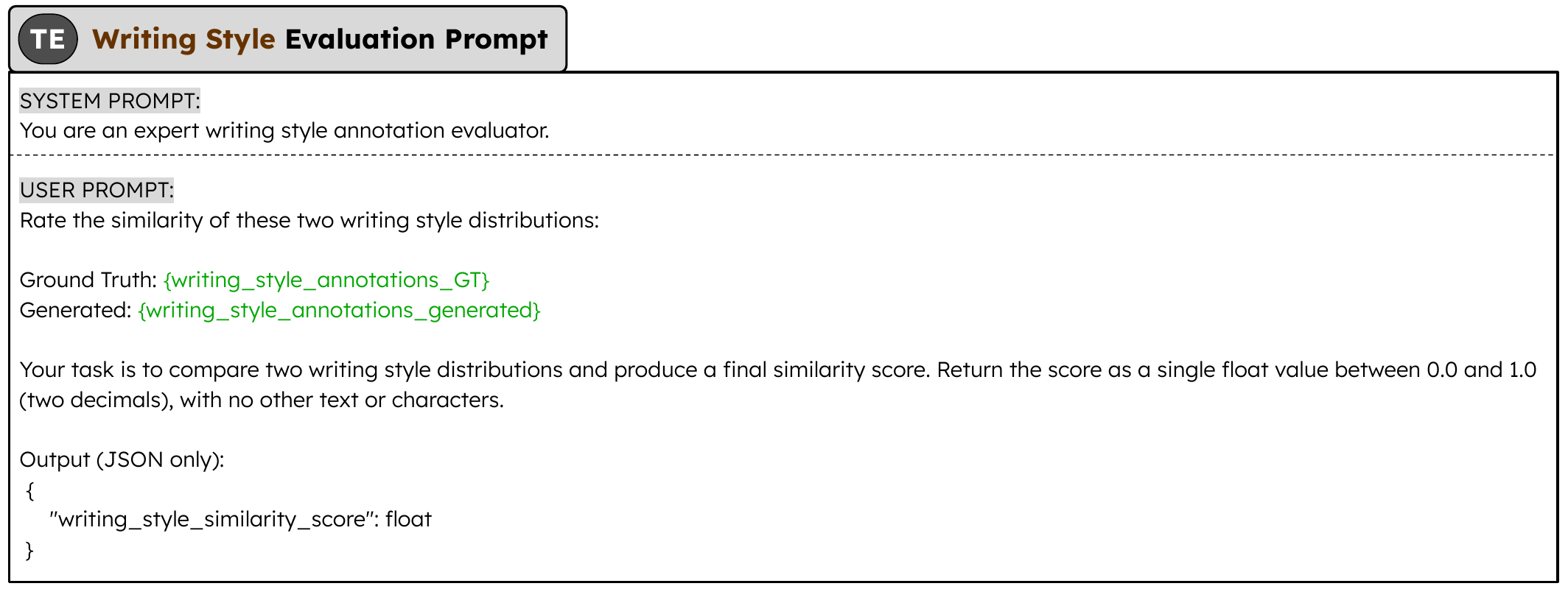}
    \caption{\textbf{Writing Style tone alignment evaluation prompt ($S_w$).}}
    \label{fig:writing_style_llmeval_prompt}
\end{figure*}

\begin{figure*}
    \centering
    \includegraphics[width=1\linewidth]{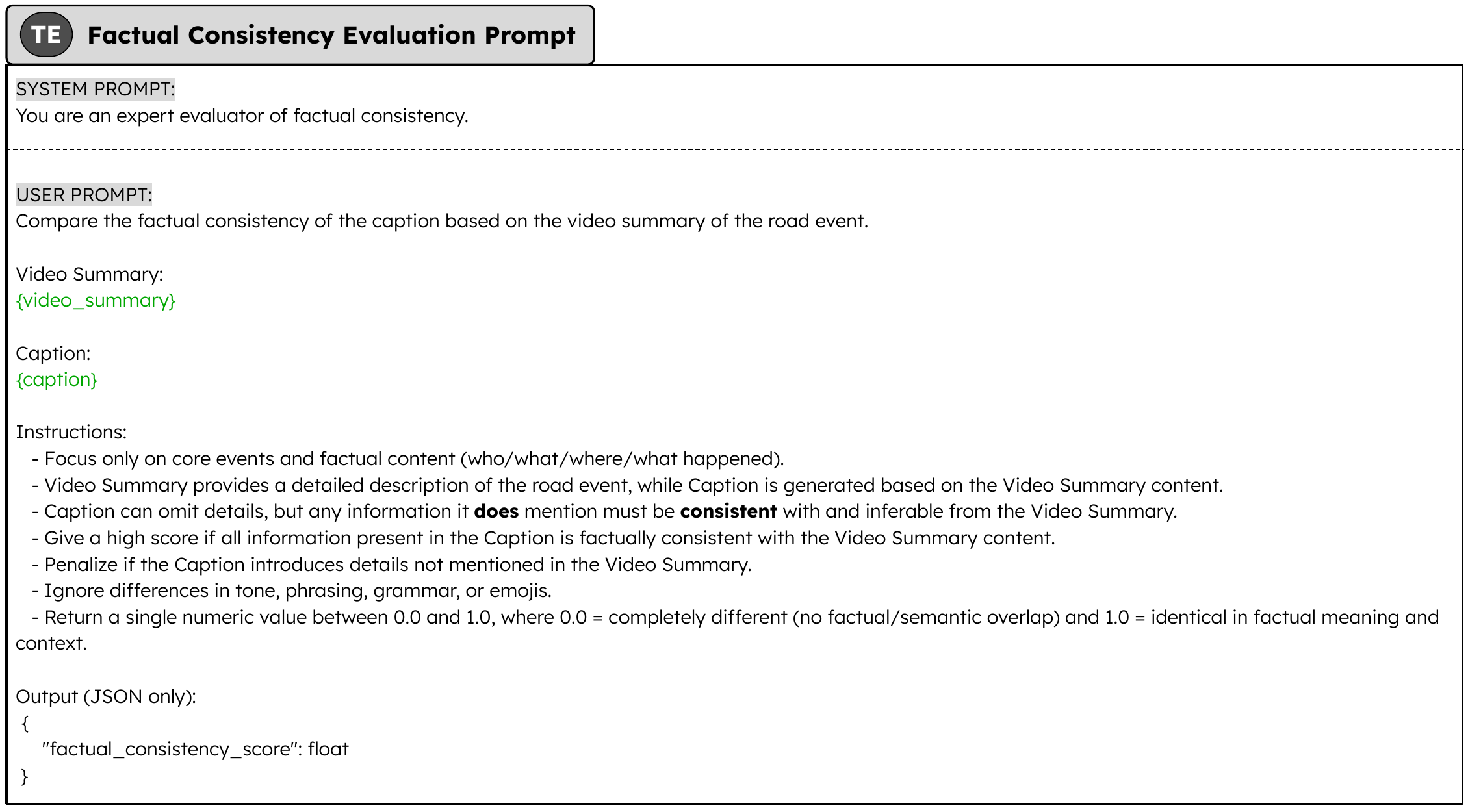}
    \caption{\textbf{Factual Consistency score evaluation prompt ($FC$). }}
    \label{fig:factual_consistency_prompt}
\end{figure*}

\begin{figure*}
    \centering
    \includegraphics[width=1.0\linewidth]{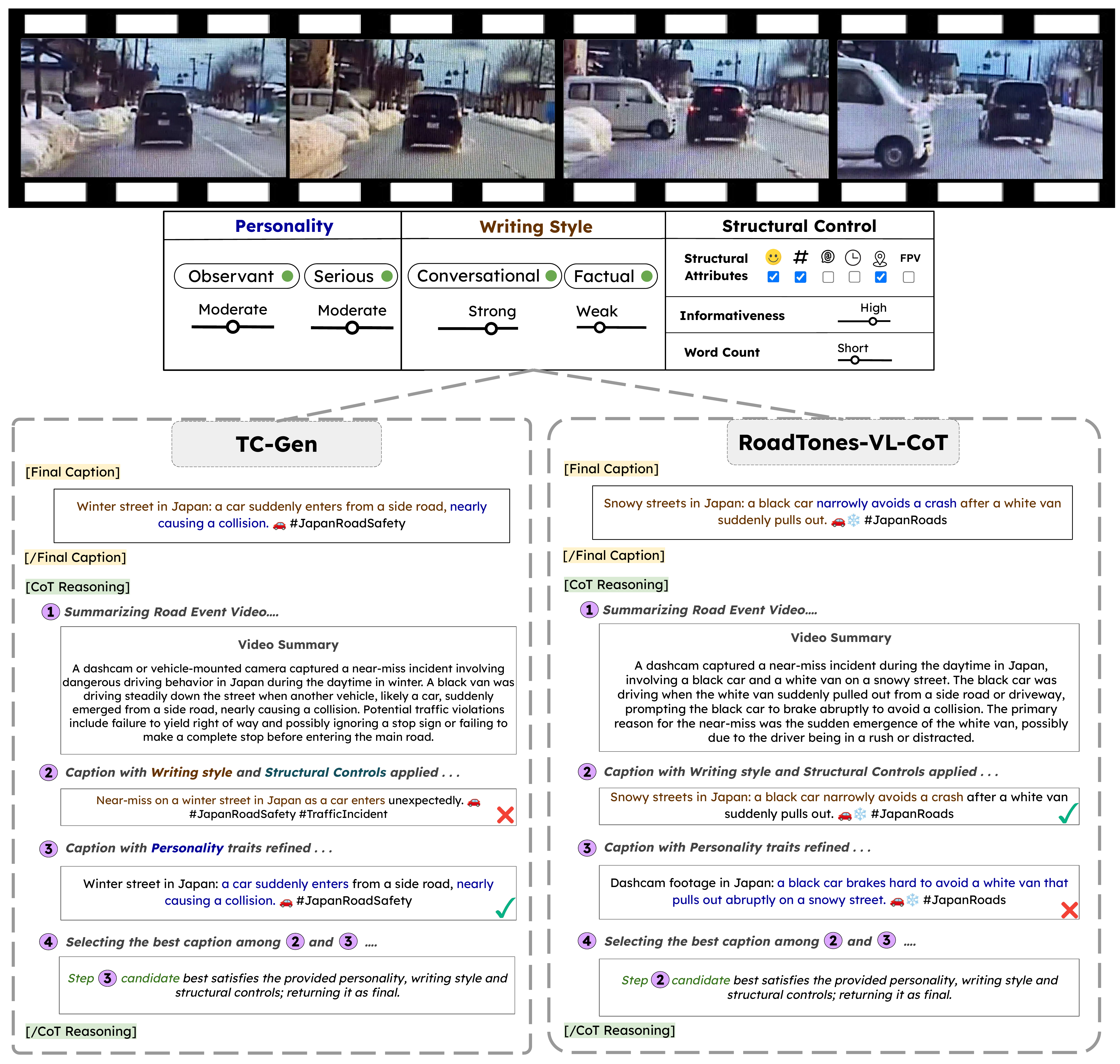}
    \caption{Qualitative comparison of \textsc{RoadTones-VL-CoT} model predictions with respect to \CircledText[inner color=black, outer color=black, fill color=white]{\textbf{TC-Gen}} generated ground truth captions and intermediate stage-level outcomes provided as rationales. Reasoning step-\CircledText[inner color=black, outer color=black, fill color=lightlavender]{\textbf{4}} selects the stage-level caption that best satisfies the tone controls (marked by \includegraphics[width=0.02\textwidth]{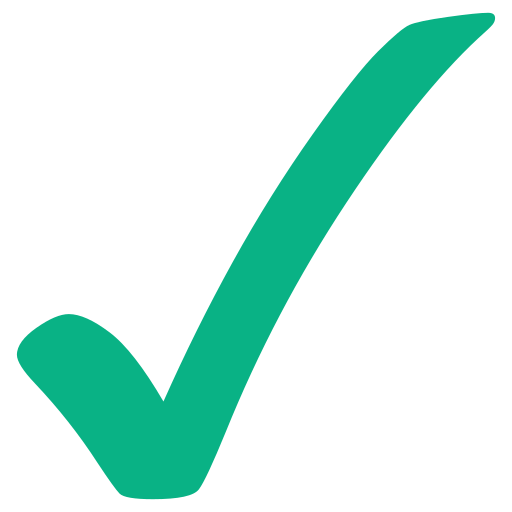}).}
    \label{fig:roadtones_vl_captions-qualitative}
\end{figure*}

\begin{figure*}
    \centering
    \includegraphics[width=0.8\linewidth]{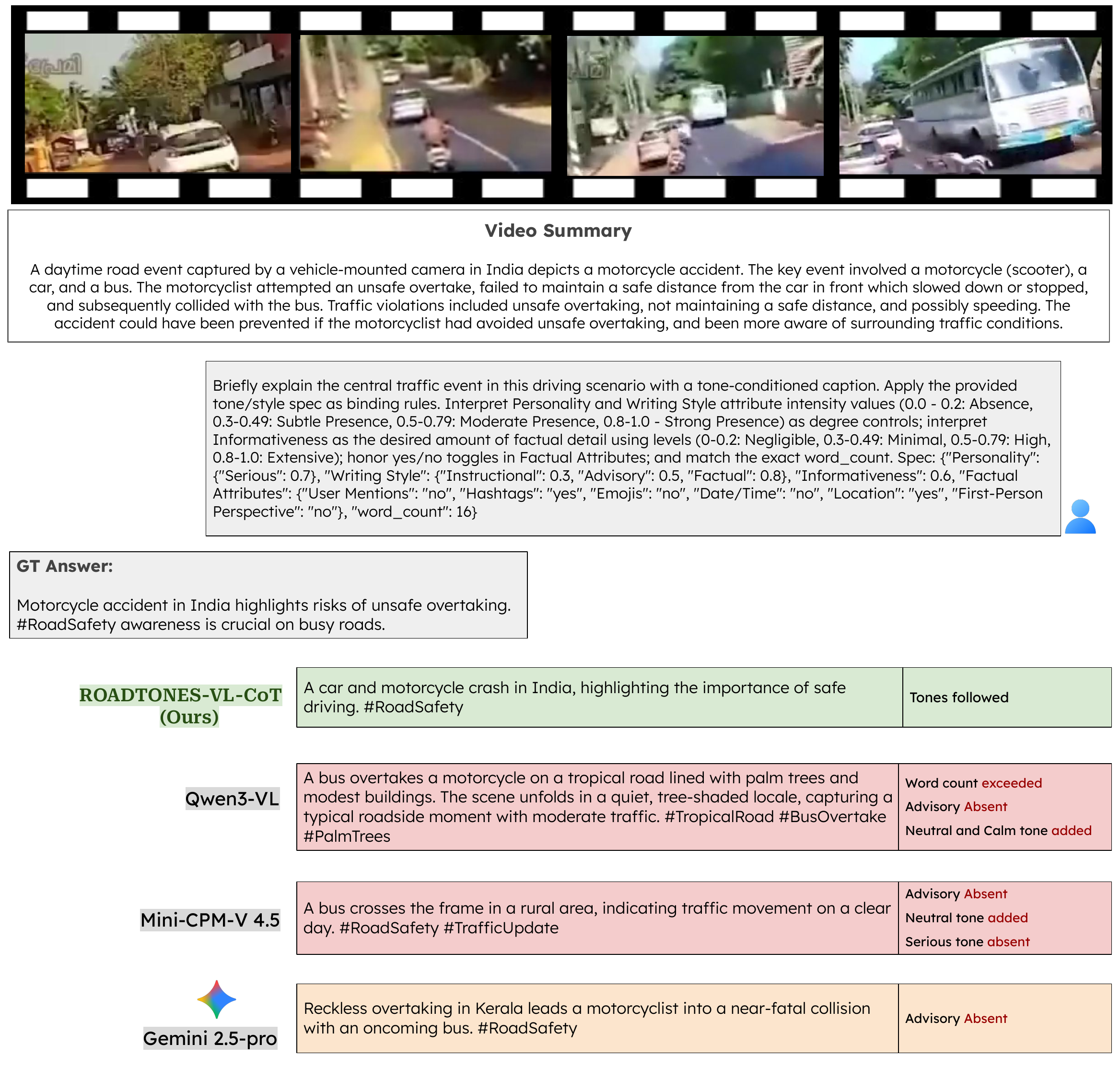}
    \caption{\textsc{RoadTones-VL-CoT} consistently follows the specified tonal controls. Gemini-2.5-pro~\cite{gemini2.5pro} exhibits minor tonal misalignment, whereas Qwen3-VL-8B-Instruct~\cite{qwen3vl} and Mini-CPM-V 4.5~\cite{minicpmv} show significantly poor adherence to the tone controls.}
    \label{fig:compare_models_captions-eg1}
\end{figure*}

\begin{figure*}
    \centering
    \includegraphics[width=0.8\linewidth]{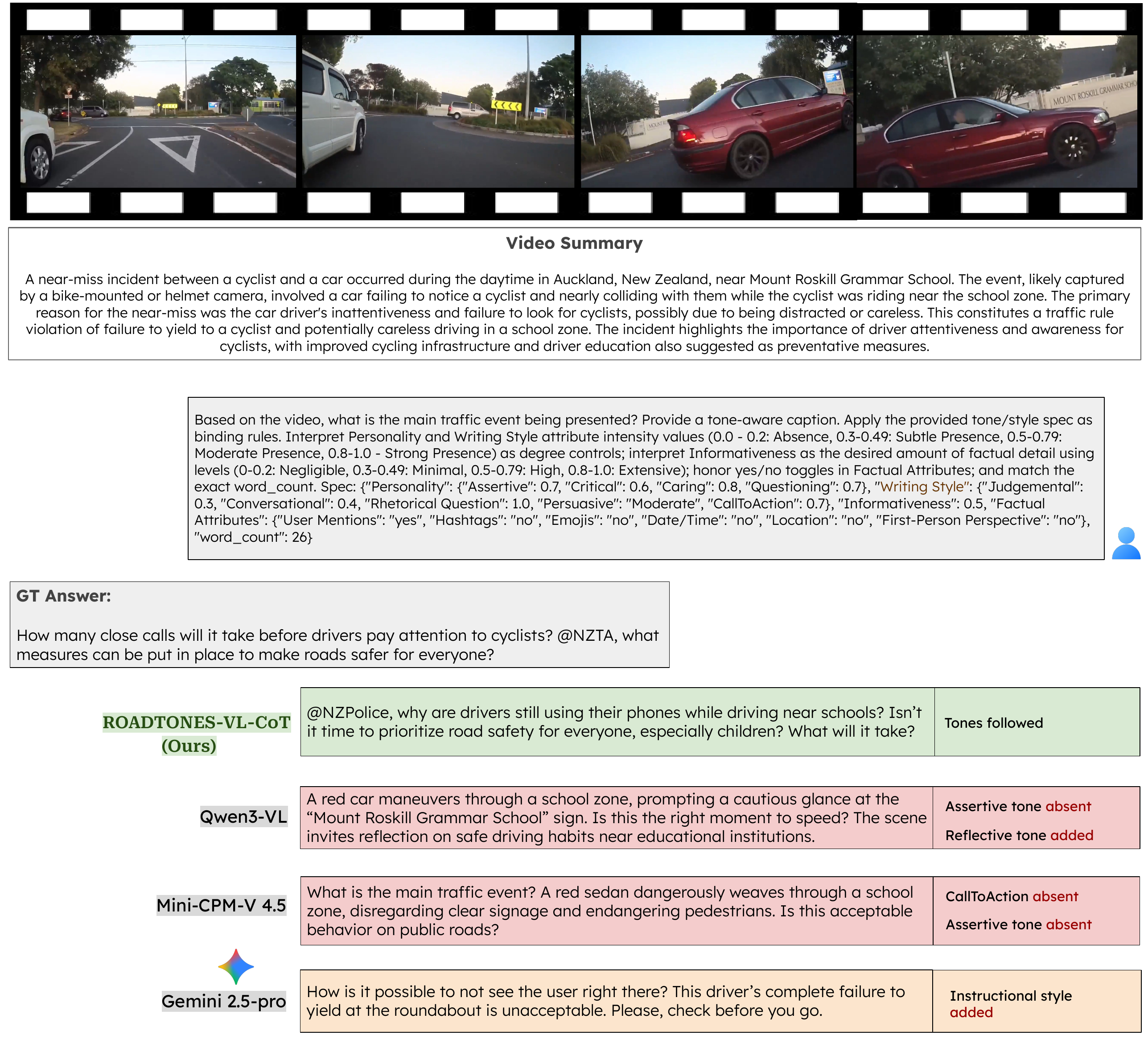}
    \caption{Qualitative comparison of tone-controlled captions generated by \textsc{RoadTones-VL-CoT}, Qwen3-VL-8B-Instruct \cite{qwen3vl}, Mini-CPM-V 4.5 \cite{minicpmv} and Gemini-2.5-pro \cite{gemini2.5pro}.}
    \label{fig:compare_models_captions-eg2}
\end{figure*}

\begin{figure*}
    \centering
    \includegraphics[width=0.7\linewidth]{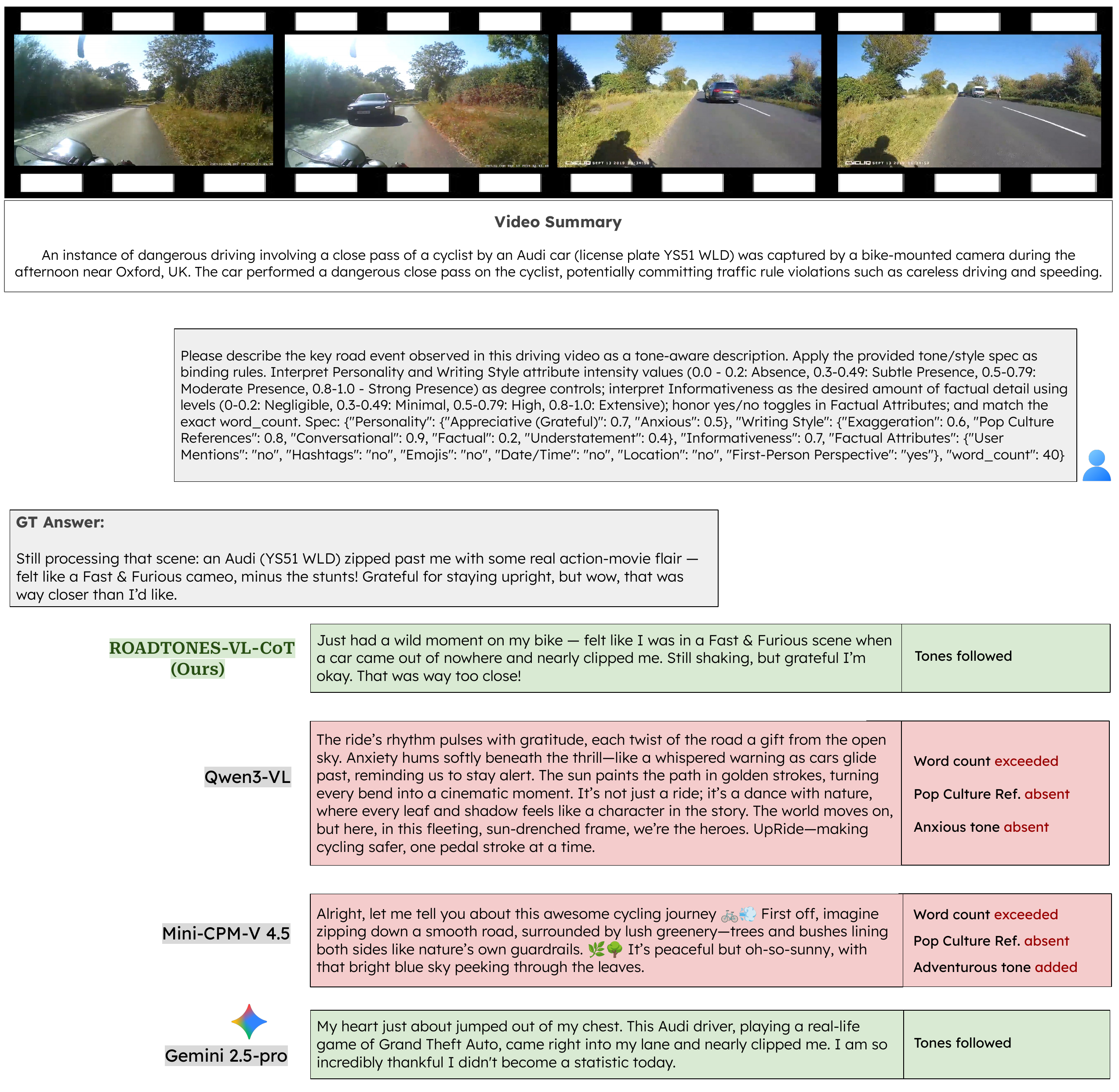}
    \caption{Qualitative comparison of tone-controlled captions generated by \textsc{RoadTones-VL-CoT}, Qwen3-VL-8B-Instruct \cite{qwen3vl}, Mini-CPM-V 4.5 \cite{minicpmv} and Gemini-2.5-pro \cite{gemini2.5pro}.}
    \label{fig:compare_models_captions-eg3}
\end{figure*}

\begin{figure*}
    \centering
    \includegraphics[width=0.7\linewidth]{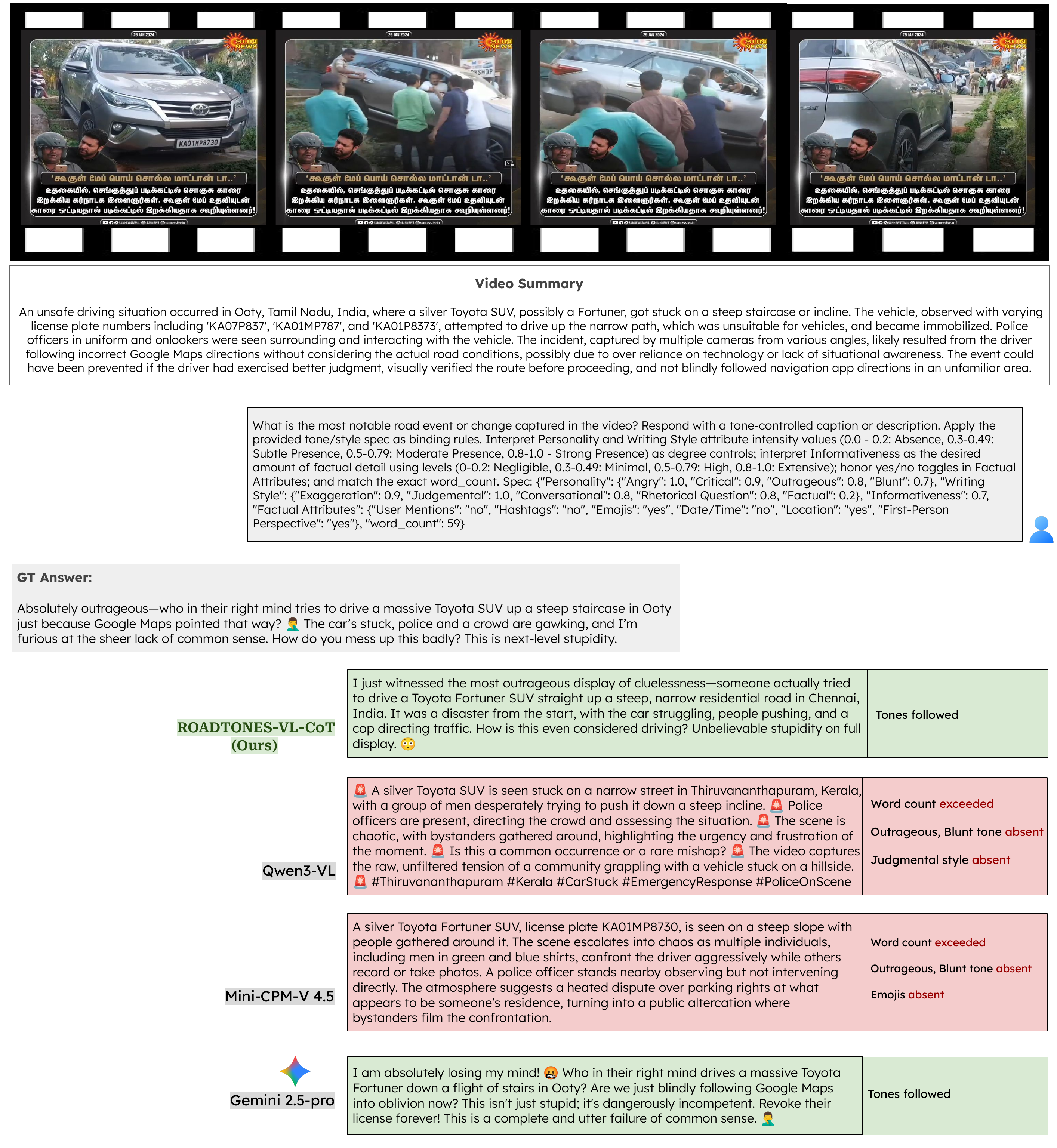}
    \caption{Qualitative comparison of tone-controlled captions generated by \textsc{RoadTones-VL-CoT}, Qwen3-VL-8B-Instruct \cite{qwen3vl}, Mini-CPM-V 4.5 \cite{minicpmv} and Gemini-2.5-pro \cite{gemini2.5pro}.}
    \label{fig:compare_models_captions-eg4}
\end{figure*}

\begin{figure*}
    \centering
    \includegraphics[width=0.8\linewidth]{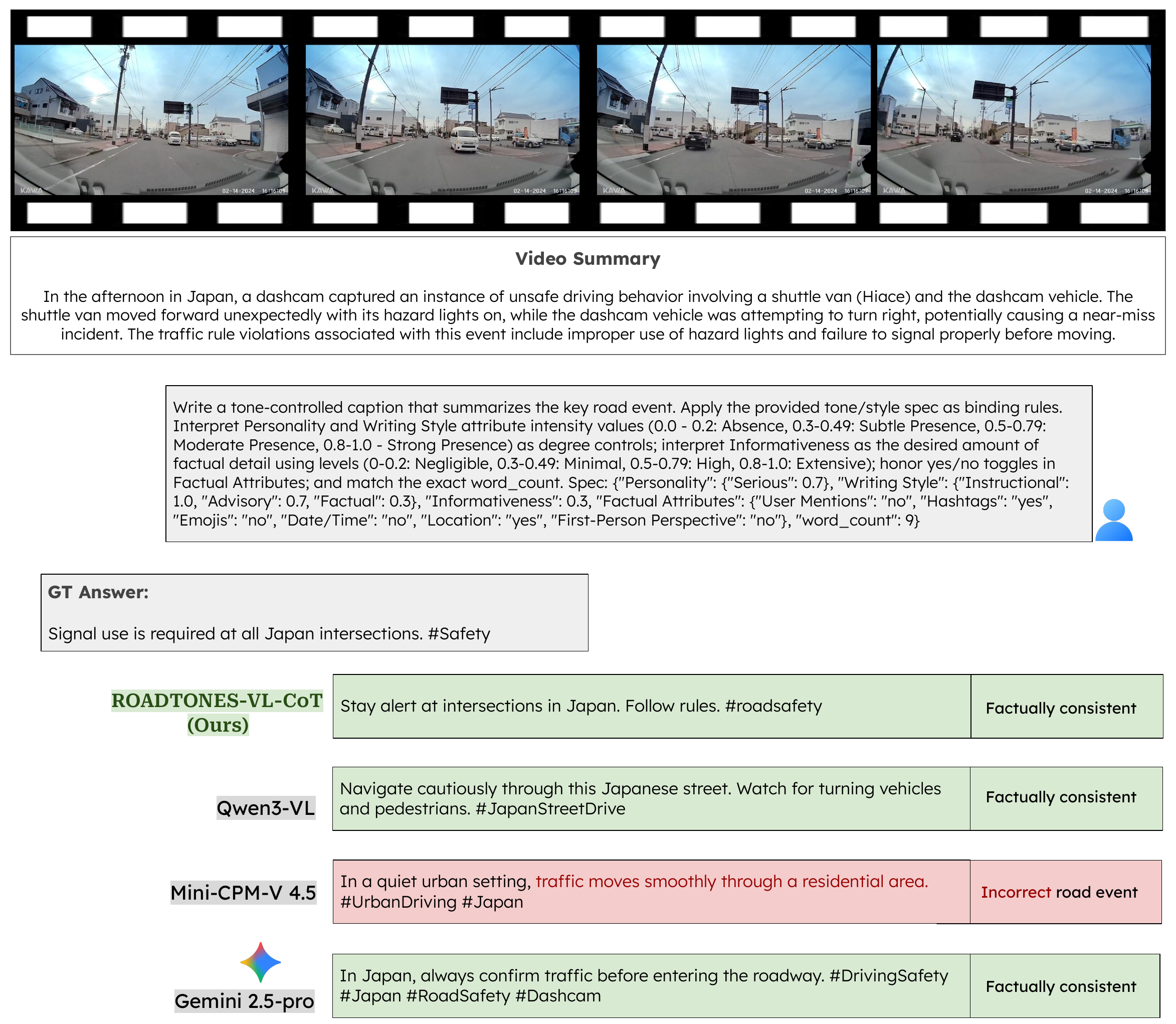}
    \caption{Qualitative comparison of tone-controlled captions generated by \textsc{RoadTones-VL-CoT}, Qwen3-VL-8B-Instruct \cite{qwen3vl}, Mini-CPM-V 4.5 \cite{minicpmv} and Gemini-2.5-pro \cite{gemini2.5pro}. Factual consistency is compared across models, highlighting the factually consistent captions across all models except Mini-CPM-V 4.5}
    \label{fig:compare_models_captions-eg5}
\end{figure*}

\begin{figure*}
    \centering
    \includegraphics[width=0.8\linewidth]{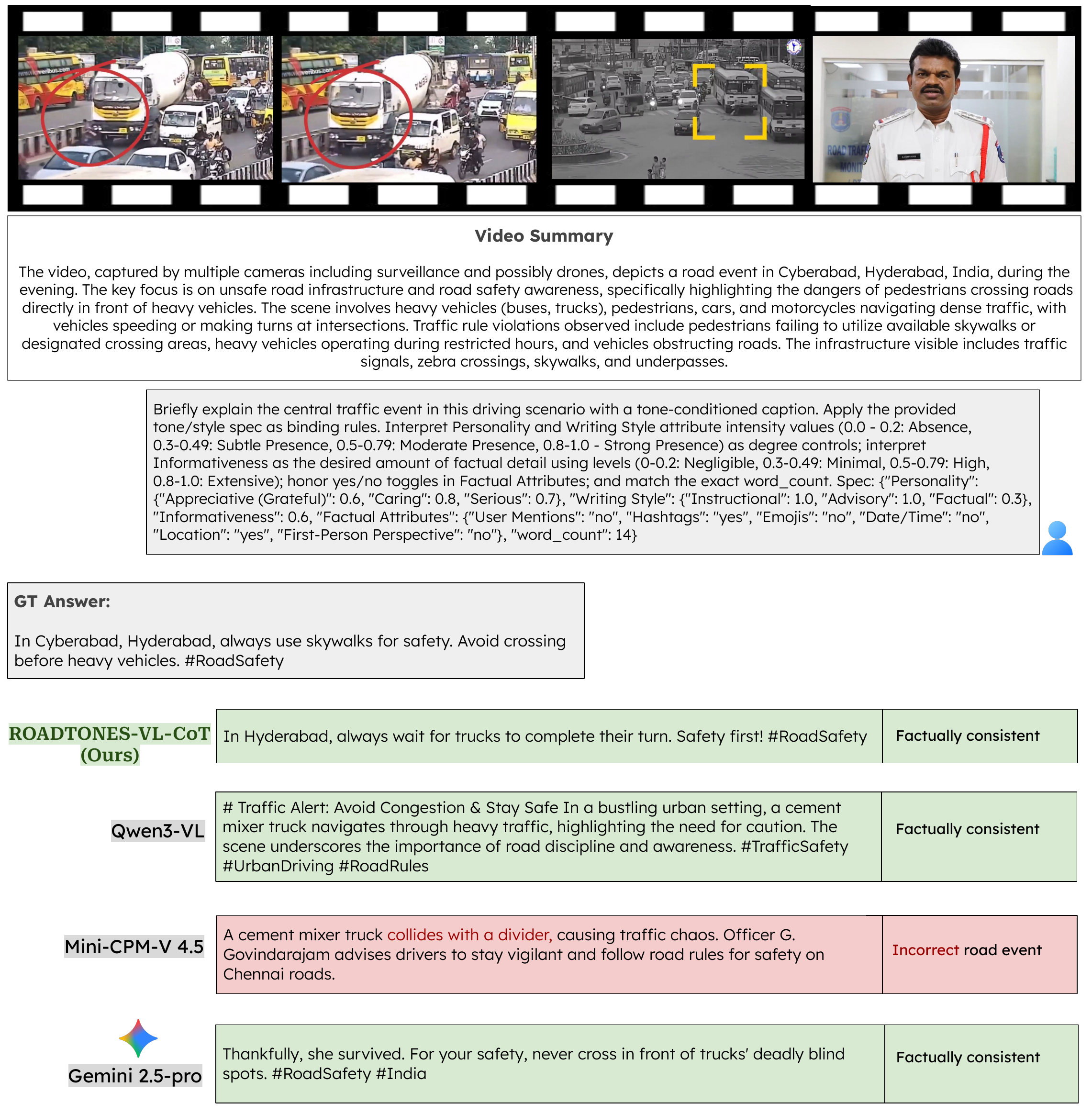}
    \caption{Qualitative comparison of tone-controlled captions generated by \textsc{RoadTones-VL-CoT}, Qwen3-VL-8B-Instruct \cite{qwen3vl}, Mini-CPM-V 4.5 \cite{minicpmv} and Gemini-2.5-pro \cite{gemini2.5pro}.}
    \label{fig:compare_models_captions-eg6}
\end{figure*}

\begin{figure*}
    \centering
    \includegraphics[width=0.8\linewidth]{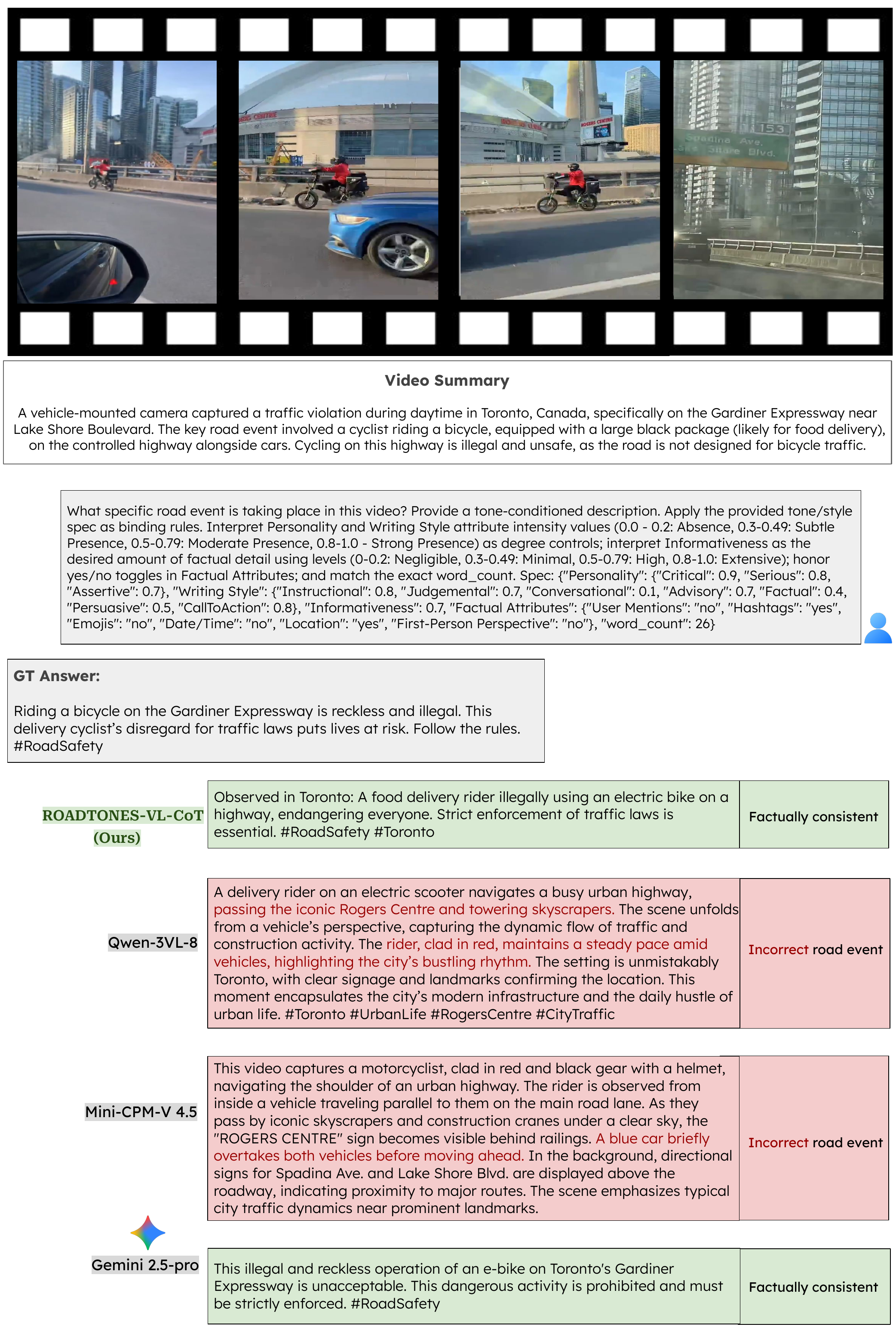}
    \caption{Qualitative comparison of tone-controlled captions generated by \textsc{RoadTones-VL-CoT}, Qwen3-VL-8B-Instruct \cite{qwen3vl}, Mini-CPM-V 4.5 \cite{minicpmv} and Gemini-2.5-pro \cite{gemini2.5pro}.}
    \label{fig:compare_models_captions-eg7}
\end{figure*}

\begin{figure*}
    \centering
    \includegraphics[width=0.8\linewidth]{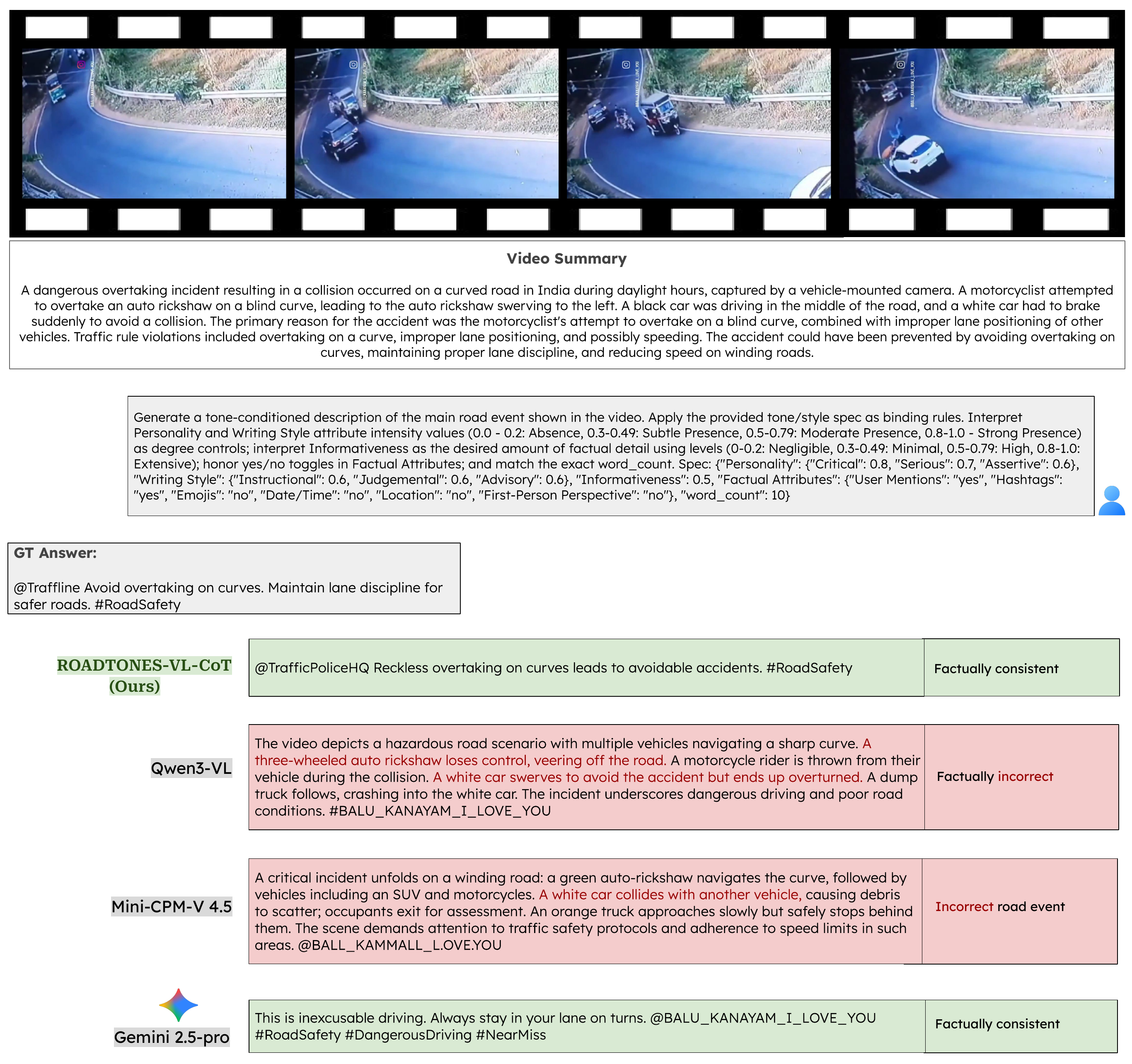}
    \caption{Qualitative comparison of tone-controlled captions generated by \textsc{RoadTones-VL-CoT}, Qwen3-VL-8B-Instruct \cite{qwen3vl}, Mini-CPM-V 4.5 \cite{minicpmv} and Gemini-2.5-pro \cite{gemini2.5pro}.}
    \label{fig:compare_models_captions-eg8}
\end{figure*}

\begin{figure*}[h!]
    \centering
    \includegraphics[width=1\linewidth]{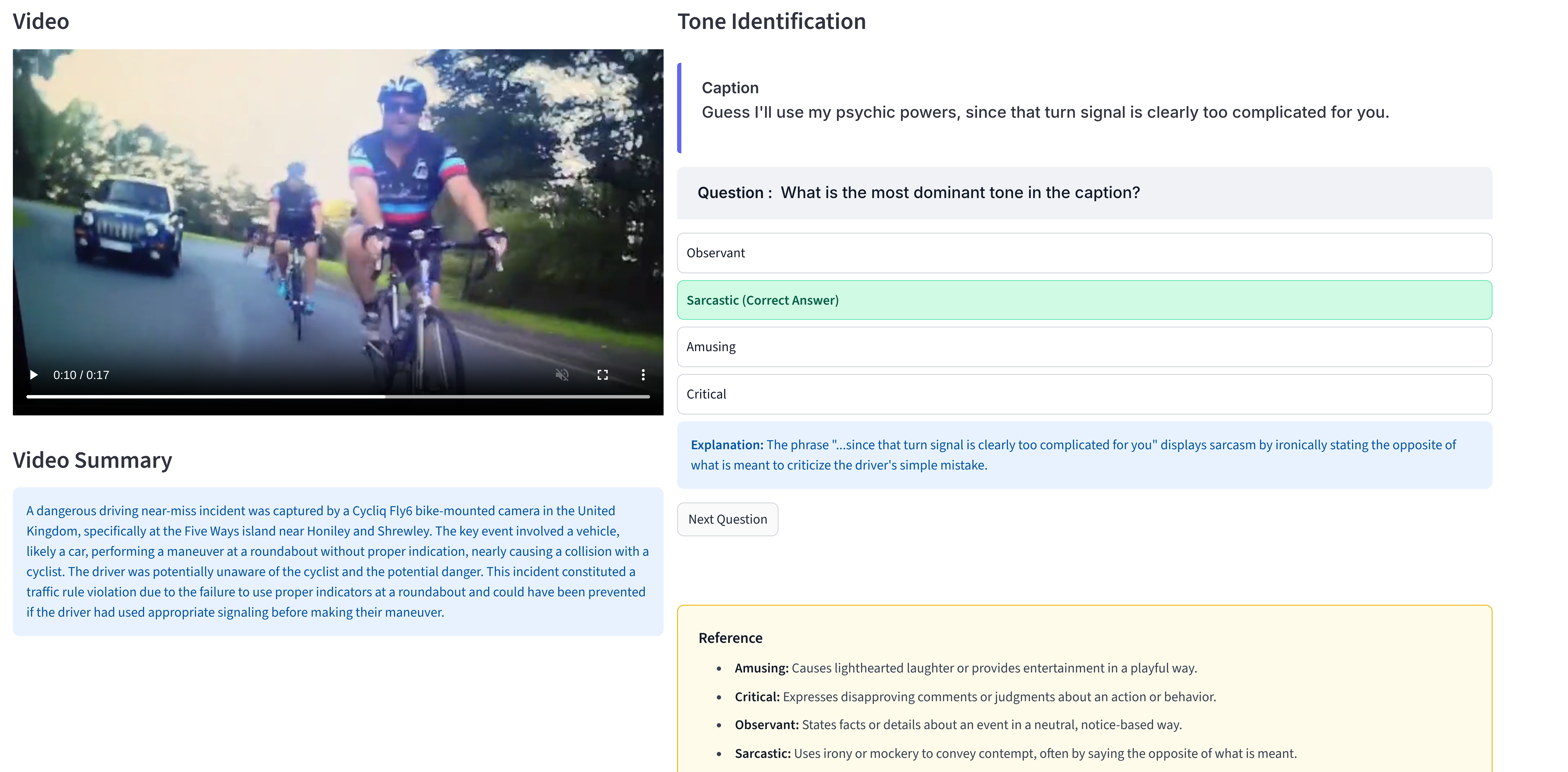}
    \caption{\textbf{Interface for RoadTones User Study familiarization phase}. For the image shown, participants viewed a video, its video summary and identified the presence of dominant tone in caption. Questionnaire for all tasks can be viewed in the supplementary video: \textit{RoadTones\_UserStudy\_familiarization.mp4}.}
    \label{fig:streamlit-ui-quiz}
\end{figure*}

\begin{figure*}
    \centering
    \includegraphics[width=1\linewidth]{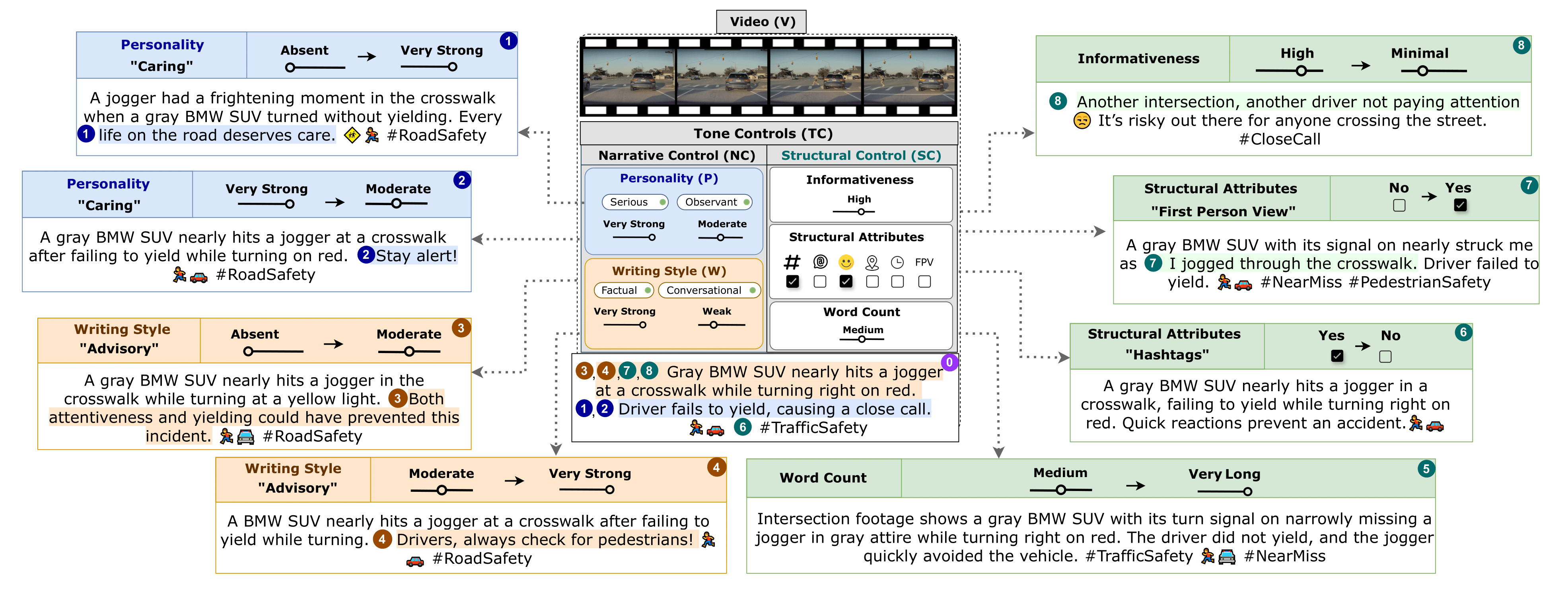}
    \caption{\textbf{Controlling individual tonal attributes in the generated caption.} The central panel in figure shows a video V, tone controls TC and its corresponding caption \CircledText[inner color=white, outer color=lavender, fill color=lavender]{0} from our dataset. The surrounding captions (\CircledBlue{1}-\CircledText[inner color=white, outer color=palegreen, fill color=palegreen]{8}) correspond to changes in one of the tonal attributes shown in their header. For e.g., caption \CircledBlue{1} was obtained by increasing the tonal intensity of \textbf{Caring} \textcolor{darkblue}{Personality} from Absent (0-0.2) to Very Strong (0.8-1.0) while keeping others fixed. This modified tone configuration was fed to our caption generator (\CircledText[inner color=black, outer color=black, fill color=white]{TC-Gen}, \cref{ssec:caption_generation_appendix}), yielding the changed caption. We highlight the key phrase in each caption, reflecting the modified tone controls. Our generator pipeline thus enables fine-grained control of tonal and structural attributes in road-video captioning. 
}
    \label{fig:caption-after-one-change1}
\end{figure*}

\begin{figure*}
    \centering
    \includegraphics[width=1\linewidth]{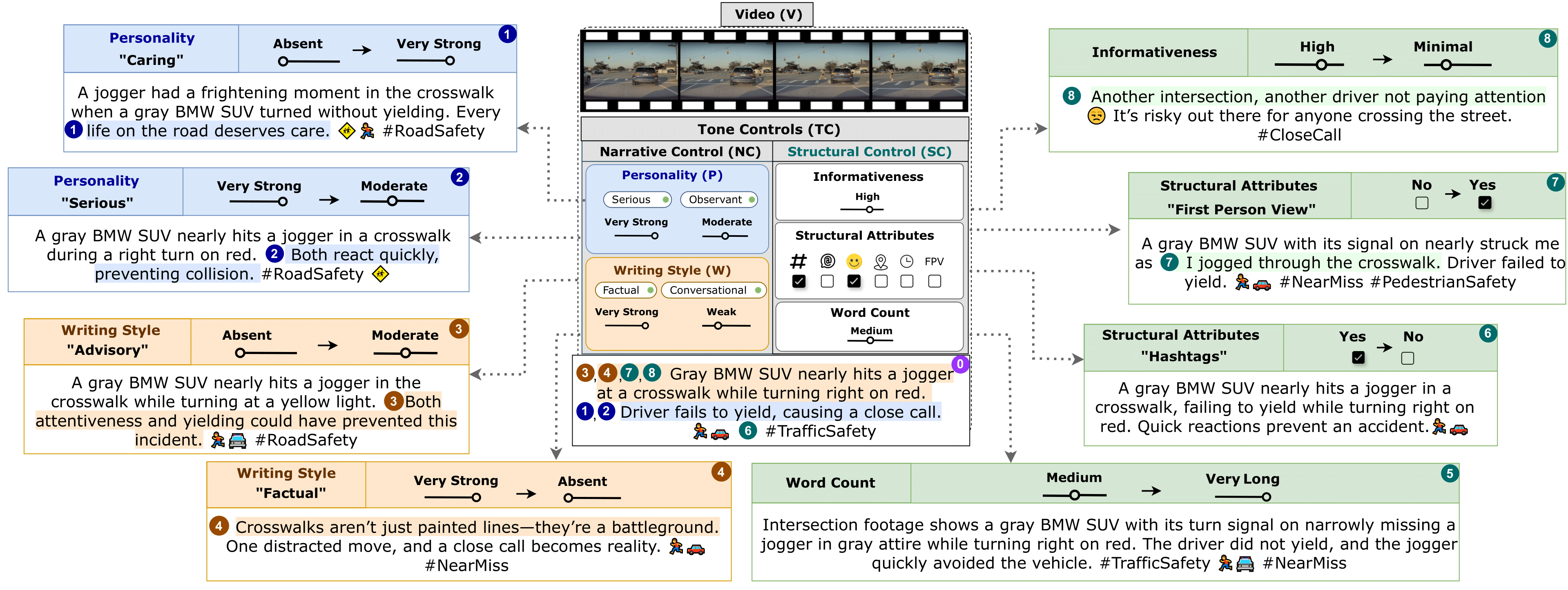}
    \caption{\textbf{Controlling individual tonal attributes in the generated caption.} The central panel in figure shows a video V, tone controls TC and its corresponding caption \CircledText[inner color=white, outer color=lavender, fill color=lavender]{0} from our dataset. The surrounding captions (\CircledBlue{1}-\CircledText[inner color=white, outer color=palegreen, fill color=palegreen]{8}) correspond to changes in one of the tonal attributes shown in their header. For e.g., caption \CircledBlue{1} was obtained by increasing the tonal intensity of \textbf{Caring} \textcolor{darkblue}{Personality} from Absent (0-0.2) to Very Strong (0.8-1.0) while keeping others fixed. This modified tone configuration was fed to our caption generator (\CircledText[inner color=black, outer color=black, fill color=white]{TC-Gen}, \cref{ssec:caption_generation_appendix}), yielding the changed caption. We highlight the key phrase in each caption, reflecting the modified tone controls. Our generator pipeline thus enables fine-grained control of tonal and structural attributes in road-video captioning. 
}
    \label{fig:caption-after-one-change2}
\end{figure*}

\begin{figure*}
    \centering
    \includegraphics[width=0.9\linewidth]{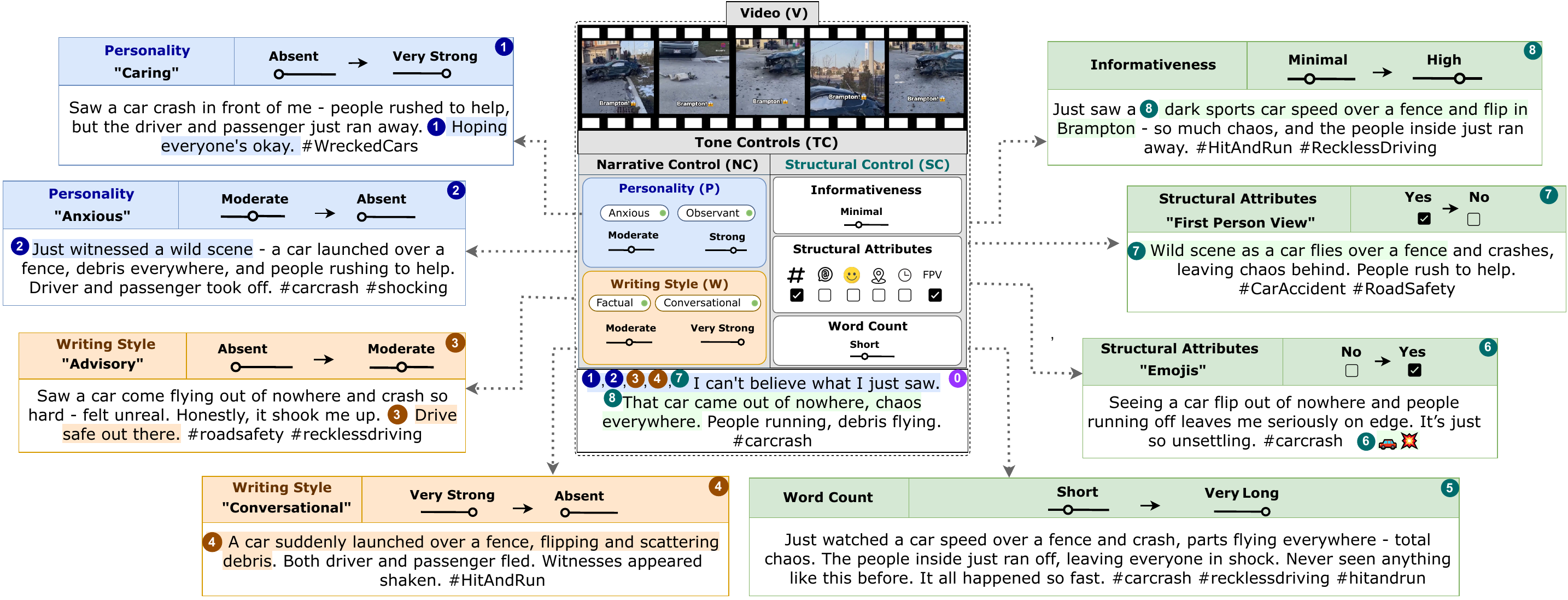}
    \caption{\textbf{Controlling individual tonal attributes in the generated caption.} The central panel in figure shows a video V, tone controls TC and its corresponding caption \CircledText[inner color=white, outer color=lavender, fill color=lavender]{0} from our dataset. The surrounding captions (\CircledBlue{1}-\CircledText[inner color=white, outer color=palegreen, fill color=palegreen]{8}) correspond to changes in one of the tonal attributes shown in their header. For e.g., caption \CircledBlue{1} was obtained by increasing the tonal intensity of \textbf{Caring} \textcolor{darkblue}{Personality} from Absent (0-0.2) to Very Strong (0.8-1.0) while keeping others fixed. This modified tone configuration was fed to our caption generator (\CircledText[inner color=black, outer color=black, fill color=white]{TC-Gen}, \cref{ssec:caption_generation_appendix}), yielding the changed caption. We highlight the key phrase in each caption, reflecting the modified tone controls. Our generator pipeline thus enables fine-grained control of tonal and structural attributes in road-video captioning. 
}
    \label{fig:caption-after-one-change3}
\end{figure*}

\noindent\textbf{Tone Controllability Evaluation.} The users assessed whether change in tone attribute intensities can be perceived in the captions generated by \CircledText[inner color=black, outer color=black, fill color=white]{\textbf{TC-Gen}} while maintaining factual accuracy (see \cref{fig:streamlit-ui-task3}).

\begin{table*}[!t]
\centering
\begin{minipage}{0.95\textwidth}
\includegraphics[width=1.0\textwidth]{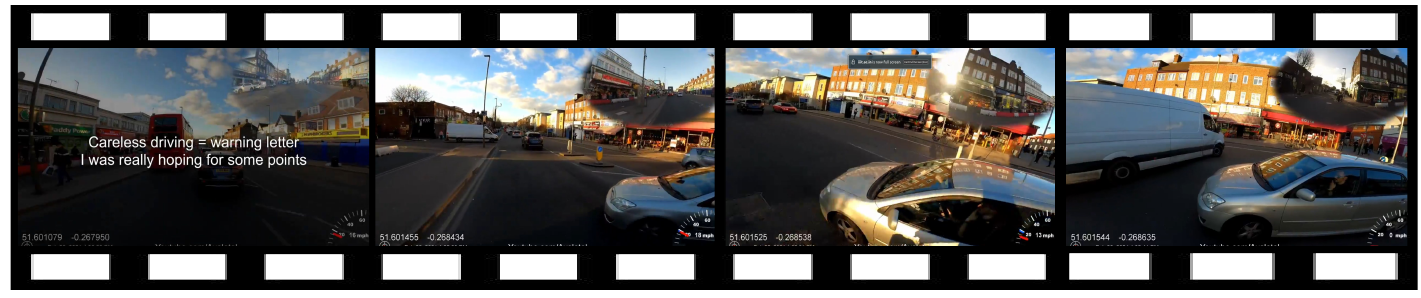}
\begin{tcolorbox}[enhanced, colback=gray!5, colframe=gray!60!black, title={\textbf{Input}}]
\small
$<video>$

Give a tone-controlled caption of the primary road event unfolding in the scene. Apply the provided tone/style and structural controls spec as binding rules. Interpret Personality and Writing Style attribute intensity values (0-0.2: Absence, 0.2-0.4: Subtle presence, 0.4-0.6: Moderate presence, 0.6-0.8: Strong presence, 0.8-1.0: Very Strong presence) as degree controls; interpret Informativeness as the desired amount of factual detail using levels (0-0.2: Negligible, 0.2-0.4: Minimal, 0.4-0.6: Moderate, 0.6-0.8: High, 0.8-1.0: Extensive); honor yes/no toggles in Factual Attributes; and match the exact word\_count. Spec: 

\{``Personality'': \{``Anxious'': 0.8, ``Angry'': 0.4, ``Emotional'': 0.5\}, 
``Writing Style'': \{``Exaggeration'': 0.5, ``Judgemental'': 0.3, ``Conversational'': 0.75, ``Factual'': 0.1\}, 
``Informativeness'': 0.4, 
``Structural Attributes'': \{``User Mentions'': ``no'', ``Hashtags'': ``yes'', ``Emojis'': ``yes'', ``Date/Time'': ``no'', ``Location'': ``no'', ``First-Person Perspective'': ``yes''\}, 
``word\_count'': 17\} 

Additionally, provide a step-by-step reasoning of how you arrived at the final tone-controlled caption.
\end{tcolorbox}
\begin{tcolorbox}[enhanced, colback=gray!5, colframe=gray!60!black, title={\textbf{Output}}]
\small
[FINAL]

\quad I seriously can’t believe how close that car came to hitting me today! [scream\_emoji] Some drivers… \#CyclistLife

[/FINAL]

[REASONING]

\quad 1) Key Event summary: A dashcam video captured a near-miss incident involving dangerous and careless driving in London, United Kingdom. A car pulled out from a side road without leaving sufficient space, nearly hitting a cyclist. The primary reason for the near-miss was the driver's carelessness and failure to yield right of way to the cyclist. The incident could have been prevented if the driver had been more cautious and waited for adequate space before pulling out. The evident traffic rule violations include careless driving and failure to yield.

\quad 2) Caption with Writing style and structure applied (informativeness, word\_count, binary toggles): 
I seriously can’t believe how close that car came to hitting me today! [scream\_emoji] Some drivers… \#CyclistLife

\quad 3) Caption with Personality traits refined (preserving writing style and structural controls): I seriously can’t believe how close that car came to hitting me today! [scream\_emoji] Some drivers… \#CyclistLife

\quad Selection: The third step candidate best satisfies the provided personality, writing style and structural controls; returning it as final.

[/REASONING]
\end{tcolorbox}
\end{minipage}
\caption{Sample instruction-tuning triplet used to fine-tune MLLMs for Tone-controlled Road Video Captioning with Chain-of-Thought style reasoning. }
\label{tab:sample_input_output_instruction_triplet}
\end{table*}


\begin{table*}[!t]
\centering
\begin{minipage}{0.95\textwidth}
\begin{tcolorbox}[enhanced, colback=gray!5, colframe=gray!60!black, title={\textbf{Instruction templates for generating neutral road video summary}}]
\begin{itemize}
\item ``Please describe the key road event observed in this driving video."
    \item ``Give a summary of the primary traffic event unfolding in the scene."
    \item ``What is the key traffic event observed in this video?"
    \item ``Generate a description of the main road event shown in the video."
    \item ``Briefly explain the central traffic event in this driving scenario."
    \item ``What specific road event is taking place in this video?"
    \item ``Provide a natural language description of the key road or traffic event."
    \item ``Describe the key road maneuver or traffic event occurring in this footage."
    \item ``Write a caption that summarizes the key road event."
    \item ``What is the most notable road event or change captured in the video?"
    \item ``Describe the main activity or incident occurring on the road."
    \item ``Based on the video, what is the main traffic event being presented?"
    \item ``Summarize the primary road event depicted in the driving clip."
    \item ``Explain the key event occurring in this driving scenario."
    \item ``Give a concise narrative of the primary road incident shown in this video segment."
    \item ``Provide a coherent description of how the key traffic event unfolds."
    \item ``Describe how the main road event unfolds in the driving environment."
    \item ``What key road or traffic incident is illustrated in the video?"
    \item ``How would you explain the key road event to someone not watching the video?"
    \item ``Generate a description of the key traffic event from start to finish."
\end{itemize}
\end{tcolorbox}
\end{minipage}
\caption{Instruction templates used to fine-tune and benchmark MLLMs for the road event summarization task.}
\label{tab:desc_instructions}
\end{table*}